\documentclass[3p,times]{elsarticle}\usepackage{lipsum}
\makeatletter
\def\ps@pprintTitle{
 \let\@oddhead\@empty
 \let\@evenhead\@empty
 \def\@oddfoot{}
 \let\@evenfoot\@oddfoot}
\makeatother

\usepackage{amssymb}

\biboptions{sort&compress,comma}

\usepackage[figuresright]{rotating}

\usepackage{amsmath}

\usepackage{xcolor,graphicx,float,verbatim,url,color}

\usepackage{setspace}
\usepackage{picins}

\usepackage{algorithm}  
\usepackage{algorithmicx} 
\usepackage{algpseudocode}  
\usepackage{caption}

\makeatletter

\newcounter{phase}[algorithm]
\newlength{\phaserulewidth}
\newcommand{\phaseTitle}{Stage} 

\DeclareMathOperator*{\argmin}{arg\,min}

\newcommand{\phase}[1]{%
  \vspace{-1.25ex}
  \Statex\leavevmode\llap{\rule{\dimexpr\labelwidth+\labelsep}{\phaserulewidth}}\rule{\linewidth}{\phaserulewidth}
  \Statex\strut\refstepcounter{phase}\item[\textbf{\phaseTitle~\thephase:~}\textit{~#1}] 
  \vspace{-1.25ex}\Statex\leavevmode\llap{\rule{\dimexpr\labelwidth+\labelsep}{\phaserulewidth}}\rule{\linewidth}{\phaserulewidth}}

\begin{document}

\begin{frontmatter}




\title{Graph Autoencoder-Based Unsupervised Feature Selection with Broad and Local Data Structure Preservation}


\author[mailaddress]{Siwei Feng}
\ead{siwei@umass.edu}
\author[mailaddress]{Marco F. Duarte}
\ead{mduarte@ecs.umass.edu}

\address[mailaddress]{Department of Electrical and Computer Engineering, University of Massachusetts Amherst, Amherst, MA 01003}

\begin{abstract}
Feature selection is a dimensionality reduction technique that selects a subset of representative features from high-dimensional data by eliminating irrelevant and redundant features. Recently, feature selection combined with sparse learning has attracted significant attention due to its outstanding performance compared with traditional feature selection methods that ignores correlation between features. These works first map data onto a low-dimensional subspace and then select features by posing a sparsity constraint on the transformation matrix. However, they are restricted by design to linear data transformation, a potential drawback given that the underlying correlation structures of data are often non-linear. To leverage a more sophisticated embedding, we propose an autoencoder-based unsupervised feature selection approach that leverages a single-layer autoencoder for a joint framework of feature selection and manifold learning. More specifically, we enforce column sparsity on the weight matrix connecting the input layer and the hidden layer, as in previous work. Additionally, we include spectral graph analysis on the projected data into the learning process to achieve local data geometry preservation from the original data space to the low-dimensional feature space. Extensive experiments are conducted on image, audio, text, and biological data. The promising experimental results validate the superiority of the proposed method.
\end{abstract}


\end{frontmatter}


\section{Introduction}
\label{intr}
In recent years, high-dimensional data can be found in many areas such as computer vision \cite{wright2010sparse, turaga2008statistical, arandjelovic2005face}, {{pattern recognition \cite{wang2015semantic, wang2015semi, wang2016multimodal, wang2016semi}}}, data mining \cite{agrawal1998automatic}, etc. {{High dimensionality}} enables data to include more information. However, learning high-dimensional data often suffer from several issues. For example, with a fixed number of training data, a large data dimensionality can cause the so-called Hughes phenomenon, i.e., a reduction in the generalization of the learned models due to overfitting during the training procedure compared with lower dimensional data~\cite{pal2010feature}. Moreover, high-dimensional data tend to include significant redundancy in adjacent features, or even noise, which leads to large amounts of useless or even harmful information being processed, stored, and transmitted~\cite{liu2011feature, jimenez1998supervised}. All these issues present challenges to many conventional data analysis problems. Moreover, several papers in the literature have shown that the intrinsic dimensionality of high-dimensional data is actually small~\cite{wang2016semi,lu2013sparse, zhu2011missing, yang2015multi}. Thus, dimensionality reduction is a popular preprocessing step for high-dimensional data analysis, which decreases time for data processing and also improves generalization of learned models. \par
Feature selection~\cite{song2007supervised, sotoca2010supervised, thoma2009near, zhao2007semi, xu2010discriminative, kong2010semi} is a set of frequently used dimensionality reduction approaches that aim at selecting a subset of features. Feature selection has the advantage of preserving the same feature space as that of raw data. Feature selection methods can be categorized into groups based on different criteria summarized below; refer to~\cite{li2016feature} for a detailed survey on feature selection. 
\begin{itemize}
\item{\textbf{Label Availability.}} Based on the availability of label information, feature selection algorithms can be classified into supervised~\cite{song2007supervised, sotoca2010supervised, thoma2009near}, semi-supervised~\cite{zhao2007semi, xu2010discriminative, kong2010semi}, and unsupervised~\cite{he2005laplacian, zhao2007spectral, du2015unsupervised, zhou2016global, cai2010unsupervised, zhao2010efficient, gu2011joint, yang2011l2, hou2011feature, hou2014joint, li2014clustering, li2015robust, zhu2015unsupervised, zhu2017non, hu2017graph, shang2016self, zhu2017subspace} methods. Since labeled data are usually expensive and time-consuming to acquire~\cite{tang2009inferring, li2010image}, unsupervised feature selection has been gaining more and more attention recently and is the subject of our focus in this work. 
\item{\textbf{Search Strategy.}} {{In terms of selection strategies, feature selection methods can be categorized into wrapper, filter, and embedded methods. Wrapper methods~\cite{guyon2003introduction, kohavi1997wrappers} are seldom used in practice since they rely on a repetition of feature subset searching and selected feature subset evaluation until some stopping criteria or some desired performance are reached, which requires an exponential search space and thus is computationally prohibitive when feature dimensionality is high. Filter feature selection methods, e.g. Laplacian score~\cite{he2005laplacian} and SPEC~\cite{zhao2007spectral}, assign a score (measuring task relevance, redundancy, etc.) to each feature and select those with the best scores. Though convenient to computation, these methods are often tailored specifically for a given task and may not provide an appropriate match to the specific application of interest \cite{li2016feature}.}} Embedded methods combine feature selection and model learning and provide a compromise between the two earlier extremes, as they are more efficient than wrapper methods and more task-specific than filter methods. In this paper, we focus on embedded feature selection methods.
\end{itemize} \par
In recent years, feature selection algorithms aiming at selecting features that preserve intrinsic data structure (such as subspace or manifold structure)~\cite{du2015unsupervised, zhou2016global, cai2010unsupervised, zhao2010efficient, gu2011joint, yang2011l2, hou2011feature, hou2014joint, li2014clustering, li2015robust, zhu2015unsupervised, zhu2017non, hu2017graph, shang2016self, zhu2017subspace} have attracted significant attention due to their good performance and interpretability~\cite{li2016feature}. In these methods, data are linearly projected onto new spaces through a transformation matrix, with fitting errors being minimized along with some sparse regularization terms. Feature importance is usually scored using the norms of corresponding rows/columns in the transformation matrix. In some methods~\cite{gu2011joint, yang2011l2, hou2011feature, hou2014joint, li2014clustering, li2015robust, hu2017graph, shang2016self, zhu2017subspace}, the local data structure, which is usually characterized by nearest neighbor graphs, is also preserved in the low-dimensional projection space. A more detailed discussion on this type of methods is in Section \ref{sparse}. One basic assumption of these methods is that the data to be processed lie in or near a completely linear low-dimensional manifold, which is then modeled as a linear subspace.\footnote{People also refer to linear manifold as subspace or linear subspace in the literature. In the sequel, we refer to such a linear manifold or subspace as a subspace for conciseness.} However, this is not always true in practice, in particular with more sophisticated data. \par
In the case when data lies on or close to more generalized or non-linear manifolds, many approaches for dimensionality reduction have been proposed that leverage the data local geometry using neighborhood graphs, such as ISOMAP~\cite{tenenbaum2000global}, Laplacian eigenmaps~\cite{belkin2003laplacian}, locally linear embedding~\cite{roweis2000nonlinear}, etc., but few developments have been reported in feature selection. In this paper, we propose a novel algorithm for graph and autoencoder-based feature selection (GAFS). The reason we choose an autoencoder for the underlying manifold learning is because of its broader goal of data reconstruction, which is a good match in spirit for an unsupervised feature selection framework: we expect to be able to infer the entire data vector from just a few of its dimensions. In this method, we integrate three objective functions into a single optimization framework: ($i$) we use a single-layer autoencoder to reconstruct the input data; ($ii$) we use an $\ell_{2,1}$-norm penalty on the columns of the weight matrix connecting the autoencoder's input layer and hidden layer to provide feature selection; {and} ($iii$) we preserve the local geometric structure of the data through to the corresponding hidden layer activations. To the best of our knowledge, we are the first to combine unsupervised feature selection with an autoencoder design and the preservation of local data structure. Extensive experiments are conducted on image data, audio data, text data, and biological data. Many experimental results are provided to demonstrate the outstanding performance achieved by the proposed method compared with other state-of-the-art unsupervised feature selection algorithms. \par
The key contributions of this paper are highlighted as follows.
\begin{itemize}
\item We propose a novel unsupervised feature selection framework which is based on an autoencoder and graph data regularization. By using this framework, the information of the underlying data subspace can be leveraged, which loosens the assumption of linear manifold in many relevant techniques.
\item We present an efficient solver for the optimization problem underlying the proposed unsupervised feature selection scheme. Our approach relies on an iterative scheme based on the gradient descent of the proposed objective function. 
\item We provide multiple numerical experiments that showcase the advantages of the flexible models used in our feature selection approach with respect to the state-of-the-art approaches from the literature.
\end{itemize}
The rest of this paper is organized as follows. Section \ref{relawork} overviews related work. The proposed framework and the corresponding optimization scheme are presented in Section \ref{framework}. Experimental results and the corresponding analysis are provided in Section \ref{exp}. Section \ref{conclusion} includes conclusion and future work. \par

\section{Related Work}
\label{relawork}
In this section, we provide a review of literature related to our proposed method and introduce the paper's notation standard. Datasets are denoted by $\mathbf{X} = [\mathbf{X}^{(1)}, \mathbf{X}^{(2)}, \cdots , \mathbf{X}^{(n)}] \in \mathbb{R}^{d \times n}$, where $\mathbf{X}^{(i)} \in \mathbb{R}^d$ is the $i$th sample in $\mathbf{X}$ for $i = 1, 2, \cdots , n$, and where $d$ and $n$ denote data dimensionality and number of data points in $\mathbf{X}$, respectively. For a matrix $\mathbf{X}$, $\mathbf{X}^{(q)}$ denotes the $q^{\mathrm{th}}$ column of the matrix, while $\mathbf{X}^{(p,q)}$ denotes the entry of the matrix at the $p^{\mathrm{th}}$ row and $q^{\mathrm{th}}$ column. \par
The $\ell_{r,p}$-norm for a matrix $\mathbf{W} \in \mathbb{R}^{a \times b}$ is denoted as
\begin{equation}
\label{lrpnorm}
\|\mathbf{W}\|_{r,p} = \left( \sum_{j=1}^b \left( \sum_{i=1}^a |\mathbf{W}^{(i,j)}|^r \right)^{p/r} \right)^{1/p}.
\end{equation}
Two common norm choices in optimization are the $\ell_{2,1}$-norm and {the Frobenius norm} (e.g.,  $r=p=2$). Note that unlike most of the literature, our outer sum is performed over the $\ell_r$-norms of the matrix columns instead of its rows; this is done for notation convenience of our subsequent mathematical expressions. \par
{{The trace of a matrix $\textbf{L} \in \mathbb{R}^{a \times a}$ is defined as
\begin{equation}
\label{trace}
{\rm{Tr}} (\mathbf{L}) = \sum_{i=1}^a \mathbf{L}^{(i,i)},
\end{equation}
which is the sum of elements on the main diagonal of $\mathbf{L}$.}} \par
We use $\mathbf{1}$ and $\mathbf{0}$ to denote an all-ones and all-zeros matrix or vector with of the appropriate size, respectively.

\subsection{Sparse Learning-Based Unsupervised Feature Selection}
\label{sparse}
Many unsupervised feature selection methods based on subspace structure preservation have been proposed in the past decades. {{For classes missing labels}}, unsupervised feature selection methods select features that are representative of the underlying subspace structure of the data~\cite{du2015unsupervised}. The basic idea is to use a transformation matrix to project data to a new space and guide feature selection based on the sparsity of the transformation matrix~\cite{zhou2016global}. To be more specific, the generic framework of these methods is based on the optimization 
\begin{equation}
\label{separate}
\min_{\mathbf{W}} \mathcal{L}(\mathbf{Y}, \mathbf{WX}) + \lambda \mathcal{R}(\mathbf{W}),
\end{equation}
where $\mathbf{Y} =  [\mathbf{Y}^{(1)}, \mathbf{Y}^{(2)}, \cdots , \mathbf{Y}^{(n)}] \in \mathbb{R}^{m \times n}$ ($m<d$) is an embedding matrix in which $\mathbf{Y}^{(i)} \in \mathbb{R}^m$ for $i = 1,2,\cdots,n$ denotes the representation of data point $\mathbf{X}^{(i)}$ in the obtained low-dimensional subspace. $\mathcal{L}(\cdot)$ denotes a loss function, and $\mathcal{R}(\cdot)$ denotes a regularization function on the transformation matrix $\mathbf{W} \in \mathbb{R}^{m \times d}$. The methods differ in their choice of embedding $\mathbf{Y}$ and loss and {{regularization}} functions; some examples are presented below. \par
Multi-cluster feature selection (MCFS)~\cite{cai2010unsupervised} and minimum redundancy spectral feature selection (MRSF)~\cite{zhao2010efficient} are two long-standing and well-known subspace learning-based unsupervised feature selection methods. In MCFS, the embedding $\mathbf{Y} \in \mathbb{R}^{m \times n}$ of each data $\mathbf{X}$ is first learned based on spectral clustering. {To be more concrete, a graph is first constructed on training data. Then spectral clustering is performed on data points using the top eigenvectors of graph Laplacian. We refer readers to \cite{cai2010unsupervised} for more details on this spectral clustering procedure. Details on the graph Laplacian are discussed in Section \ref{obj}.} After that, all data points are regressed to the learned embedding through a transformation matrix $\mathbf{W} \in \mathbb{R}^{m \times d}$. The loss function is set to the Frobenius norm of the linear transformation error and the regularization function is set to the $\ell_{1,1}$ norm of the transformation matrix, which promotes sparsity. {{Thus, MCFS can be formulated mathematically as the following optimization problem
\begin{equation}
\min_{\mathbf{W}} \| \mathbf{Y} - \mathbf{W}\mathbf{X} \|_F^2 + \lambda \| \mathbf{W} \|_{1,1}.
\end{equation}
A score for each feature is measured by the maximum absolute value of the corresponding column of the transformation matrix:
\begin{equation}
MCFS(p) = \max_{q = 1,2,\cdots,m} | \mathbf{W}^{(q,p)} |,
\end{equation} 
where $p = 1,2,\cdots,d$. This score is then used in a filter-based feature selection scheme.}}  MRSF is an extension of MCFS that changes the regularization function to an $\ell_{2,1}$-norm that {enforces} column sparsity on the transformation matrix. Ideally, the selected features should be representative enough to keep the loss value close to that obtained when using all features. In order to achieve feature selection, we expect that $\mathbf{W}$ holds a sparsity property with its columns, which means only a subset of the columns are nonzeros. We use the $\ell_2$-norm of a $\mathbf{W}$ column to measure the importance of the corresponding feature, leading to an $\ell_{2,1}$-norm regularization function. Furthermore, MRSF ranks the importance of each feature according to the $\ell_2$-norm of the corresponding column of the transformation matrix. {Compared with MCFS, the use of $\ell_{2,1}$-norm in MRSF can provide the learned subspace with consistent column sparseness. MRSF can be formulated as
\begin{equation}
\min_{\mathbf{W}} \| \mathbf{Y} - \mathbf{WX} \|_F^2 + \lambda \| \mathbf{W} \|_{2,1},
\end{equation}
where $\| \mathbf{W} \|_{2,1}$ is the $\ell_{2,1}-$norm of $\mathbf{W}$. A score is assigned to each feature based on the following scheme
\begin{equation}
MRSF(p) = \| \mathbf{W}^{(p)} \|_2
\end{equation}}

Both MCFS and MRSF are able to select features that provide a suitable subspace approximation to the spectral clustering embedding that detects cluster structure. However, the performance of these two methods is often degraded by the separate nature of subspace learning and feature selection~\cite{shang2016self}. In order to address this problem, many approaches on joint subspace learning and feature selection have been proposed. For example, Gu et. al.~\cite{gu2011joint} proposed a joint framework that combines subspace learning and feature selection. In this framework, data are linearly projected to a low-dimensional subspace with a transformation matrix, and the local data structure captured by a nearest neighbor graph is preserved in data embeddings on low-dimensional subspace. {Meanwhile}, an $\ell_{2,1}-$norm {penalty is applied to the} transformation matrix to guide feature selection simultaneously. That is, subspace learning and feature selection are not two separate steps but combined into a single framework.
Studies like~\cite{yang2011l2, hou2011feature, hou2014joint, li2014clustering, li2015robust} made further modifications to~\cite{gu2011joint}: besides combining subspace learning and feature selection into a single framework, these methods also exploit the discriminative information of the data for unsupervised feature selection. {{For example, in unsupervised discriminative feature selection (UDFS)~\cite{yang2011l2}, data instances are assumed to come from $c$ classes. Furthermore, a linear classifier $\mathbf{W} \in \mathbb{R}^{c \times d}$ is assumed to project data $\mathbf{X}$ onto a $c$-dimensional subspace which captures the discriminative information of data, which can be written as $\mathbf{G} = \mathbf{WX}$, where $\mathbf{G} \in \mathbb{R}^{c \times n}$ denotes the data representation of $\mathbf{X}$ on low-dimensional subspace. Furthermore, for each data point $\mathbf{X}^{(i)}$, a local set $\mathcal{N}_k (\mathbf{X}^{(i)})$ is constructed, which contains $\mathbf{X}^{(i)}$ and its $k$ nearest neighbors $\mathbf{X}_1^{(i)},\mathbf{X}_2^{(i)},\cdots,\mathbf{X}_k^{(i)}$. Denoting $\mathbf{X}_i = [\mathbf{X}^{(i)},\mathbf{X}_1^{(i)},\mathbf{X}_2^{(i)},\cdots,\mathbf{X}_k^{(i)}] \in \mathbb{R}^{d \times (k+1)}$ as the local data matrix containing $\mathbf{X}^{(i)}$ and its $k$ nearest neighbors, we define the local total scatter matrix $\mathbf{S}_{ti} = \mathbf{\tilde{X}}_i \mathbf{\tilde{X}}_i^T \in \mathbb{R}^{d \times d}$ and interclass scatter matrix $\mathbf{S}_{bi} = \mathbf{\tilde{X}}_i \mathbf{G}_i^T \mathbf{G}_i \mathbf{\tilde{X}}_i^T \in \mathbb{R}^{d \times d}$, where $\mathbf{G}_i = [\mathbf{G}^{(i)},\mathbf{G}_1^{(i)},\mathbf{G}_2^{(i)},\cdots,\mathbf{G}_k^{(i)}] \in \mathbb{R}^{c \times (k+1)}$ and $\mathbf{\tilde{X}}_i = \mathbf{X}_i\mathbf{H}_{k+1}$. To be more specific, $\mathbf{G}^{(i)} = \mathbf{WX}^{(i)}$ and $\mathbf{G}_j^{(i)} = \mathbf{WX}_j^{(i)}$ for $j = 1,2,\cdots,k$, and $\mathbf{\tilde{X}}_i$ is a centered version of $\mathbf{X}_i$ and $\mathbf{H}_{k+1} = \mathbf{I}_{k+1} - \frac{1}{k+1}\mathbf{1}\mathbf{1}^T \in \mathbb{R}^{(k+1) \times (k+1)}$. We can also define a selection matrix $\mathbf{P}_i \in \{ 0,1 \}^{n \times (k+1)}$ so that $\mathbf{G}_i = \mathbf{GP}_i$. The local discriminative score $DS_i$ for $\mathbf{X}^{(i)}$ is
$DS_i = {\rm{Tr}} \left[ \left( \mathbf{S}_{ti} + \lambda \mathbf{I}_d \right)^{-1} \mathbf{S}_{bi} \right]  = {\rm{Tr}} \left[ \mathbf{WXP}_i \mathbf{\tilde{X}_i}^T \left( \mathbf{\tilde{X}}_i \mathbf{\tilde{X}}_i^T + \lambda \mathbf{I}_d \right)^{-1} \mathbf{\tilde{X}}_i \mathbf{P}_i^T \mathbf{X}^T \mathbf{W}^T \right]$,
where $\lambda$ is a parameter to make the term $\left( \mathbf{\tilde{X}} \mathbf{\tilde{X}}^T + \lambda \mathbf{I}_d \right)$ invertible. A larger $DS_i$ value means a higher discriminative capability $\mathbf{W}$ has with respect to $\mathbf{X}^{(i)}$. The objective of UDFS is to train a $\mathbf{W}$ corresponding to the highest discriminative scores for data $\mathbf{X}$. Therefore the following objective function is optimized
\begin{equation}
\begin{aligned}
\mathbf{W^*} = & {\argmin}_{\mathbf{WW}^T = \mathbf{I}} \sum_{i=1}^n \left\{ {\rm{Tr}} \left[ \mathbf{G}_i \mathbf{H}_{k+1} \mathbf{G}_i^T \right] - DS_i \right\} + \gamma \| \mathbf{W} \|_{2,1} \\
= & {\argmin}_{\mathbf{WW}^T = \mathbf{I}} {\rm{Tr}} (\mathbf{WMW}^T) + \gamma \| \mathbf{W} \|_{2,1},
\end{aligned}
\end{equation}
where $\mathbf{M} = \mathbf{X} \left[ \sum_{i=1}^n \left( \mathbf{P}_i \mathbf{H}_{k+1} \left( \mathbf{\tilde{X}}_i^T \mathbf{\tilde{X}}_i + \lambda \mathbf{I}_{k+1} \right)^{-1} \mathbf{H}_{k+1} \mathbf{P}_i^T \right) \right] \mathbf{X}^T $ and $\gamma$ is a balancing parameter.}} The orthogonal constraint is to avoid both arbitrary scaling and the trivial solutions of all zeros. We refer readers to~\cite{yang2011l2} for more details on UDFS. 
Though unsupervised, one drawback of these discriminative exploitation feature selection methods is that the feature selection performance relies on an accurate estimation of {the} number of classes.  \par
Instead of projecting data onto a low-dimensional subspace, some approaches consider combining unsupervised feature selection methods with self-representation. In these methods, each feature is assumed to be representable as a linear combination of all (other) features, i.e., $\mathbf{X} = \mathbf{WX} + \mathbf{E}$, where $\mathbf{W} \in \mathbb{R}^{d \times d}$ is a representation matrix and $\mathbf{E} \in \mathbb{R}^{d \times n}$ denotes a reconstruction error. 
That is, the data are linearly projected into the same data space so that the relationships between features can be gleaned from the transformation matrix. This type of method can be regarded as a special case of subspace learning-based feature selection methods where the embedding subspace is equal to the original space. Zhu et. al.~\cite{zhu2015unsupervised} proposed a regularized self-representation (RSR) model for unsupervised feature selection that sets both the loss function and the regularization function to $\ell_{2,1}$-norms on the representation error $\mathbf{E}$ (for robustness to outlier samples) and transformation matrix $\mathbf{W}$ (for feature selection), respectively. RSR can therefore be written as
\begin{equation}
\min_{\mathbf{W}} \| \mathbf{X} - \mathbf{WX} \|_{2,1} + \lambda \| \mathbf{W} \|_{2,1}.
\end{equation}
RSR has been extended to non-convex RSR~\cite{zhu2017non}, where the regularization function is instead set to an $\ell_{2,p}$-norm for $0<p<1$. Unsupervised graph self-representation sparse feature selection (GSR\_SFS)~\cite{hu2017graph} further extends~\cite{zhu2017non} by changing the loss function to a Frobenius norm, as well as by considering local data structure preservation on embedding $\mathbf{WX}$ through spectral graph analysis. GSR\_SFS can be written in the following formulation
\begin{equation}
\min_{\mathbf{W}} \frac{1}{2} ||\mathbf{X}-\mathbf{WX}||_F^2 + {\lambda}_1 \rm{Tr} ({\mathbf{X}}^T{\mathbf{W}}^T\mathbf{LWX}) + {\lambda}_2 ||\mathbf{W}||_{2,1},
\end{equation}
where $\mathbf{L}$ is the graph Laplacian matrix, which will be elaborated in Section \ref{obj}. Self-representation based dual-graph regularized feature selection clustering (DFSC)~\cite{shang2016self} considers the error of self-representation for both the columns and the rows of $\mathbf{X}$ (i.e., both for features and data samples). Moreover, spectral graph analysis on both domains is considered. Subspace clustering guided unsupervised feature selection (SCUFS)~\cite{zhu2017subspace} combines both self-representation and subspace clustering with unsupervised feature selection. In addition, SCUFS also exploits discriminative information for feature selection.

\subsection{Single-Layer Autoencoder}
\label{single-layer autoencoder}
A single-layer autoencoder is an artificial neural network that aims to learn a function $h(\mathbf{x};\mathbf{\Theta}) \approx \mathbf{x}$ with a single hidden layer, where $\mathbf{x} \in \mathbb{R}^d$ is the input data, $h(\cdot)$ is a nonlinear function, and $\mathbf{\Theta}$ is a set of parameters. To be more specific, the workflow of an autoencoder contains two steps:
\begin{itemize}
\item{Encoding:} mapping the input data $\mathbf{x}$ to a compressed data representation $\mathbf{y} \in \mathbb{R}^m$:
\begin{equation}
\mathbf{y} = \sigma ( \mathbf{W_1x} + \mathbf{b}_1 ),
\end{equation}
where $\mathbf{W}_1 \in \mathbb{R}^{m \times d}$ is a weight matrix, $\mathbf{b}_1 \in \mathbb{R}^m$ is a bias vector, and $\sigma (\cdot)$ is an elementary nonlinear activation function. Commonly used activation functions include the sigmoid function, the hyperbolic tangent function, the rectified linear unit, etc. 
\item{Decoding:} mapping the compressed data representation $\mathbf{y}$ to a vector in the original data space $\mathbf{\bar{\mathbf{X}} \in \mathbb{R}^d}$:
\begin{equation}
\mathbf{\bar{\mathbf{X}}} = \sigma ( \mathbf{W_2y} + \mathbf{b}_2 ),
\end{equation}
where $\mathbf{W}_2 \in \mathbb{R}^{d \times m}$ and $\mathbf{b}_2 \in \mathbb{R}^d$ are the corresponding weight matrix and bias vector, respectively.
\end{itemize}
The optimization problem brought by the autoencoder is to minimize the difference between the input data and the reconstructed/output data. To be more specific, given a set of data $\mathbf{X} = [\mathbf{X}^{(1)}, \mathbf{X}^{(2)}, \cdots, \mathbf{X}^{(n)} ]$, the parameters $\mathbf{W}_1$, $\mathbf{W}_2$, $\mathbf{b}_1$, and $\mathbf{b}_2$ are adapted to minimize the reconstruction error $\sum_{i=1}^n \| \mathbf{X}^{(i)} - \mathbf{\bar{X}}^{(i)} \|_2^2$, where $\mathbf{\bar{X}}^{(i)}$ is the output of autoencoder to the input $\mathbf{X}^{(i)}$. The general approach to minimize the reconstruction error is by selecting the parameter values via the backpropagation algorithm{~\cite{schmidhuber2015deep}}. \par
The data reconstruction capability of the autoencoder makes it suitable to capture the essential information of the data while discarding information that is not useful or redundant. Therefore, it is natural to assume that the compressed representation in the hidden layer of a single-layer autoencoder can capture the manifold structure of the input data when such manifold structure exists and is approximated well by the underlying weighting and nonlinearity operations. 
There are many variations of autoencoders, e.g., sparse autoencoder, denoising autoencoder, variational autoencoder, contractive autoencoder, etc. In this paper we only consider the baseline (standard) autoencoder model, which will be elaborated in Section \ref{framework}. We will explore the combination of unsupervised feature selection and other specific variations of autoencoder in future work.

\section{Proposed Method}
\label{framework}

In this section, we introduce our proposed graph autoencoder-based unsupervised feature selection (GAFS). Our proposed framework performs broad data structure preservation 
through a single-layer autoencoder and also preserves local data structure through spectral graph analysis. In contrast to existing methods that exploit discriminative information for unsupervised feature selection by imposing orthogonal constraints on the transformation matrix~\cite{yang2011l2} or low-dimensional data representation~\cite{hou2011feature, hou2014joint}, GAFS does not include such constraints. More specifically, we do not add orthogonal constraints on the transformation matrix because feature weight vectors are not necessarily orthogonal with each other in real-world applications~\cite{qian2013robust}, allowing GAFS to be applicable to a larger set of applications~\cite{zhou2016global}. Furthermore, methods posing orthogonal constraints on low-dimensional data representations {require accurate estimates of the number of classes in order} to obtain reliable label indicators for those algorithms; such estimation is difficult to achieve in an unsupervised framework.  \par

\subsection{Objective Function}
\label{obj}
The objective function of GAFS includes three parts: a term based on a single-layer autoencoder promoting broad data structure preservation; {a regularization term promoting feature selection; and a term based on spectral graph analysis promoting local data structure preservation.}
As mentioned in Section \ref{single-layer autoencoder}, a single-layer autoencoder aims at minimizing the reconstruction error between output and input data by optimizing a reconstruction error-driven loss function:
\begin{equation}
\label{reconErr}
\mathcal{L}(\mathbf{\Theta}) = \frac{1}{2n} \sum_{i=1}^n \| \mathbf{X}^{(i)} - h(\mathbf{X}^{(i)};\mathbf{\Theta}) \|_2^2 = \frac{1}{2n} \| \mathbf{X} - h(\mathbf{X};\mathbf{\Theta}) \|_F^2,
\end{equation}
where $\mathbf{\Theta} = [\mathbf{W}_1, \mathbf{W}_2, \mathbf{b}_1, \mathbf{b}_2]$, {$h(\mathbf{X^{(i)}};\mathbf{\Theta}) = \sigma \left( \mathbf{W}_2 \cdot \sigma ( \mathbf{W_1X^{(i)}} + \mathbf{b}_1 )  + \mathbf{b}_2 \right)$. We} use the sigmoid function as the activation function: $\sigma(z) = 1/(1+ {\rm{exp}} (-z))$. \par
Since $\mathbf{W}_1$ is a weight matrix applied directly on the input data, each column of $\mathbf{W}_1$ can be used to measure the importance of the corresponding data feature. Therefore, $\mathcal{R}(\Theta) = \|\mathbf{W}_1\|_{2,1}$ can be used as a regularization function to promote feature selection as detailed in Section~\ref{sparse}. 
The objective function for the single-layer autoencoder based unsupervised feature selection can be obtained by combining this regularization function with the loss function of \eqref{reconErr}, providing us with the optimization
\begin{equation}
\label{aeFeatSelect}
\min_{\mathbf{\Theta}} \frac{1}{2n} \| \mathbf{X} - h(\mathbf{X};\mathbf{\Theta}) \|_F^2 + \lambda \| \mathbf{W}_1 \|_{2,1},
\end{equation}
where $\lambda$ is a balance parameter.\par
Local geometric structures of the data often contain discriminative information of neighboring data point pairs~\cite{cai2010unsupervised}. They assume that nearby data points should have similar representations. It is often more efficient to combine both broad and local data information during low-dimensional subspace learning~\cite{bottou1992local}. 
In order to characterize the local data structure, we construct a $k$-nearest neighbor ($k$NN) graph $\mathbb{G}$ on the data space. The edge weight between two connected data points is determined by the similarity between those two points. In this paper, we choose cosine distance as similarity measurement {due to} its simplicity. Therefore the adjacency matrix $\mathbf{A}$ for the graph $\mathbb{G}$ is defined as
\begin{equation}
\label{knn_graph}
\mathbf{A}^{(i,j)} = \left\{
\begin{array}{ll}
\dfrac{{\mathbf{X}^{(i)}}^T\mathbf{X}^{(j)}}{\|\mathbf{X}^{(i)}\|_2\|\mathbf{X}^{(j)}\|_2} & \textrm{if}~ \mathbf{X}^{(i)} \in \mathcal{N}_k(\mathbf{X}^{(j)}) ~\textrm{or}~ \mathbf{X}^{(j)} \in \mathcal{N}_k(\mathbf{X}^{(i)}), \\
0 & \textrm{otherwise,} \\
\end{array}\right.
\end{equation}
where $\mathcal{N}_k(\mathbf{X}^{(i)})$ denotes the $k$-nearest neighborhood set for $\mathbf{X}^{(i)}$, and ${\mathbf{X}^{(i)}}^T$ refers to the transpose of $\mathbf{X}^{(i)}$. The Laplacian matrix $\mathbf{L}$ of the graph $\mathbb{G}$ is defined as $\mathbf{L} = \mathbf{D} - \mathbf{A}$, where $\mathbf{D}$ is a diagonal matrix whose $i^{\mathrm{th}}$ element on the diagonal is defined as $\mathbf{D}^{(i,i)} = \sum_{j=1}^n \mathbf{A}^{(i,j)}$. \par
In order to preserve the local data structure in the learned subspace (i.e., if two data points $\mathbf{X}^{(i)}$ and $\mathbf{X}^{(j)}$ are close in original data space then the corresponding low-dimensional representations $\mathbf{Y}^{(i)}$ and $\mathbf{Y}^{(j)}$ are also close in the low-dimensional embedding space), we set up the following minimization objective:
\begin{equation}
\label{localStruct}
\begin{aligned}
\mathcal{G}(\mathbf{\Theta}) &= \frac{1}{2} \sum_{i=1}^n \sum_{j=1}^n \| \mathbf{Y}^{(i)} - \mathbf{Y}^{(j)} \|_2^2 \mathbf{A}^{(i,j)} \\
&=  \frac{1}{2} \sum_{i=1}^n \sum_{j=1}^n ({\mathbf{Y}^{(i)}}^T \mathbf{Y}^{(i)}-{\mathbf{Y}^{(i)}}^T \mathbf{Y}^{(j)}-{\mathbf{Y}^{(j)}}^T \mathbf{Y}^{(i)}+{\mathbf{Y}^{(j)}}^T \mathbf{Y}^{(j)})\mathbf{A}^{(i,j)} \\
&=  \sum_{i=1}^n {\mathbf{Y}^{(i)}}^T\mathbf{Y}^{(i)}\mathbf{D}^{(i,i)} - \sum_{i=1}^n \sum_{j=1}^n {\mathbf{Y}^{(i)}}^T\mathbf{Y}^{(j)}\mathbf{A}^{(i,j)} \\
&=  \rm{Tr} (\mathbf{Y}(\mathbf{\Theta})\mathbf{DY}(\mathbf{\Theta})^T) -  \rm{Tr} (\mathbf{Y}(\mathbf{\Theta})\mathbf{AY}(\mathbf{\Theta})^T) 
=  \rm{Tr} (\mathbf{Y}(\mathbf{\Theta})\mathbf{LY}(\mathbf{\Theta})^T),
\end{aligned}
\end{equation}
where $\rm{Tr}(\cdot)$ denotes the trace operator, $\mathbf{Y}^{(i)}(\mathbf{\Theta}) = \sigma ( \mathbf{W_1x}^{(i)} + \mathbf{b}_1 )$ for $i = 1,2,\cdots,n$ (and we often drop the dependence on $\mathbf{\Theta}$ for readability), and  $\mathbf{Y}(\mathbf{\Theta}) = [\mathbf{Y}^{(1)}(\mathbf{\Theta}), \mathbf{Y}^{(2)}(\mathbf{\Theta}), \cdots, \mathbf{Y}^{(n)}(\mathbf{\Theta}) ]$. \par
Therefore, by combining the single-layer autoencoder based feature selection objective (\ref{aeFeatSelect}) and the local data structure preservation into consideration, the resulting objective function of GAFS can be written in terms of the following minimization with respect to the parameters $\mathbf{\Theta} = [\mathbf{W}_1, \mathbf{W}_2, \mathbf{b}_1, \mathbf{b}_2]$:
\begin{equation}
\begin{aligned}
\label{objFun}
\hat{\mathbf{\Theta}} &= \arg\min_\mathbf{\Theta} \mathcal{F}(\mathbf{\Theta}) = \arg\min_\mathbf{\Theta} \mathcal{L}(\mathbf{\Theta})+\mathcal{R}(\mathbf{\Theta})+\mathcal{G}(\mathbf{\Theta}) \\
&= \arg \min_\mathbf{\Theta} \left[\frac{1}{2n} \| \mathbf{X} - h(\mathbf{X};\mathbf{\Theta}) \|_F^2 + \lambda \| \mathbf{W}_1 \|_{2,1} + \gamma \rm{Tr} (\mathbf{Y(\mathbf{\Theta})}\mathbf{LY(\mathbf{\Theta})}^T)\right],
\end{aligned}
\end{equation}
where $\lambda$ and $\gamma$ are two balance parameters. Filter-based feature selection is then performed using the score function $GAFS(q) = \| \mathbf{W}^{(q)}_1 \|_2$ based on the weight matrix $\mathbf{W}_1$ from $\hat{\mathbf{\Theta}}$. The pseudocode of GAFS is listed in Algorithm \ref{alg_wholeprocedure}. \par
\begin{algorithm}[tbp]
\caption{GAFS Algorithm}
\label{alg_wholeprocedure}
\begin{algorithmic}[1]
\Require High-dimensional dataset $\mathbf{X} = [\mathbf{X}^{(1)}, \mathbf{X}^{(2)}, \cdots, \mathbf{X}^{(n)}] \in \mathbb{R}^{d \times n}$; 
neighborhood size $k$ ; 
hidden layer size $m$; 
balance parameters $\lambda$ and $\gamma$; 
number of features to keep $n_F$.
\Ensure Selected feature index set $\{ r_1, r_2, \cdots r_{n_F} \}$.
\phase{Graph construction}
\State Construct a $k$NN graph $\mathbb{G}$ with adjacency matrix $\mathbf{A}$ described in~\eqref{knn_graph};
\State Calculate the Laplacian matrix $\mathbf{L}$ of the graph $\mathbb{G}$ from the obtained adjacency matrix $\mathbf{A}$;
\phase{Objective optimization}
\State Optimize~\eqref{objFun} by using the scheme described in Section \ref{opt};
\phase{Feature selection}
\State Compute the scores for all features $GAFS(p) = \| \mathbf{W}_1^{(p)} \|_2$ for $p = 1,2,\cdots,d$;
\State Sort these scores and return the indices of the $n_F$ features with largest score values.
\end{algorithmic}  
\end{algorithm} 

\subsection{Optimization}
\label{opt}
The objective function of GAFS shown in \eqref{objFun} does not have a closed-form solution. By following \cite{le2011optimization}, we use a limited memory Broyden-Fletcher-Goldfarb-Shanno (L-BFGS) algorithm to do the optimization. {{Compared with the predominant stochastic gradient descent methods used in neural network training, the L-BFGS algorithm can provide great simplification in parameter tuning and parallel computation. For example, the dimensionality of {the} parameter $\mathbf{\Theta}$ is the sum of {the} dimensionalities of $\mathbf{W}_1 \in \mathbb{R}^{m \times d}$, $\mathbf{W}_2 \in \mathbb{R}^{d \times m}$, $\mathbf{b}_1 \in \mathbb{R}^m$, and $\mathbf{b}_2 \in \mathbb{R}^d$, which is $2md+d+m$. 
Then compared with conventional BFGS algorithm, which requires the computing and storing of $(2md+d+m) \times (2md+d+m)$ Hessian matrices, the L-BFGS algorithm saves a few vectors\footnote{The number of saved vectors is a parameter that can be adjusted.} that represent the approximations implicitly. Therefore, the computational complexity of L-BFGS algorithm are nearly linear in $2md+d+m$, which makes it suitable for optimization problems with large datasets. To be more specific, L-BFGS algorithm save the past $l$ updates of $\mathbf{\Theta}$ and corresponding gradients. Therefore, denoting the number of iterations in the optimization by $t$, the corresponding computational complexity of L-BFGS is $O(tlmd)$.}} We refer readers to~\cite{liu1989limited} for more details on L-BFGS algorithm. In this paper, we implement the L-BFGS algorithm using the \emph{minFunc} toolbox~\cite{schmidt2005minfunc} to solve the GAFS optimization problem. We set number of iterations $t$ to be $400$ and number of storing updates $l$ to be $100$. The solver requires the gradients of the objective function in \eqref{objFun} with respect to its parameters $\mathbf{\Theta}$.  \par
The gradients for the loss term $\mathcal{L} (\mathbf{\Theta})$ can be obtained through a back-propagation algorithm. We {skip} the details for the derivation of the gradients of the error term, which are standard in the formulation of backpropagation for an autoencoder. The resulting gradients are as follows: 
\begin{equation}
\begin{aligned}
\frac{\partial \mathcal{L} (\mathbf{\Theta})}{\partial \mathbf{W}_1} & = \frac{1}{n} \mathbf{\Delta_2}\mathbf{X}^T, \\
\frac{\partial \mathcal{L} (\mathbf{\Theta})}{\partial \mathbf{W}_2} & = \frac{1}{n} \mathbf{\Delta_3}\mathbf{Y}^T, \\
\frac{\partial \mathcal{L} (\mathbf{\Theta})}{\partial \mathbf{b}_1} & = \frac{1}{n} \sum_{i=1}^n \mathbf{\Delta_2}^{(i)} = \frac{1}{n} \mathbf{\Delta_21}, \\
\frac{\partial \mathcal{L} (\mathbf{\Theta})}{\partial \mathbf{b}_2}  &= \frac{1}{n} \sum_{i=1}^n \mathbf{\Delta_3}^{(i)} = \frac{1}{n} \mathbf{\Delta_31}.
\end{aligned}
\end{equation}
Each column $\mathbf{\Delta}_2^{(i)}$ and $\mathbf{\Delta}_3^{(i)}$ of $\mathbf{\Delta_2} \in \mathbb{R}^{m \times n}$ and $\mathbf{\Delta_3} \in \mathbb{R}^{d \times n}$, respectively, contains the error term of the corresponding data point for the hidden layer and the output layer, respectively, 
\begin{equation}
\begin{aligned}
\label{delta}
\mathbf{\Delta}_2^{(q,i)} & = \left( \sum_{p=1}^d \mathbf{W}_2^{(q,p)} \mathbf{\Delta}_3^{(p,i)} \right) \cdot \mathbf{Y}^{(q,i)} \cdot ( \mathbf{1} - \mathbf{Y}^{(q,i)} ),\\
\mathbf{\Delta}_3^{(p,i)} & = ( \bar{\mathbf{X}}^{(p,i)} - \mathbf{X}^{(p,i)} ) \cdot \bar{\mathbf{X}}^{(p,i)} \cdot ( \mathbf{1} - \bar{\mathbf{X}}^{(p,i)} ), 
\end{aligned}
\end{equation}
for $p = 1,2,\cdots,d$, $q = 1,2,\cdots,m$, and $i = 1,2,\cdots,n$, and where $\bar{\mathbf{X}}$ denotes the reconstructed data output of the autoencoder. {We can rewrite \eqref{delta} in matrix form as}
\begin{equation}
\begin{aligned}
\mathbf{\Delta_3} & = ( \mathbf{\bar{X}} - \mathbf{X} ) \bullet \mathbf{\bar{X}} \bullet ( \mathbf{1} - \mathbf{\bar{X}} ), \\
\mathbf{\Delta_2} & = ( \mathbf{W}_2^T\mathbf{\Delta_3} ) \bullet \mathbf{Y} \bullet ( \mathbf{1} - \mathbf{Y} ),
\end{aligned}
\end{equation}
where $\bullet$ denotes the element-wise product operator.  \par
The regularization term $R(\mathbf{\Theta}) = \|\mathbf{W}_1\|_{2,1}$ { and its derivative} do not exist for its $i$th column $\mathbf{W}_1^{(i)}$ when $\mathbf{W}_1^{(i)} = \mathbf{0}$ for $i = 1,2,\cdots,d$. In this case, 
\begin{equation}
\frac{\partial \mathcal{R}(\mathbf{\Theta})}{\partial\mathbf{W}_1} = \mathbf{W_1U},
\end{equation}
where $\mathbf{U} \in \mathbb{R}^{d \times d}$ is a diagonal matrix whose $i$th element on the diagonal is 
\begin{equation}
\mathbf{U}^{(i,i)} = \left\{\begin{array}{ll}
\left(\|\mathbf{W}_1^{(i)}\|_2+\epsilon\right)^{-1}, & \|\mathbf{W}_1^{(i)}\|_2 \neq 0, \\
0, & \mathrm{otherwise.}
\end{array}\right.
\end{equation}
where $\epsilon$ is a small constant added to avoid overflow~\cite{shang2016self}. Since {elements in $\|\mathbf{W}_1\|_{2,1}$ are not differentiable if their values are $0$}, we calculate the subgradient for each element in $\mathbf{W}_1$ in that case. That is, for each element in $\mathbf{W}_1$, the subgradient at {$0$} can be an arbitrary value in the interval $[-1,1]$, and so we set the gradient to $0$ for computational convenience. In summary, the gradients for the regularization term is:
\begin{equation}
\begin{aligned}
\frac{\partial \mathcal{R} (\mathbf{\Theta})}{\partial \mathbf{W}_1} & = \lambda \mathbf{W_1U}, \\
\frac{\partial \mathcal{R} (\mathbf{\Theta})}{\partial \mathbf{W}_2} & = \mathbf{0}, \\
\frac{\partial \mathcal{R} (\mathbf{\Theta})}{\partial \mathbf{b}_1} & = \mathbf{0}, \\
\frac{\partial \mathcal{R} (\mathbf{\Theta})}{\partial \mathbf{b}_2} & = \mathbf{0},
\end{aligned}
\end{equation}
\par
The gradients of the graph term $\mathcal{G}(\mathbf{\Theta}) = \gamma \rm{Tr} (\mathbf{YL}\mathbf{Y}^T)$ can be obtained in a straightforward fashion as follows:
\begin{equation}
\begin{aligned}
\frac{\partial \mathcal{L} (\mathbf{\Theta})}{\partial \mathbf{W}_1}  & = \frac{\partial \rm{Tr} (\gamma\mathbf{YL}\mathbf{Y}^T)}{\partial \mathbf{Y}} \cdot \frac{\partial \mathbf{Y}}{\partial \mathbf{W}_1} = 2\gamma \left( \mathbf{YL} \bullet \mathbf{Y} \bullet ( \mathbf{1} - \mathbf{Y} ) \right) \mathbf{X}^T, \\
\frac{\partial \mathcal{L} (\mathbf{\Theta})}{\partial \mathbf{b}_1}  & = \frac{\partial \rm{Tr} (\gamma\mathbf{YL}\mathbf{Y}^T)}{\partial \mathbf{Y}} \cdot \frac{\partial \mathbf{Y}}{\partial \mathbf{b}_1} = 2\gamma \left( \mathbf{YL} \bullet \mathbf{Y} \bullet ( \mathbf{1} - \mathbf{Y} ) \right) \mathbf{1}, \\
\frac{\partial \mathcal{L} (\mathbf{\Theta})}{\partial \mathbf{W}_2}  & = \mathbf{0}, \\
\frac{\partial \mathcal{L} (\mathbf{\Theta})}{\partial \mathbf{b}_2} & = \mathbf{0}.
\end{aligned}
\end{equation}
To conclude, the gradients of the GAFS objective function with respect to $\mathbf{\Theta} = [\mathbf{W}_1, \mathbf{W}_2, \mathbf{b}_1, \mathbf{b}_2]$ can be written as
\begin{equation}
\begin{aligned}
\frac{\partial \mathcal{F} (\mathbf{\Theta})}{\partial \mathbf{W}_1} & = \frac{1}{n} \mathbf{\Delta_2}\mathbf{X}^T + \lambda \mathbf{W_1U} + 2\gamma \left( \mathbf{YL} \bullet \mathbf{Y} \bullet ( \mathbf{1} - \mathbf{Y} ) \right) \mathbf{X}^T, \\
\frac{\partial \mathcal{F} (\mathbf{\Theta})}{\partial \mathbf{W}_2} & = \frac{1}{n} \mathbf{\Delta_3}\mathbf{Y}^T, \\
\frac{\partial \mathcal{F} (\mathbf{\Theta})}{\partial \mathbf{b}_1} & = \frac{1}{n} \mathbf{\Delta_21} + 2\gamma \left( \mathbf{YL} \bullet \mathbf{Y} \bullet ( \mathbf{1} - \mathbf{Y} ) \right) \mathbf{1}, \\
\frac{\partial \mathcal{F} (\mathbf{\Theta})}{\partial \mathbf{b}_2} & = \frac{1}{n} \mathbf{\Delta_31}
\end{aligned}
\end{equation}

\subsection{Computational Complexity Analysis}
\label{complexity}

{In this subsection, we provide the computational complexity analysis of the proposed GAFS algorithm. \par
The time complexity for calculating the similarity values for a single instance in dataset $\mathbf{X} \in \mathbb{R}^{d \times n}$ is $O(dn)$, where $n$ is the number of instances in the dataset and $d$ is data dimensionality. Therefore, the computational complexity of $k$NN graph construction for the whole dataset is $O(dn^2)$. \par
As mentioned in Section \ref{opt}, the time complexity of using L-BFGS algorithm to optimize ~\eqref{objFun} is $O(tlmd)$, where $t$ is the number of iterations for parameter updating and $l$ is the number of steps stored in memory. If the hidden layer size of the autoencoder we use is $m$, then in each iteration, parameter updating requires an operation of time complexity $O(mdn^2)$, which leads to a time complexity of $O(tmdn^2)$ for $t$ optimization iterations. Therefore, the time complexity of objective function optimization is $O(tmdn^2+tlmd)$. \par
After we obtain $\mathbf{W}_1$, the computation of score for each feature requires a $O(dn)$ operation. After that, we use a quick sort algorithm with time complexity $O(n{\rm{log}}n)$ to sort the obtained scores. \par
Therefore, the overall time complexity of GAFS is $O(tmdn^2+tlmd)$. It is obvious that the time complexity of GAFS largely depends on the objective optimization stage.}

\section{Experiments}
\label{exp}

In this section, we evaluate the feature selection performance of GAFS in terms of both supervised and unsupervised tasks, e.g. clustering and classification, on several benchmark datasets. We also compare GAFS with other state-of-the-art unsupervised feature selection algorithms. To be more specific, we first select $p$ representative features and then perform both clustering and classification on those selected features. The performance of clustering and classification is used as the metric to evaluate feature selection algorithms.

\subsection{Data Description}
\label{data}
We perform experiments on $10$ benchmark datasets,\footnote{{{Caltech-UCSD Birds 200 is downloaded from \url{http://www.vision.caltech.edu/visipedia/CUB-200.html}. \\ Caltech101 is downloaded from \url{http://www.vision.caltech.edu/Image_Datasets/Caltech101/}. \\ All other datasets are downloaded from \url{http://featureselection.asu.edu/datasets.php}.}}} including $5$ image datasets (MNIST, COIL20, Yale, {{Caltech101, CUB200}}), $3$ text datasets (PCMAC, BASEHOCK, RELATHE), $1$ audio dataset (Isolet), and $1$ biological dataset (Prostate\_GE). {{For all datasets except Caltech101 and CUB200, we use the original features for feature selection. For both Caltech101 and CUB200, we do not use pixels as features due to the high dimensionality of each image and the differences between image sizes. Because of the significant success of deep convolutional neural network (CNN) features on computer vision, we also adopt CNN features in our experiments for both Caltech101 and CUB200. To be more specific, we use the Keras tool \cite{chollet2015keras} with the pre-trained VGG-19 model \cite{simonyan2014very}. We use the $4096$-dimensional output of the second fully connected layer as the feature vector.}} \par
In order to eliminate the side effects caused by imbalanced classes, for each dataset we set the number of instances from each class {to be} the same for both {training and testing sets}. For example, when an experiment is conducted on Yale, for each class $6$ instances are used for training and $5$ instances are used for testing. {{The Caltech101 dataset contains both a ``Faces" and ``Faces\_easy" class, with each consisting of different versions of the same human face images. However, the images in ``Faces" contain more complex backgrounds. To avoid confusion between these two similar classes of images, we drop the ``Faces\_easy" class from consideration. Therefore, we keep $100$ classes for Caltech101. For CUB200, we removed $5$ classes with sample sizes smaller than $50$, with $195$ remaining for experiments. Properties of these datasets are summarized in Table \ref{datasets}.}} \par
\begin{table}[tbp]
\centering
\begin{tabular}{llllll}
\hline
Dataset & Features & Training Instances & Testing Instances & Classes & Type \\
\hline
MNIST & 784 & 1000 & 27100 & 10 & Hand Written Digit Image \\
COIL20 & 1024 & 720 & 720 & 20 & Object Image \\
Yale & 1024 & 90 & 75 & 15 & Human Face Image \\
PCMAC & 3289 & 960 & 960 & 2 & Text \\
BASEHOCK & 4862 & 994 & 994 & 2 & Text \\
RELATHE & 4322 & 648 & 648 & 2 & Text \\
Prostate\_GE & 5966 & 50 & 50 & 2 & Biology \\
Isolet & 617 & 780 & 780 & 26 & Audio \\
Caltech101 & 4096 & 2000 & 1000 & 100 & Natural Image \\
CUB200 & 4096 & 7800 & 1950 & 195 & Natural Image \\
\hline
\end{tabular}
\caption{{Details of datasets used in our experiment.}}
\label{datasets}
\end{table}

\subsection{Evaluation Metric}
We perform both supervised (i.e., classification) and unsupervised (i.e., clustering) tasks on datasets formulated by the selected features in order to evaluate the effectiveness of feature selection algorithms. For classification, we employ softmax classifier {due to} its simplicity and compute the classification accuracy as the evaluation metric for feature selection effectiveness. For clustering, we use $k$-means clustering on the selected features and use two different evaluation metrics to evaluate the clustering performance of all methods. The first is clustering accuracy (ACC), defined as 
$$\mathrm{ACC} = \frac{1}{n}\sum_{i=1}^n \delta(g_i,\mathrm{map}(c_i)),$$ 
where $n$ is the total number of data samples, $\delta(a,b) = 1$ when $a=b$ and $0$ when $a \neq b$, $\mathrm{map}(\cdot)$ is the optimal mapping function between cluster labels and class labels obtained using the Hungarian algorithm~\cite{kuhn1955hungarian}, and $c_i$ and $g_i$ are the clustering and ground truth labels of a given data sample $\mathbf{x_i}$, respectively. The second is normalized mutual information (NMI), which is defined as 
$$\mathrm{NMI} = \frac{\mathrm{MI}(C,G)}{\max (H(C),H(G))},$$ 
where $C$ and $G$ are clustering labels and ground truth labels, respectively, $\mathrm{MI}(C,G)$ is the mutual information between $C$ and $G$, and $H(C)$ and $H(G)$ denote the entropy of $C$ and $G$, respectively. More details about NMI are available in~\cite{strehl2002cluster}. For both ACC and NMI, $20$ clustering processes are repeated with random initialization for each case following the setup of~\cite{cai2010unsupervised} and~\cite{yang2011l2}, and we report the corresponding mean values of ACC and NMI. \par


\subsection{Experimental Setup}

{In our experiment}, we compare GAFS with LapScore\footnote{Available at http://www.cad.zju.edu.cn/home/dengcai/Data/code/LaplacianScore.m}~\cite{he2005laplacian}, SPEC\footnote{Available at https://github.com/matrixlover/LSLS/blob/master/fsSpectrum.m}~\cite{zhao2007spectral}, MRSF\footnote{Available at https://sites.google.com/site/alanzhao/Home}~\cite{zhao2010efficient}, UDFS\footnote{Available at http://www.cs.cmu.edu/~yiyang/UDFS.rar}~\cite{yang2011l2}, and RSR\footnote{Available at https://github.com/guangmingboy/githubs\_doc}~\cite{zhu2015unsupervised}. Among these methods, LapScore and SPEC are filter feature selection methods which are based on data similarity. LapScore uses spectral graph analysis to set a score for each feature. SPEC is an extension to LapScore and can be applied to both supervised and unsupervised scenarios {by varying the construction of graph}. Details on MRSF, UDFS, and RSR can be found in Section \ref{sparse}. Details on {the} computational complexity of GAFS and these five methods are listed in Table~\ref{time_complexity}. 
\begin{table}[tbp]
\centering
\begin{tabular}{|l|l|}  
\hline  
Method & Time Complexity \\ 
\hline  
GAFS & $O(tmdn^2+tlmd)$ \\ 
\hline 
LapScore & $O(dn^2)$  \\  
\hline 
SPEC & $O(rn^3+dn^2)$ \\ 
\hline 
MRSF & $O(mn^3+dn^2)$  \\  
\hline 
UDFS & $O(n^2+td^3)$ \\  
\hline 
RSR & $O(td^2n+td^3)$ \\ 
\hline
\end{tabular}  
\caption{{Computational Complexity of GAFS and Five Comparing Methods. In this table, $d$ denotes data dimensionality, $n$ denotes number of samples, $t$ denotes number of iterations for optimization. For SPEC, $r$ is a parameter that controls the use of graph Laplacian matrix $\mathbf{L}$. We refer readers to \cite{zhao2007spectral} for more details on the definition of $r$. For MRSF, $m$ denotes subspace dimensionality.}}  
\label{time_complexity}
\end{table}  
Besides the five methods, we also compare GAFS with the performance of using all features as the baseline. 

Both GAFS and {the} compared algorithms include parameters to adjust. In this experiment, we fix some parameters and tune others according to a ``{grid search}" strategy. For all algorithms, we select $p \in \{2\%$, $4\%$, $6\%$, $8\%$, $10\%$, $20\%$, $30\%$, $40\%$, $50\%$, $60\%$, $70\%$, $80\%\}$ of all features for each dataset. For all graph-based algorithms, the number of nearest neighbor in a $k$NN graph is set to $5$. For all algorithms projecting data onto a low-dimensional space, the space dimensionality is set in the range of $m \in \{10, 20, 30, 40\}$. In GAFS, the range for the hidden layer size is set to match that of the subspace dimensionality $m$,\footnote{We will alternatively use the terminologies {\em subspace dimensionality} and {\em hidden layer size} in descriptions of GAFS.} while the balance parameters are given ranges $\lambda \in \{10^{-4},10^{-3}, 10^{-2}, 10^{-1}, 1\}$ and $\gamma \in \{0, 10^{-4}, 5 \times 10^{-4}, 10^{-3}, 5 \times 10^{-3}\}$, respectively. For UDFS, we use the range $\gamma \in \{10^{-9}, 10^{-6}, 10^{-3}, 1, 10^3, 10^6, 10^9\}$, 
and $\lambda$ is fixed to $10^3$. For RSR, we use the range $\lambda\in \{10^{-3}, 5 \times 10^{-3}, 10^{-2}, 5 \times 10^{-2}, 10^{-1}, 5 \times 10^{-1}, 1, 5, 10, 10^2\}$. \par
For each specific value of $p$ on a certain dataset, we tune the parameters for each algorithm in order to achieve the best results among all possible combinations. For classification, we report the highest classification accuracy. For clustering, we report the highest average values for both ACC and NMI from $20$ repetitions. 

\subsection{Parameter Sensitivity}
\label{parasens}
We study the performance variation of GAFS with respect to the hidden layer size $m$ and the two balance parameters $\lambda$ and $\gamma$. We show the results on all the $8$ datasets in terms of ACC. \par
\begin{figure*}
\begin{minipage}{0.2\linewidth}
  \centerline{\includegraphics[width=4.0cm]{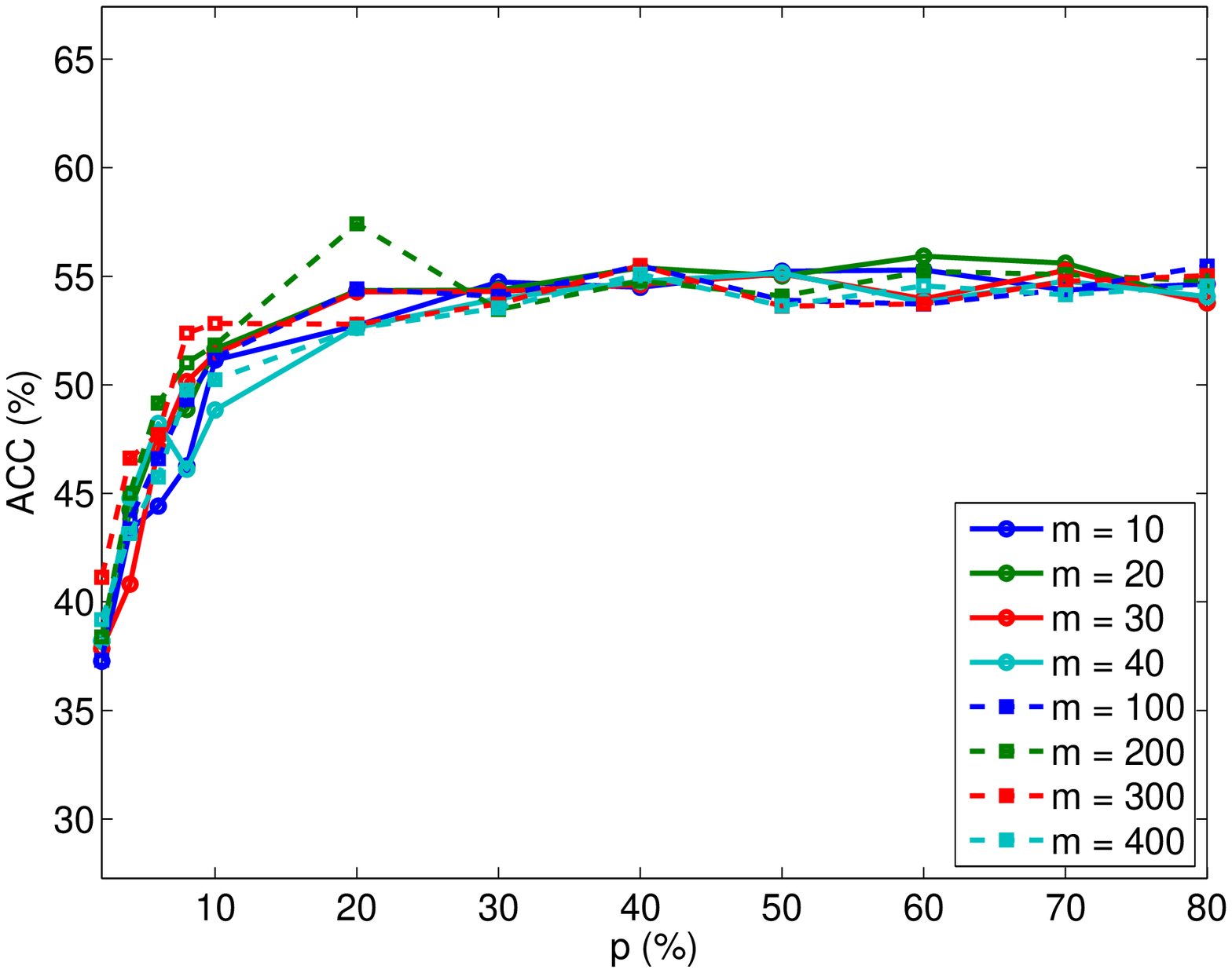}}
  \centerline{(a) MNIST}
\end{minipage}
\hfill
\begin{minipage}{0.2\linewidth}
  \centerline{\includegraphics[width=4.0cm]{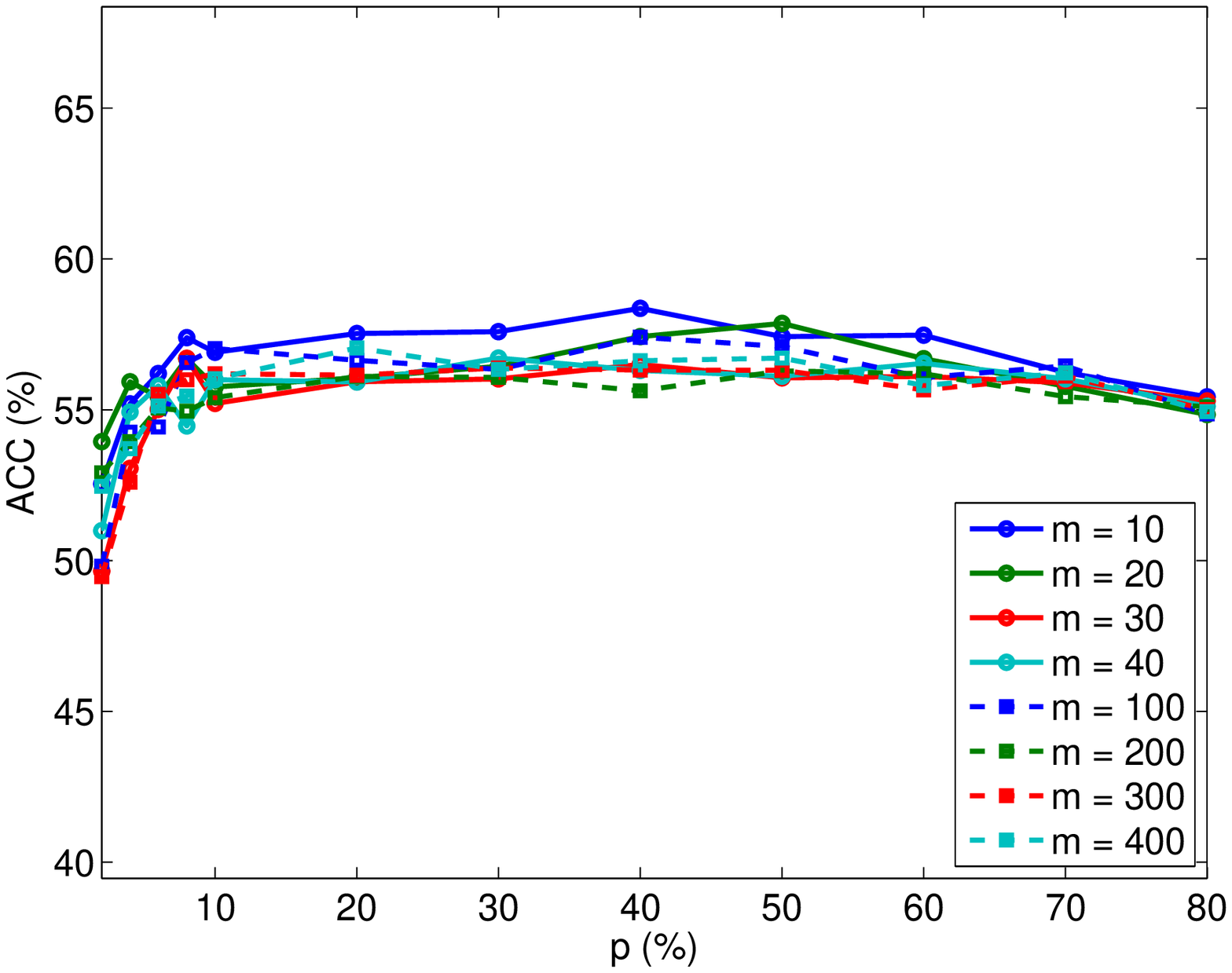}}
  \centerline{(b) COIL20}
\end{minipage}
\hfill
\begin{minipage}{0.2\linewidth}
  \centerline{\includegraphics[width=4.0cm]{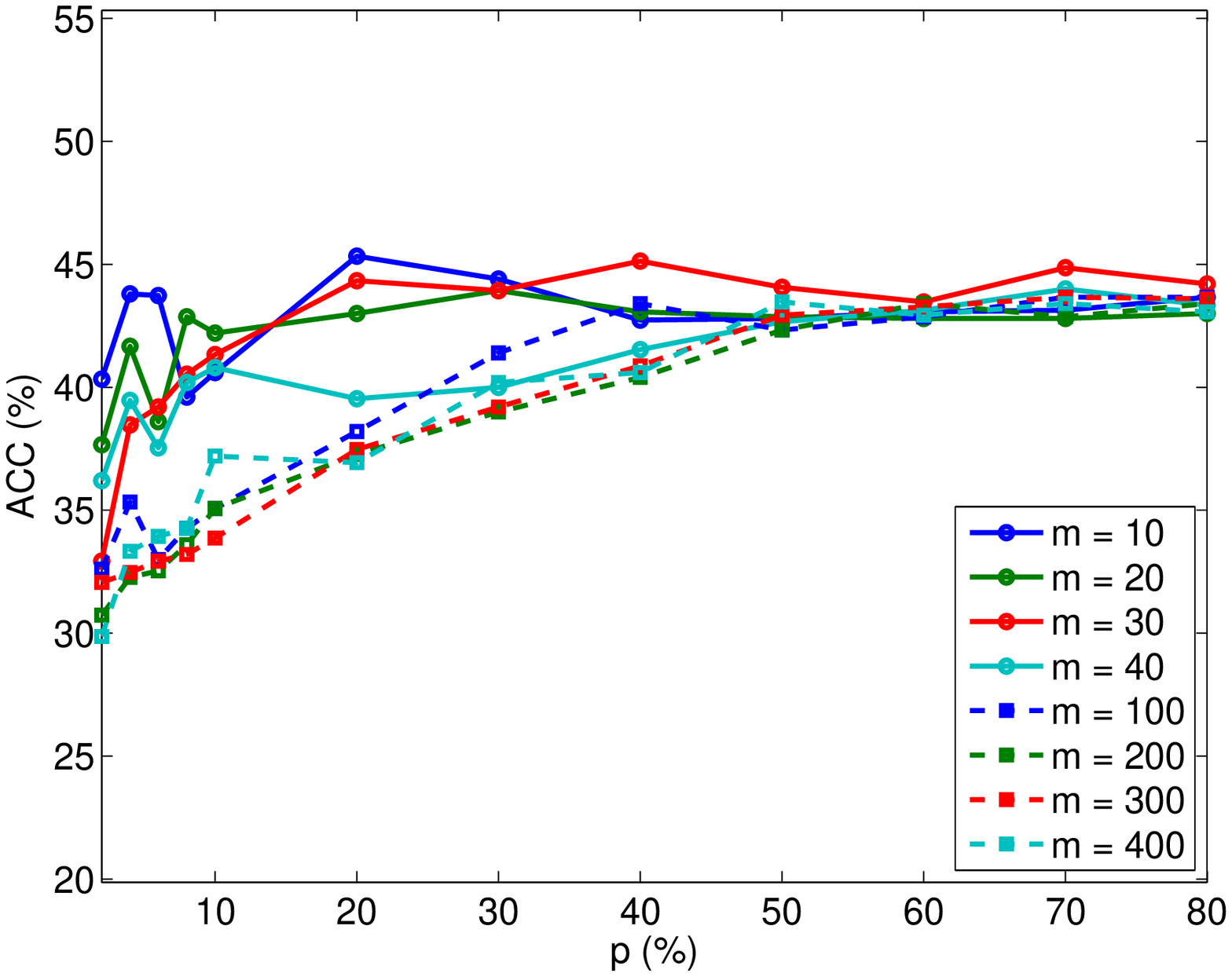}}
  \centerline{(c) Yale}
\end{minipage}
\hfill
\begin{minipage}{0.2\linewidth}
  \centerline{\includegraphics[width=4.0cm]{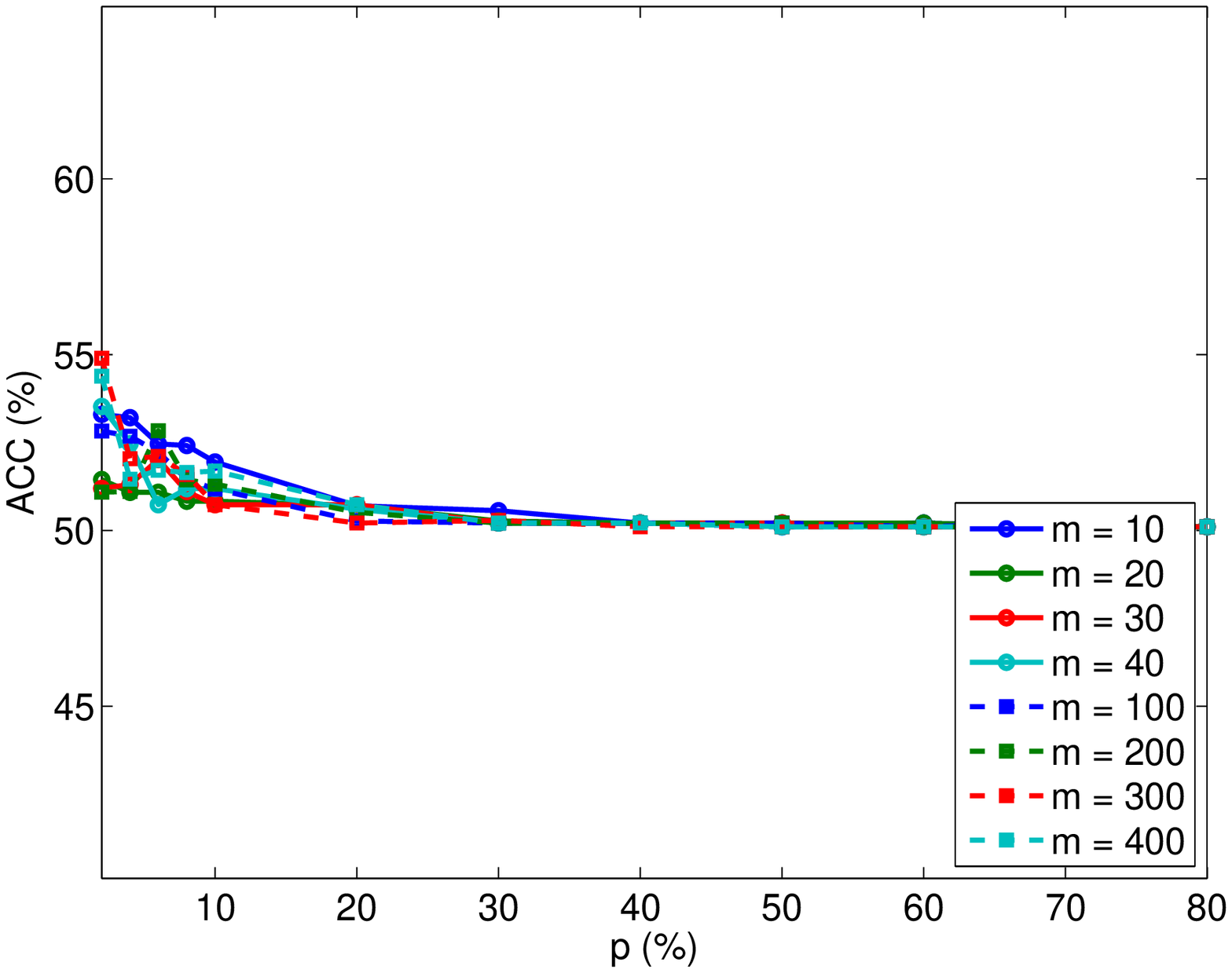}}
  \centerline{(d) PCMAC}
\end{minipage}
\vfill
\begin{minipage}{0.2\linewidth}
  \centerline{\includegraphics[width=4.0cm]{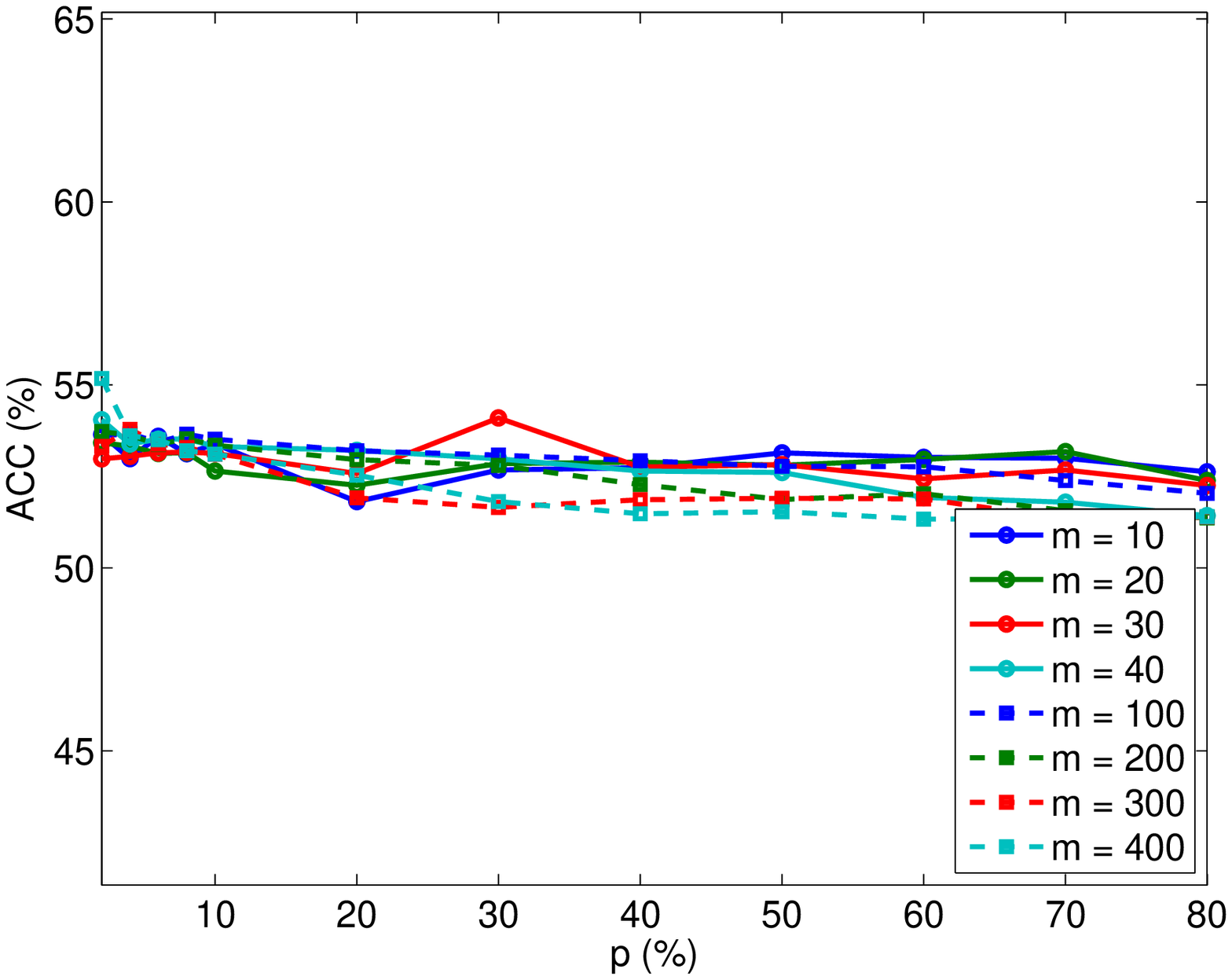}}
  \centerline{(e) BASEHOCK}
\end{minipage}
\hfill
\begin{minipage}{0.2\linewidth}
  \centerline{\includegraphics[width=4.0cm]{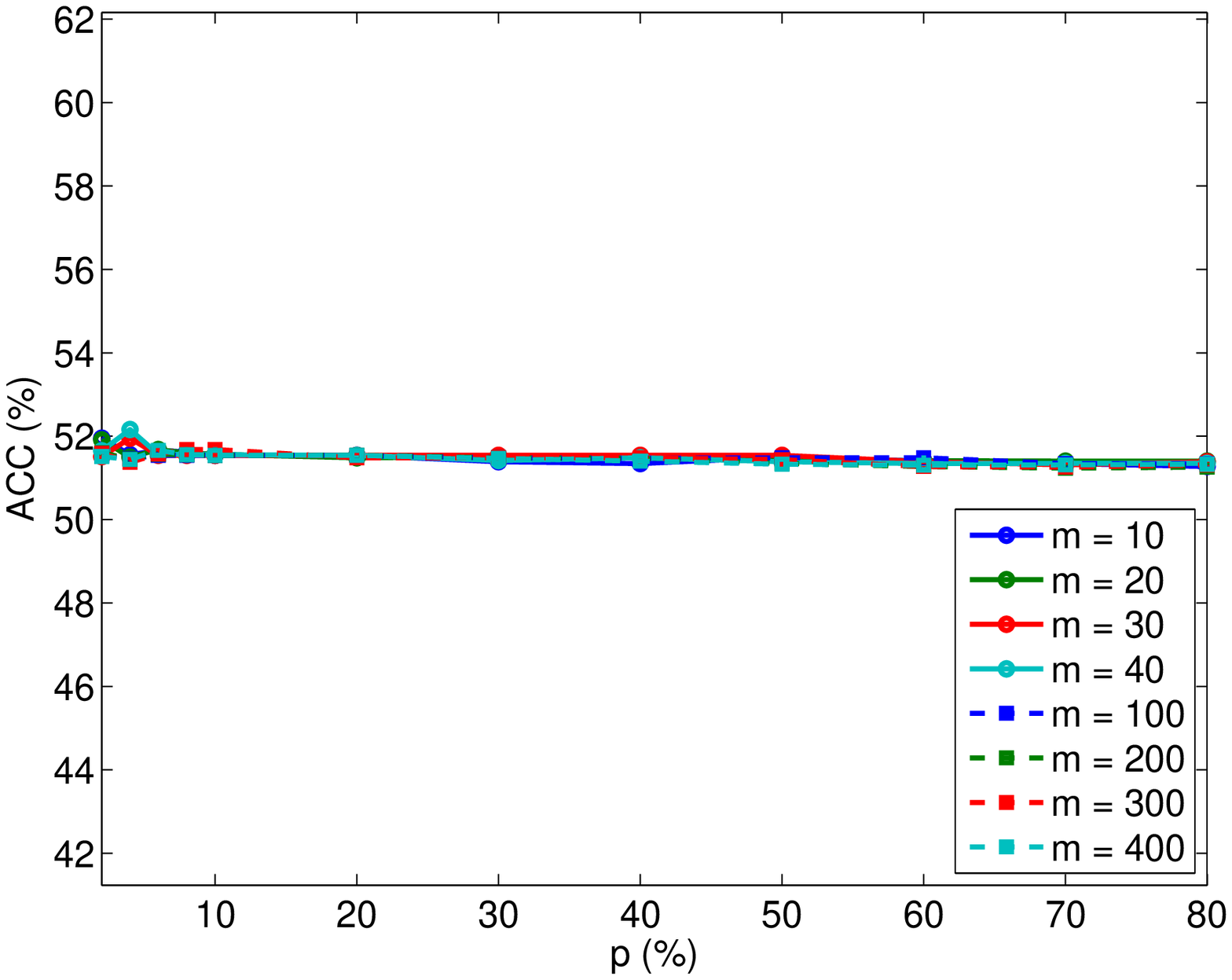}}
  \centerline{(f) RELATHE}
\end{minipage}
\hfill
\begin{minipage}{0.2\linewidth}
  \centerline{\includegraphics[width=4.0cm]{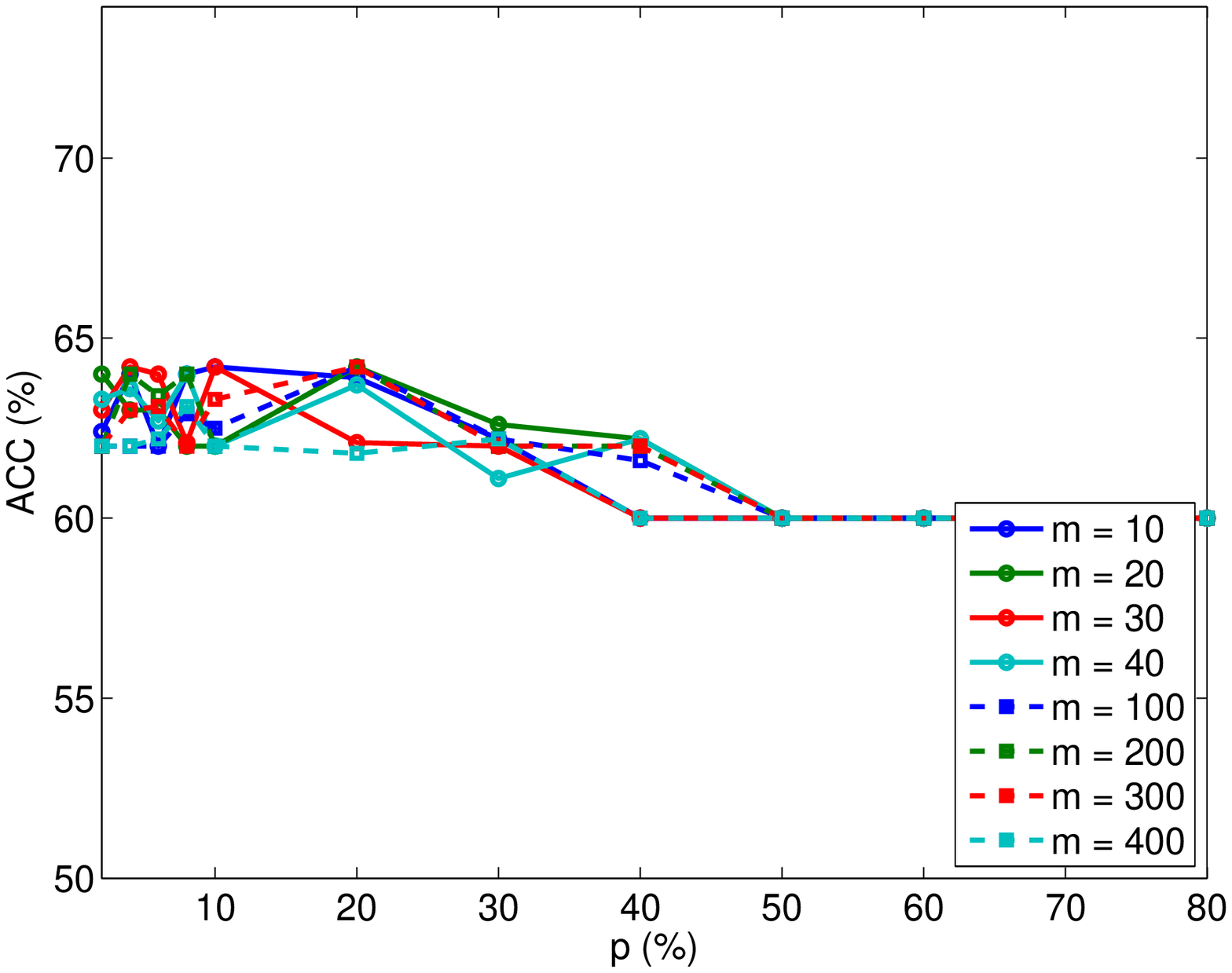}}
  \centerline{(g) Prostate\_GE}
\end{minipage}
\hfill
\begin{minipage}{0.2\linewidth}
  \centerline{\includegraphics[width=4.0cm]{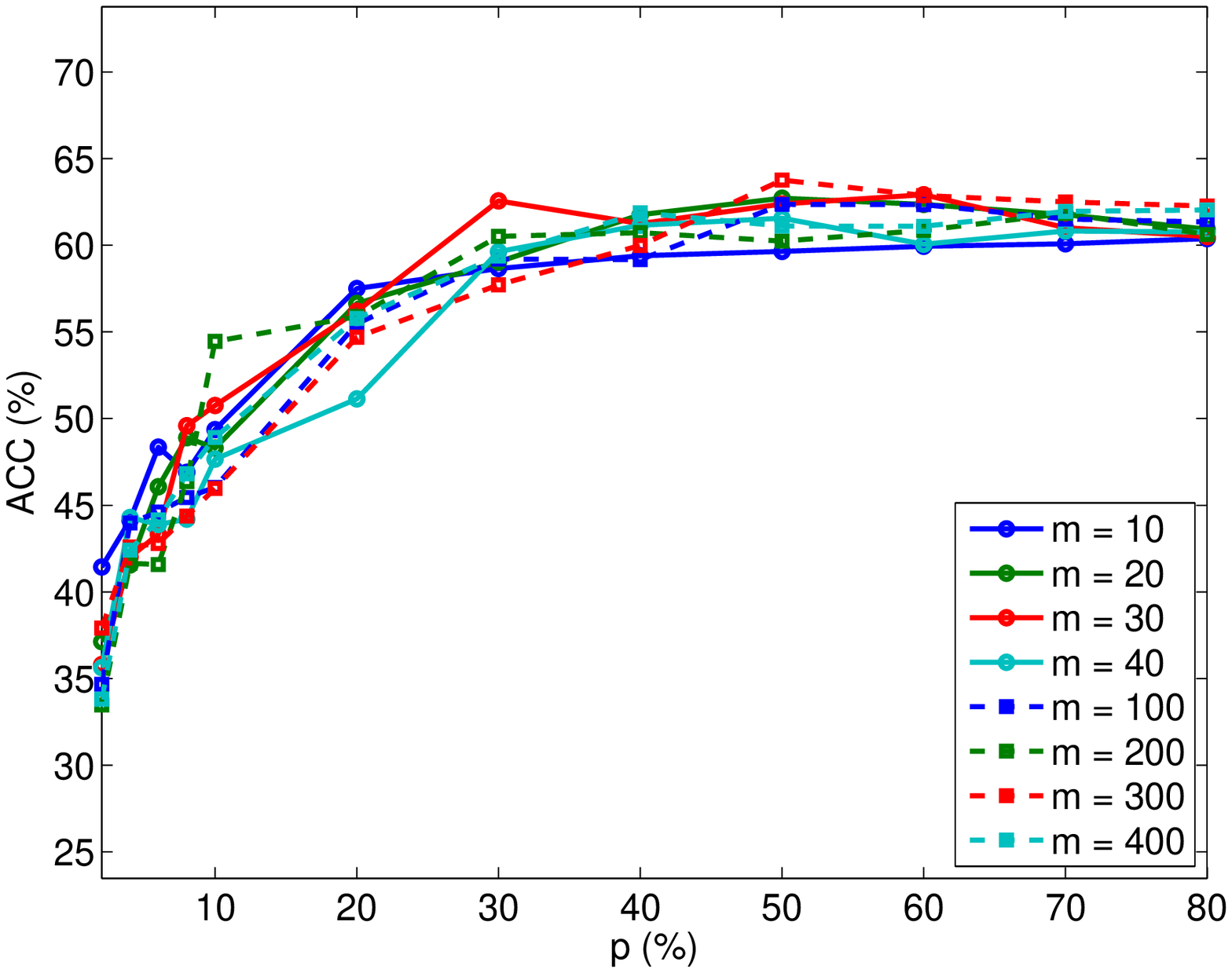}}
  \centerline{(h) Isolet}
\end{minipage}
\vfill
\begin{minipage}{0.2\linewidth}
\end{minipage}
\hfill
\begin{minipage}{0.2\linewidth}
  \centerline{\includegraphics[width=4.0cm]{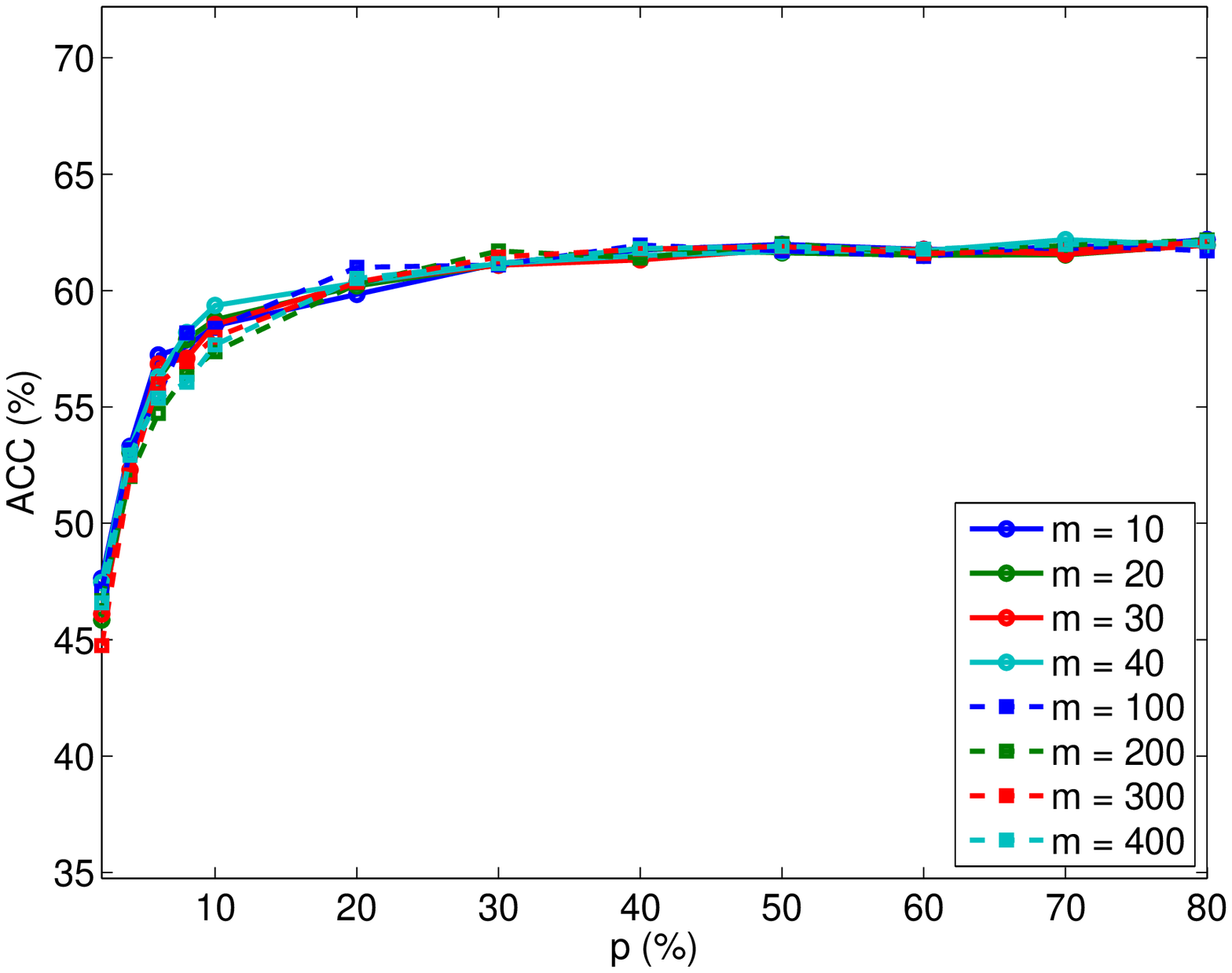}}
  \centerline{(g) Caltech101}
\end{minipage}
\hfill
\begin{minipage}{0.2\linewidth}
  \centerline{\includegraphics[width=4.0cm]{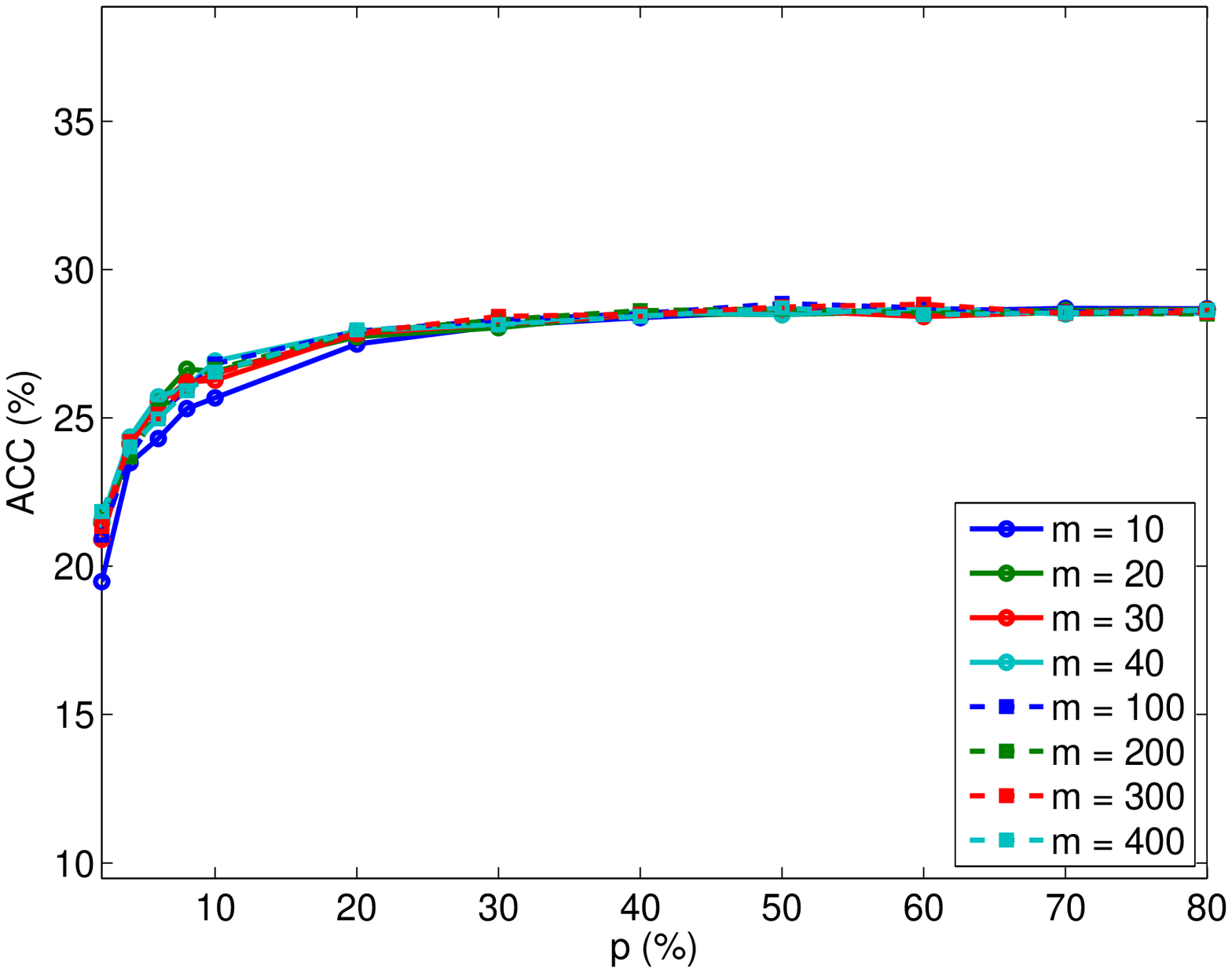}}
  \centerline{(h) CUB200}
\end{minipage}
\hfill
\begin{minipage}{0.2\linewidth}
\end{minipage}
\caption{{Performance of GAFS in clustering as a function of the percentage of features selected $p$ ($\%$) for varying sizes of the autoencoder hidden layer $m$. Clustering accuracy is used as the evaluation metric.}}
\label{sensHL}
\end{figure*}

We first study the parameter sensitivity of GAFS with respect to {the} subspace dimensionality $m$. Besides the aforementioned manifold dimensionality range $m \in \{10, 20, 30, 40\}$, we also conducted experiments with hidden layer size values of $m \in \{100, 200, 300, 400\}$ to investigate the performance change for a larger range of reduced dimensionality values. The results in Fig.~\ref{sensHL} show that the performance of GAFS is not too sensitive to hidden layer size on the given datasets, with the exception of Yale, where the performance with hidden layer size of $m \in \{10, 20, 30, 40\}$ is apparently better than that with reduced dimensionality $m \in \{100, 200, 300, 400\}$, while the performance variations are small in the latter set. One possible reason behind this behavior is that for a human face image dataset like Yale, the differences between data instances can be subtle since they may only lie in a small area of relevance such as eyes, mouth, nose, etc. Therefore, in this case a small subspace dimensionality can be enough for information preservation, while a large subspace dimensionality may introduce redundant information that may harm feature selection performance. \par
\begin{figure*}
\begin{minipage}{0.2\linewidth}
  \centerline{\includegraphics[width=4.0cm]{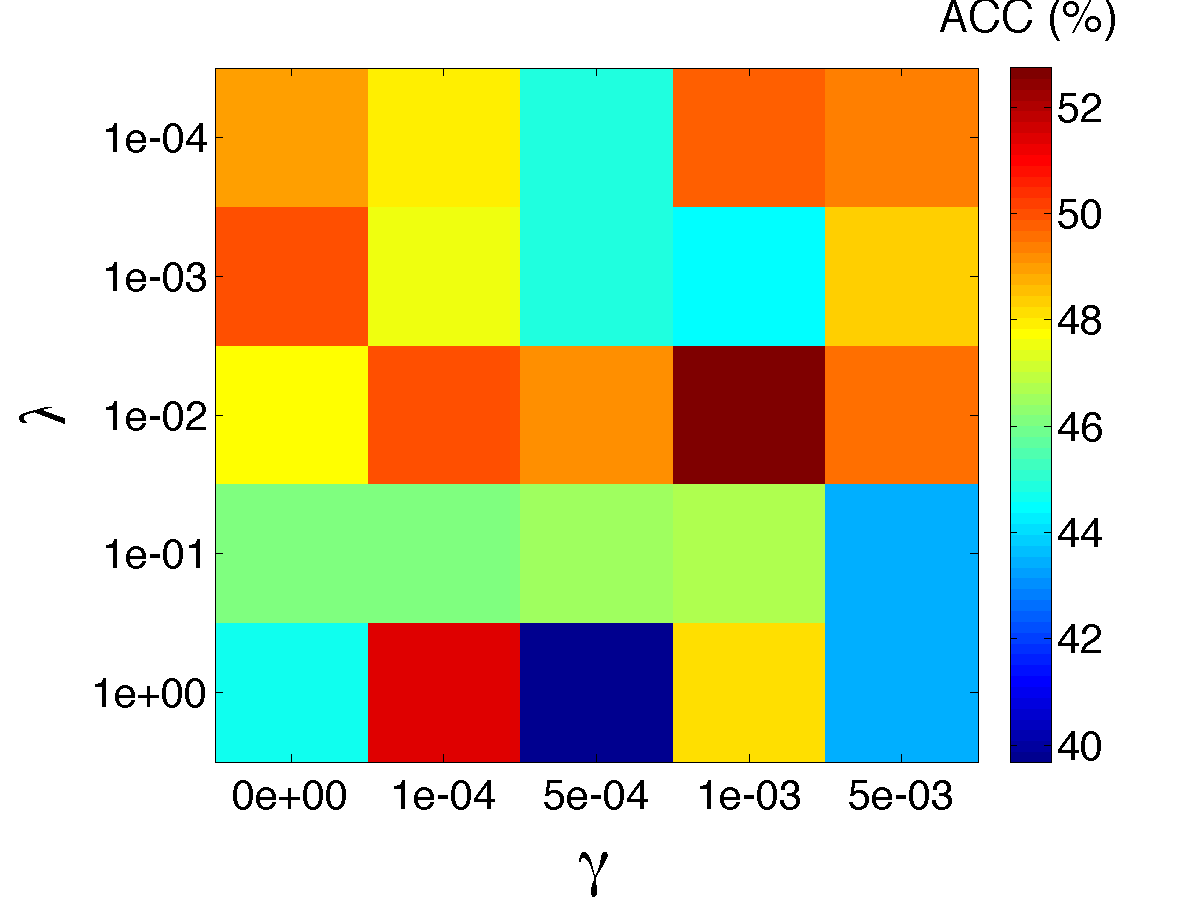}}
  \centerline{(a) MNIST}
\end{minipage}
\hfill
\begin{minipage}{0.2\linewidth}
  \centerline{\includegraphics[width=4.0cm]{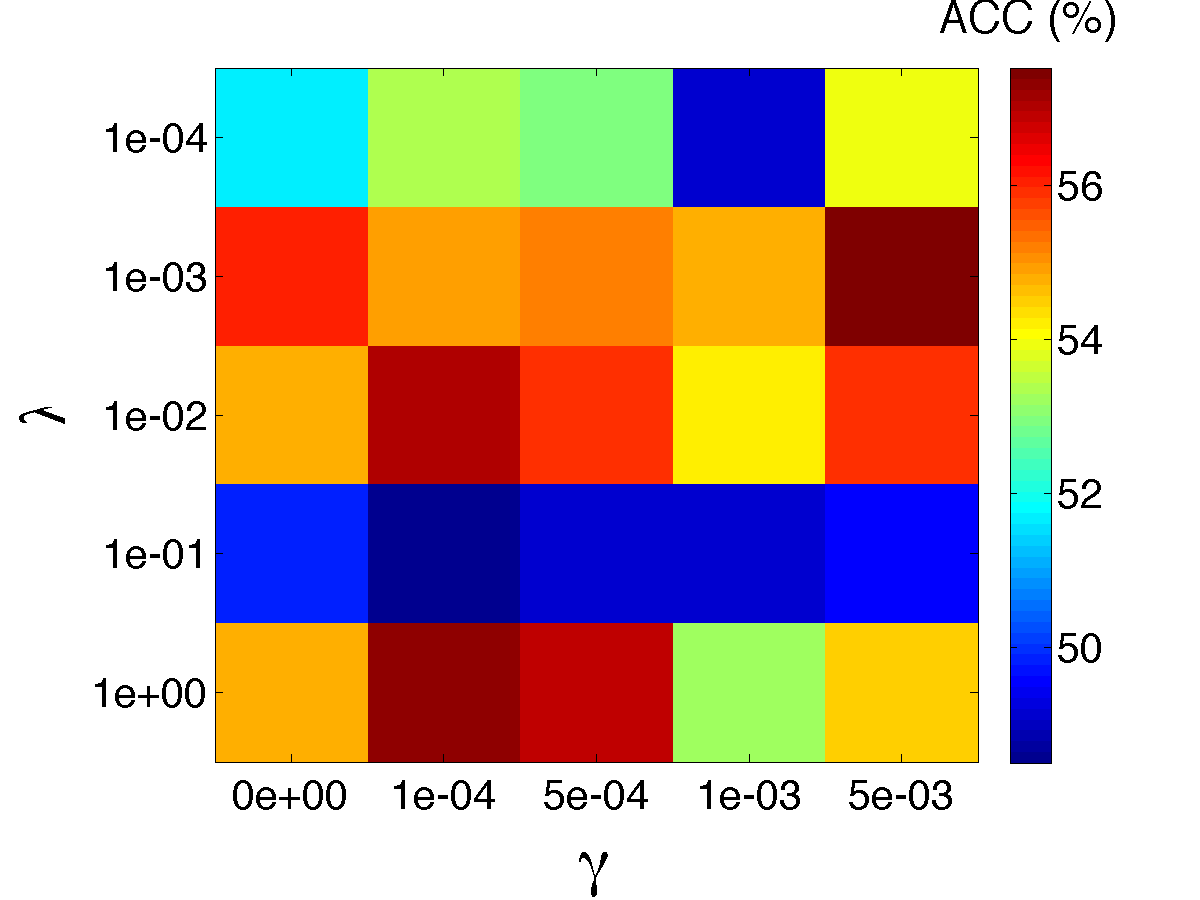}}
  \centerline{(b) COIL20}
\end{minipage}
\hfill
\begin{minipage}{0.2\linewidth}
  \centerline{\includegraphics[width=4.0cm]{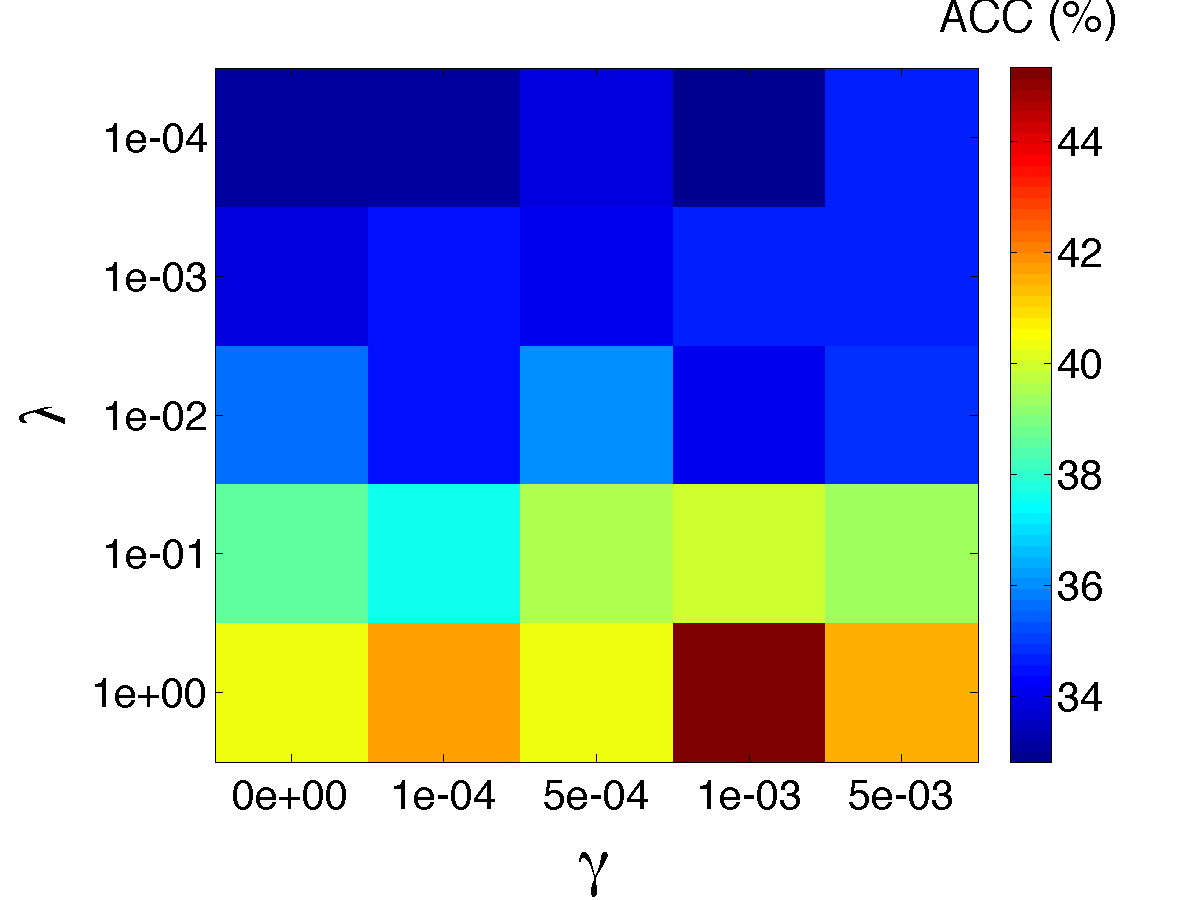}}
  \centerline{(c) Yale}
\end{minipage}
\hfill
\begin{minipage}{0.2\linewidth}
  \centerline{\includegraphics[width=4.0cm]{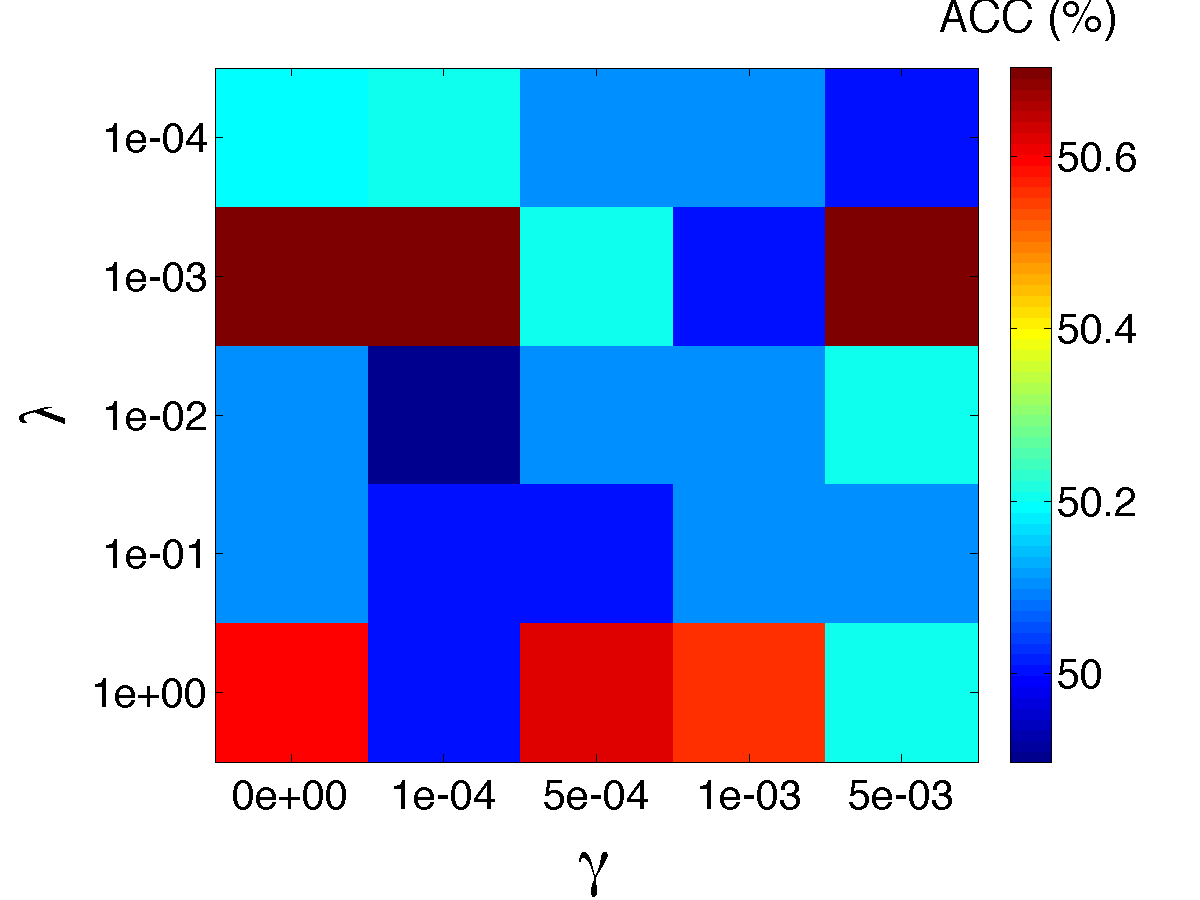}}
  \centerline{(d) PCMAC}
\end{minipage}
\vfill
\begin{minipage}{0.2\linewidth}
  \centerline{\includegraphics[width=4.0cm]{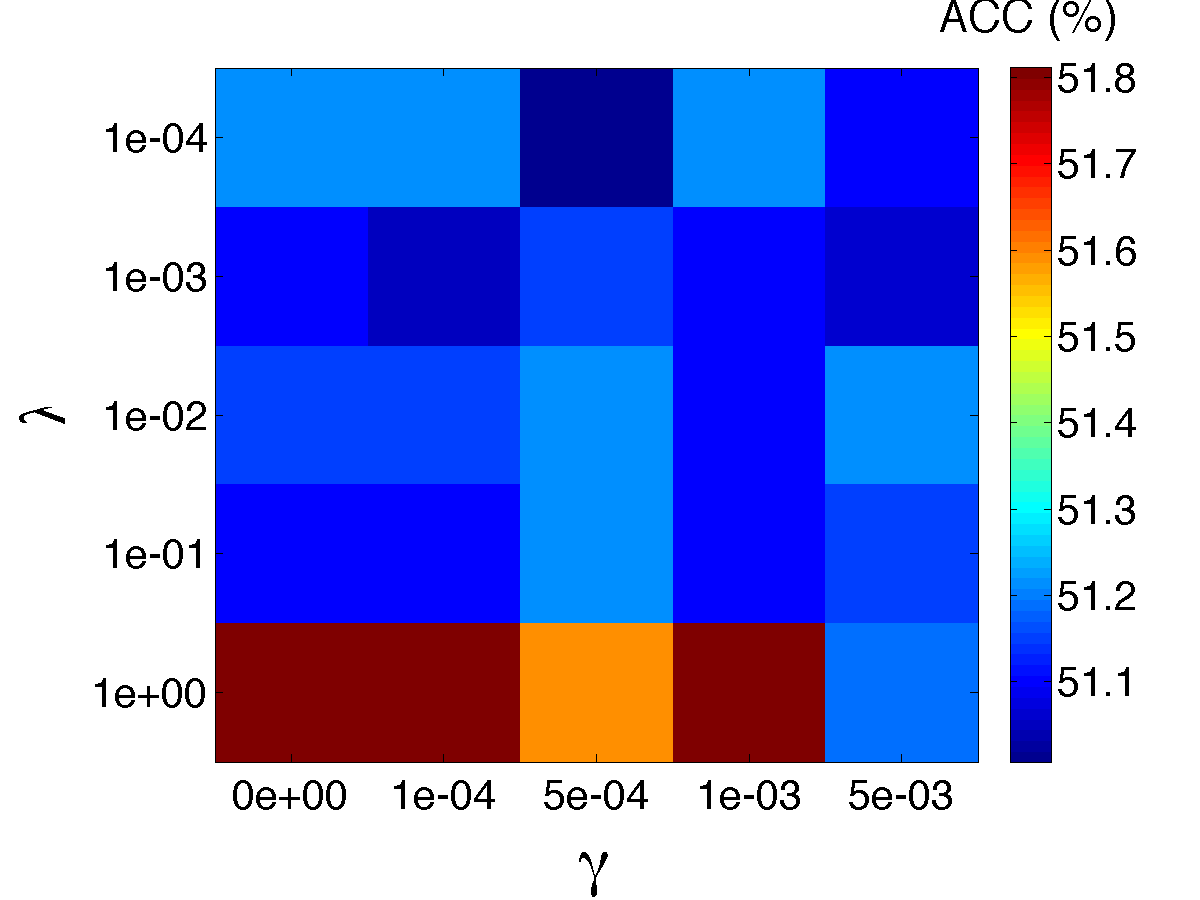}}
  \centerline{(e) BASEHOCK}
\end{minipage}
\hfill
\begin{minipage}{0.2\linewidth}
  \centerline{\includegraphics[width=4.0cm]{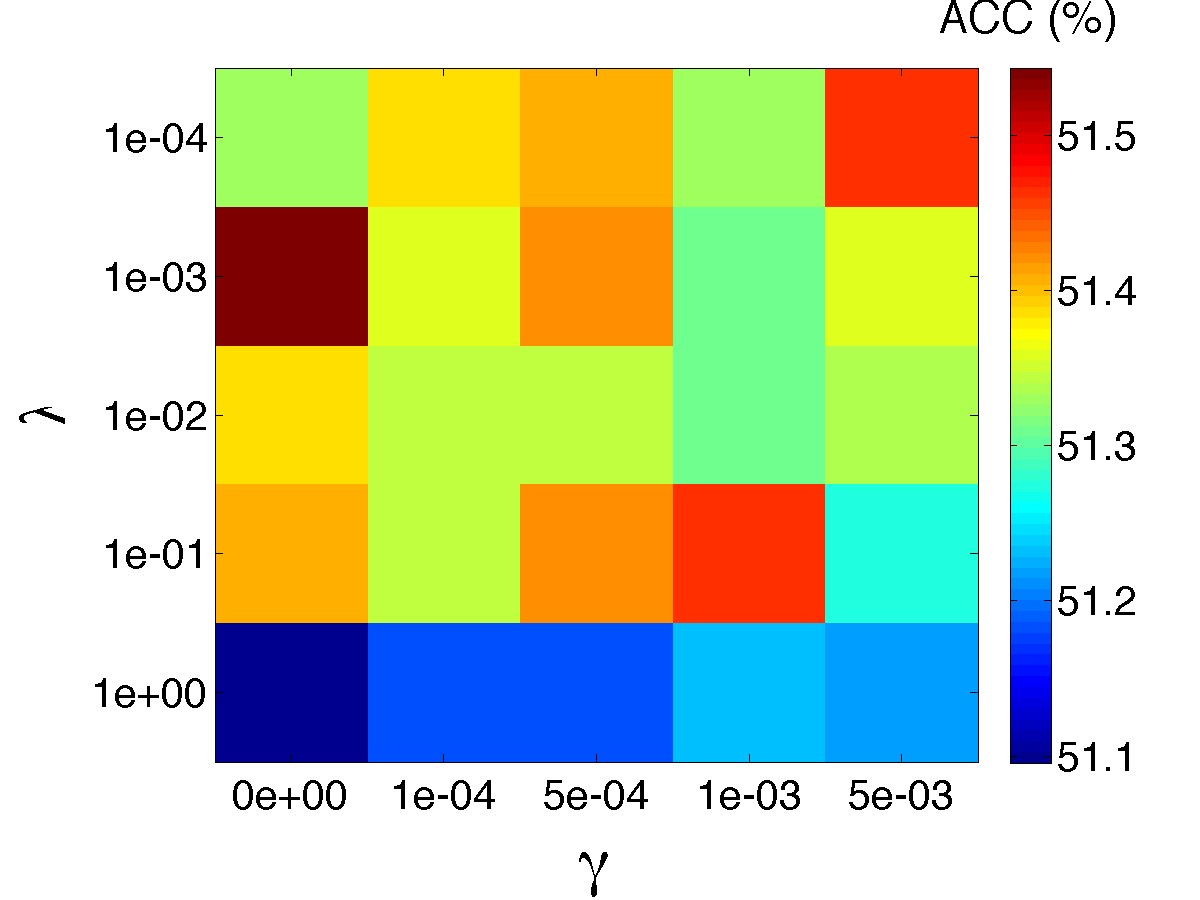}}
  \centerline{(f) RELATHE}
\end{minipage}
\hfill
\begin{minipage}{0.2\linewidth}
  \centerline{\includegraphics[width=4.0cm]{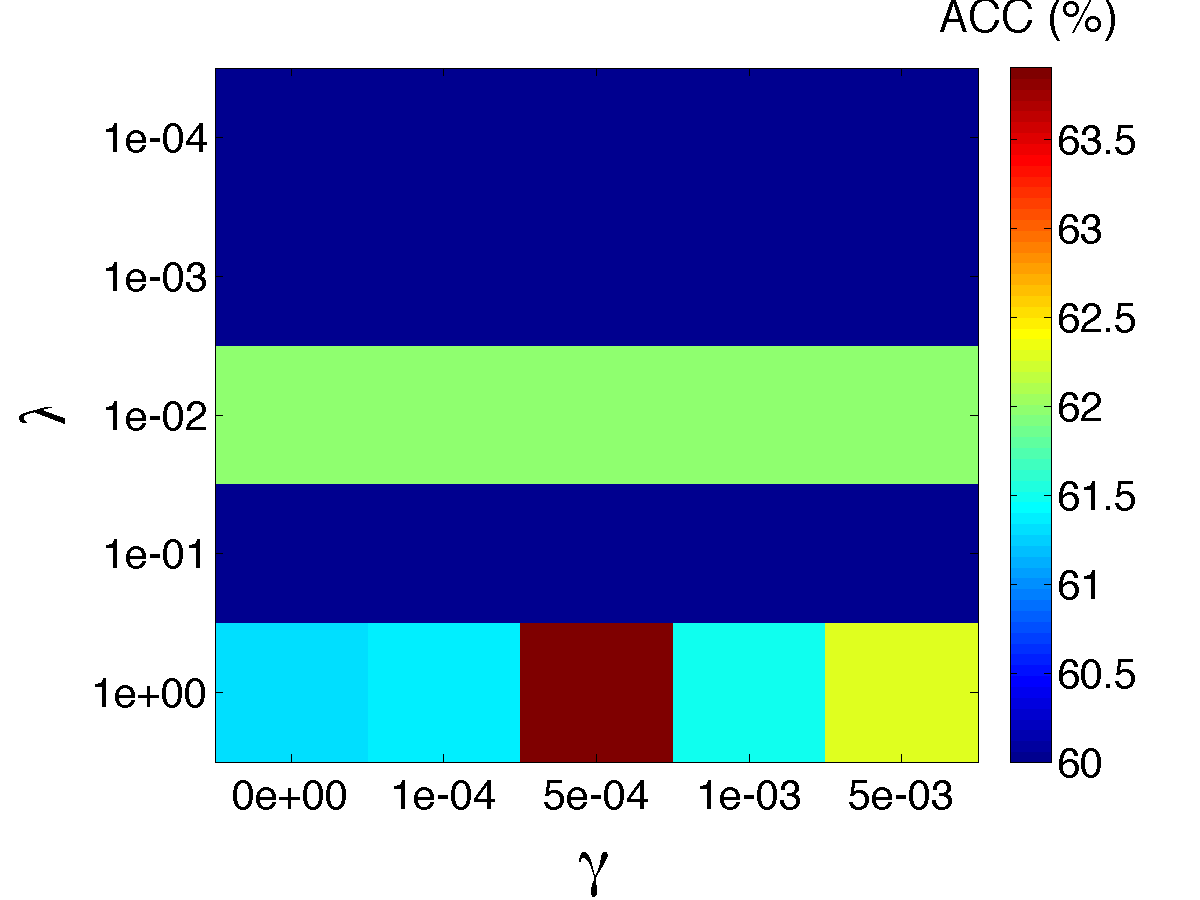}}
  \centerline{(g) Prostate\_GE}
\end{minipage}
\hfill
\begin{minipage}{0.2\linewidth}
  \centerline{\includegraphics[width=4.0cm]{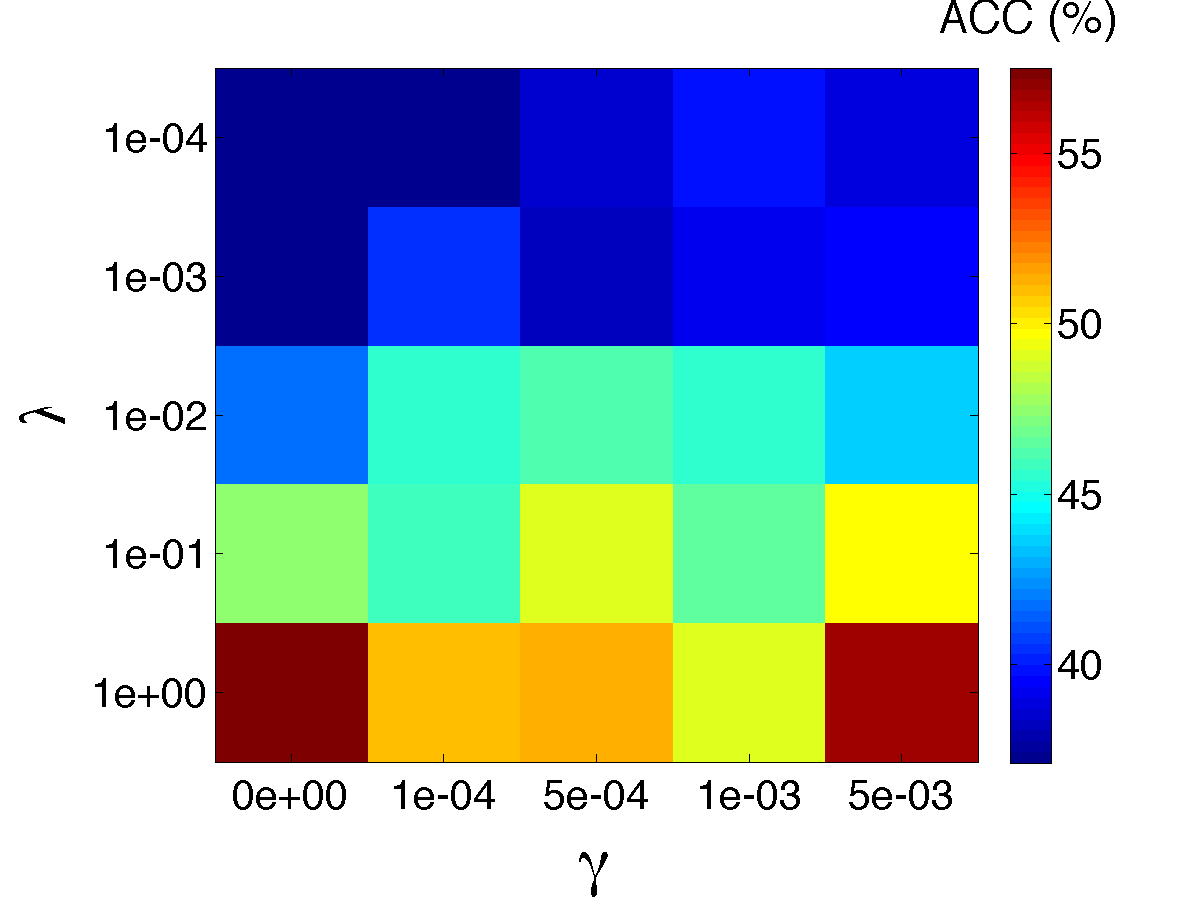}}
  \centerline{(h) Isolet}
\end{minipage}
\vfill
\begin{minipage}{0.2\linewidth}
\end{minipage}
\hfill
\begin{minipage}{0.2\linewidth}
  \centerline{\includegraphics[width=4.0cm]{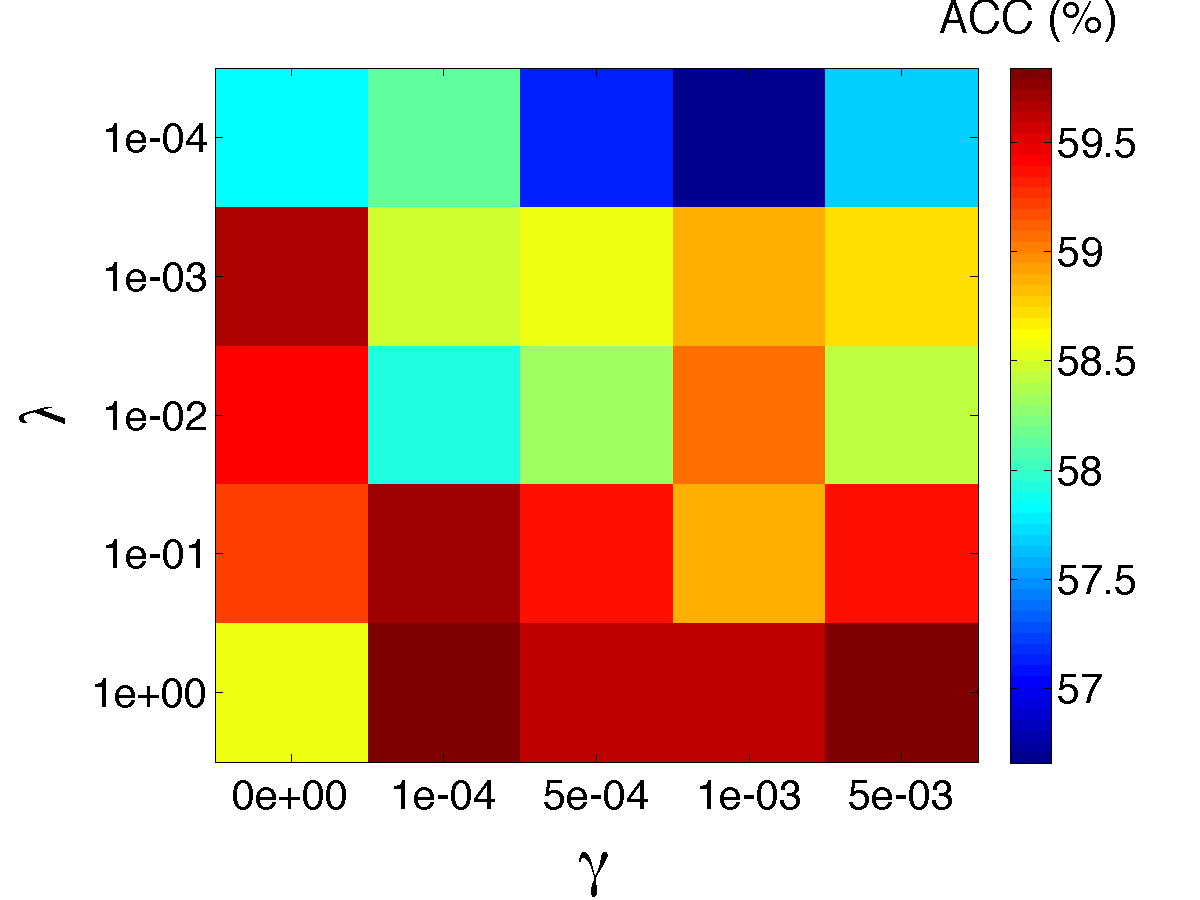}}
  \centerline{(g) Caltech101}
\end{minipage}
\hfill
\begin{minipage}{0.2\linewidth}
  \centerline{\includegraphics[width=4.0cm]{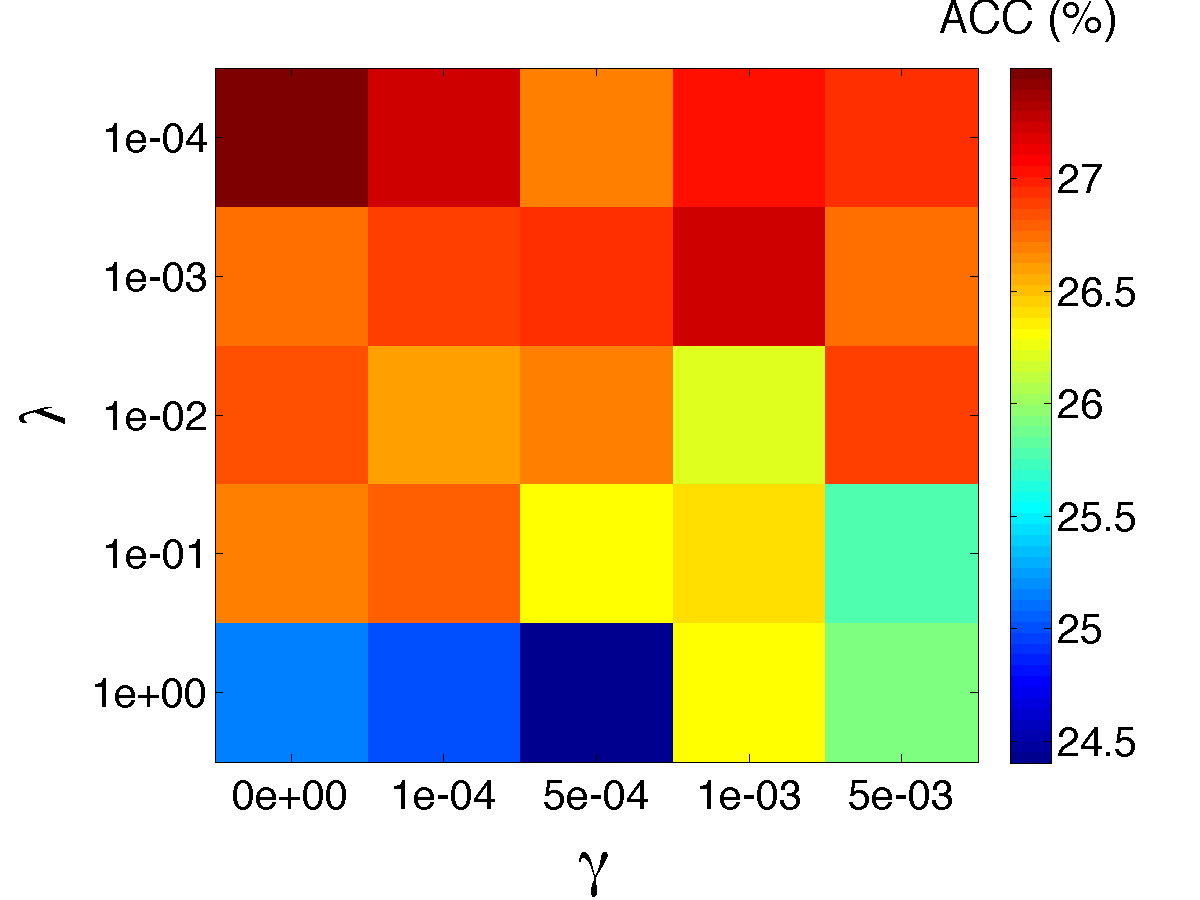}}
  \centerline{(h) CUB200}
\end{minipage}
\hfill
\begin{minipage}{0.2\linewidth}
\end{minipage}
\caption{{Performance of GAFS in clustering as a function of the percentage of features selected $p$ ($\%$) for varying values of the balance parameters $\lambda$ and $\gamma$. Clustering accuracy is used as the evaluation metric.}}
\label{sensBP}
\end{figure*}

We also study the {parameter sensitivity of GAFS with respect to the} balance parameters $\lambda$ and $\gamma$, {under a} fixed percentage of selected features and hidden layer size. We set $p=20\%$, as Fig.~\ref{sensHL} shows that the performance stabilizes starting at that value of $p$. For subspace dimensionality, we choose $m=10$ since Fig.~\ref{sensHL} shows that the performance of GAFS is not sensitive to the value of $m$. The performance results are shown in Fig. \ref{sensBP}, where we find that different datasets present different trends on the ACC values with respect to $\lambda$ and $\gamma$. However, we also find that the performance differences on PCMCA, BASEHOCK, and RELATHE are not greater than $0.8\%$, $0.8\%$, and $0.4\%$, respectively. Therefore we cannot make any conclusion on the influence from two balance parameters on ACC based on these $3$ datasets. For the parameter $\lambda$, which controls the column sparsity of $\mathbf{W}_1$, we can find that for Yale the performance monotonically improves as the value of $\lambda$ increases for each fixed value of $\gamma$, even though the number of selected features $m$ is fixed. We believe this is further evidence that a small number of selected features receiving large score (corresponding to large $\lambda$) is sufficient to obtain good learning performance, while having a large number of highly scoring features (corresponding to small $\lambda$) may introduce irrelevant features to the selection. We also find a similar behavior for Prostate\_GE and Isolet. For both MNIST and COIL20, we can find that the overall performance is best when $\lambda=10^{-2}$ and both smaller and larger values of $\lambda$ degrade the performance. This is because the diversity among instances of these two datasets is large enough: a large value of $\lambda$ may remove informative features, while a small value of $\lambda$ prevents the exclusion of small, irrelevant, or redundant features. For the parameter $\gamma$, which controls local data structure preservation, we can find that both large values and small values of $\gamma$ degrade performance. On one hand, we can conclude that local data structure preservation does help improve feature selection performance to a certain degree. On the other hand, large weights on local data structure preservation may also harm feature selection performance. \par

\subsection{Feature Selection Illustration}
\begin{figure*}[htbp] 
\centering{\includegraphics[width=16.5cm]{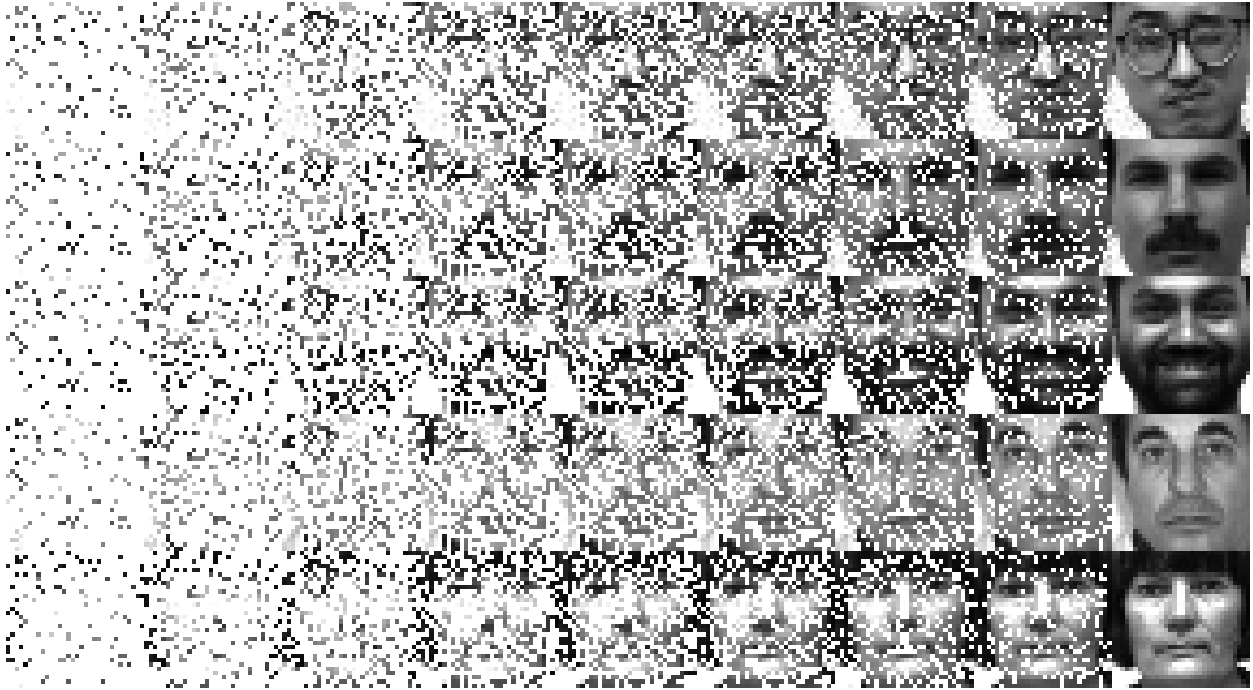}}
\caption{Feature selection illustration on Yale. Each row corresponds to a sample human face image and each column refers to percentages of features selected $p \in \{10\%, 20\%, 30\%, 40\%, 50\%, 60\%, 70\%, 80\%, 100\%\}$ from left to right.}
\label{featSelectIllust}
\end{figure*}

We randomly select five samples from the Yale dataset to illustrate the choices made by different feature selection algorithms. For each sample, $p \in \{10\%, 20\%, 30\%, 40\%, 50\%, 60\%, 70\%, 80\%, 100\%\}$ features are selected. Figure \ref{featSelectIllust} shows images corresponding to the selected features (i.e., pixels) for each sample and value of $p$, with unselected pixels shown in white.  The figure shows that GAFS is able to capture the most discriminative parts on human face such as eyes, nose, and mouse. \par

\subsection{Performance Comparison}
\begin{figure*}
\begin{minipage}{0.2\linewidth}
  \centerline{\includegraphics[width=4.0cm]{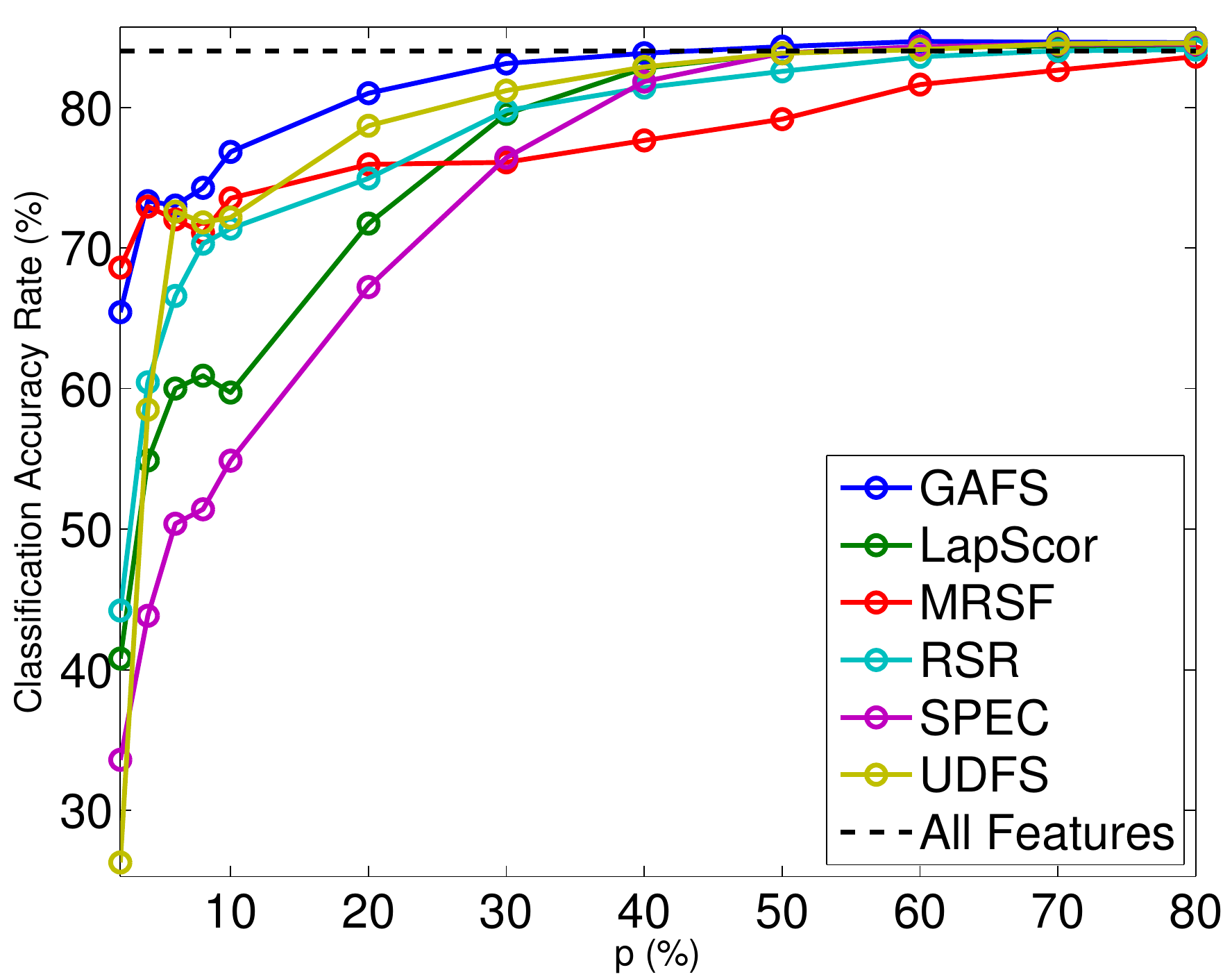}}
  \centerline{(a) MNIST}
\end{minipage}
\hfill
\begin{minipage}{0.2\linewidth}
  \centerline{\includegraphics[width=4.0cm]{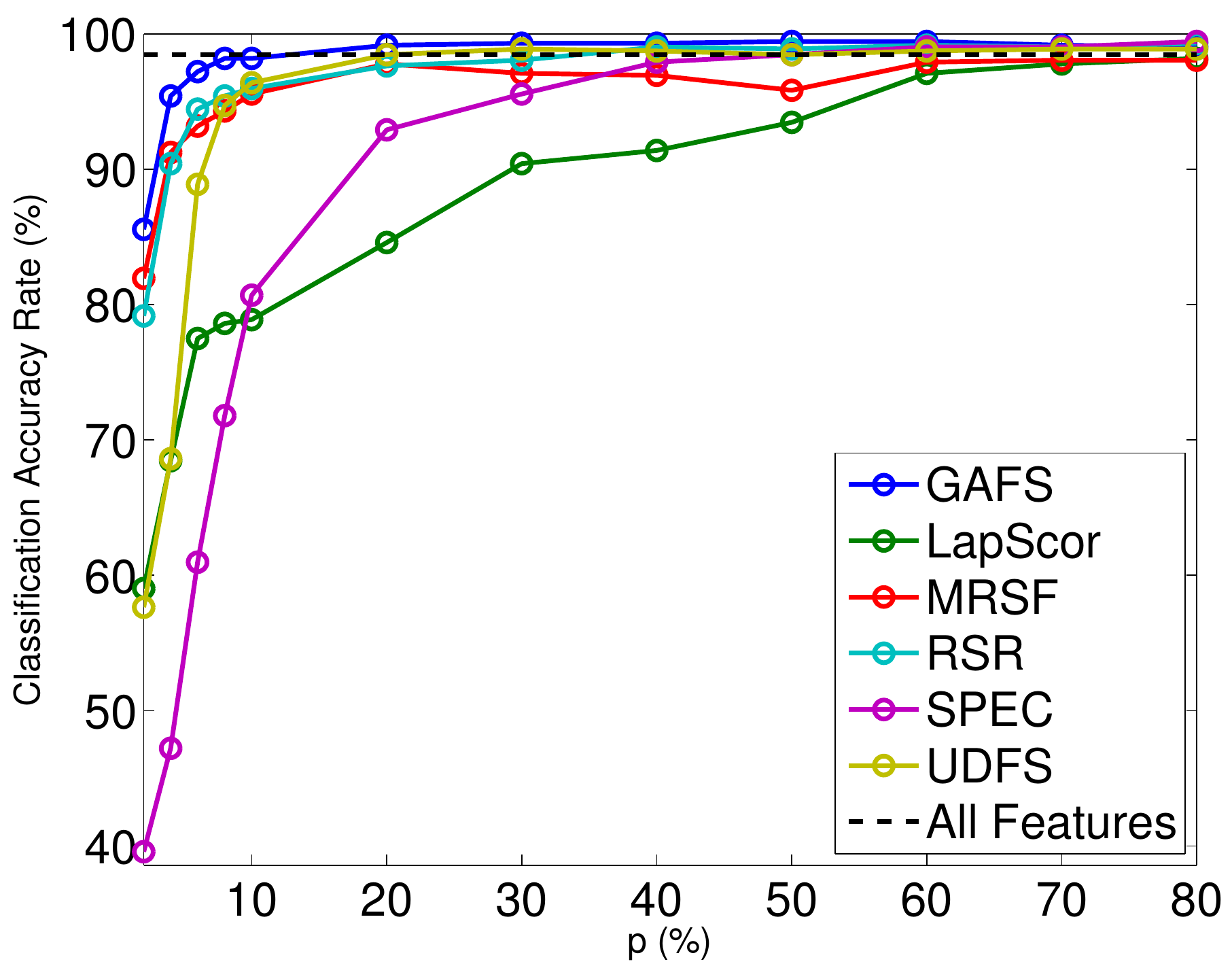}}
  \centerline{(b) COIL20}
\end{minipage}
\hfill
\begin{minipage}{0.2\linewidth}
  \centerline{\includegraphics[width=4.0cm]{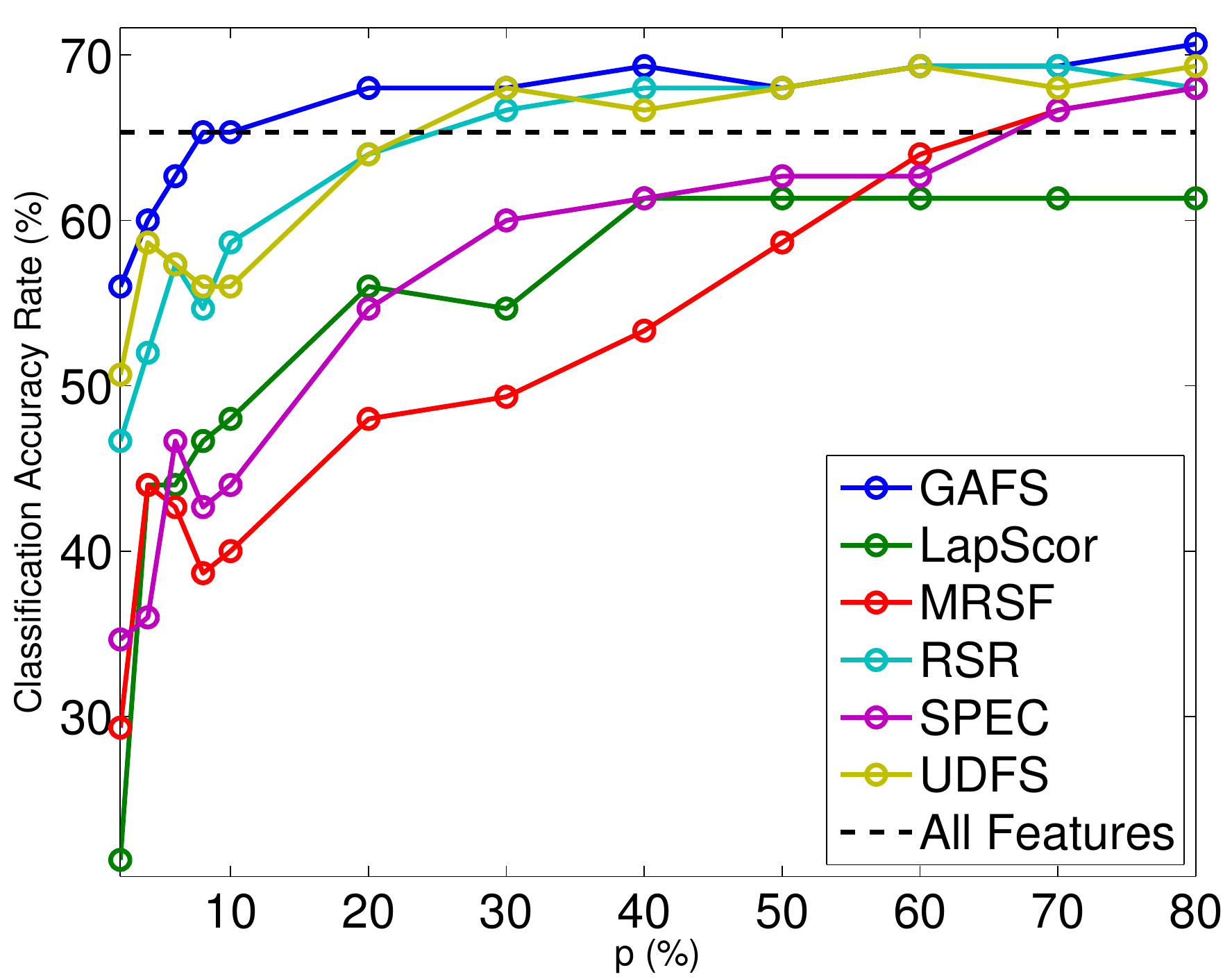}}
  \centerline{(c) Yale}
\end{minipage}
\hfill
\begin{minipage}{0.2\linewidth}
  \centerline{\includegraphics[width=4.0cm]{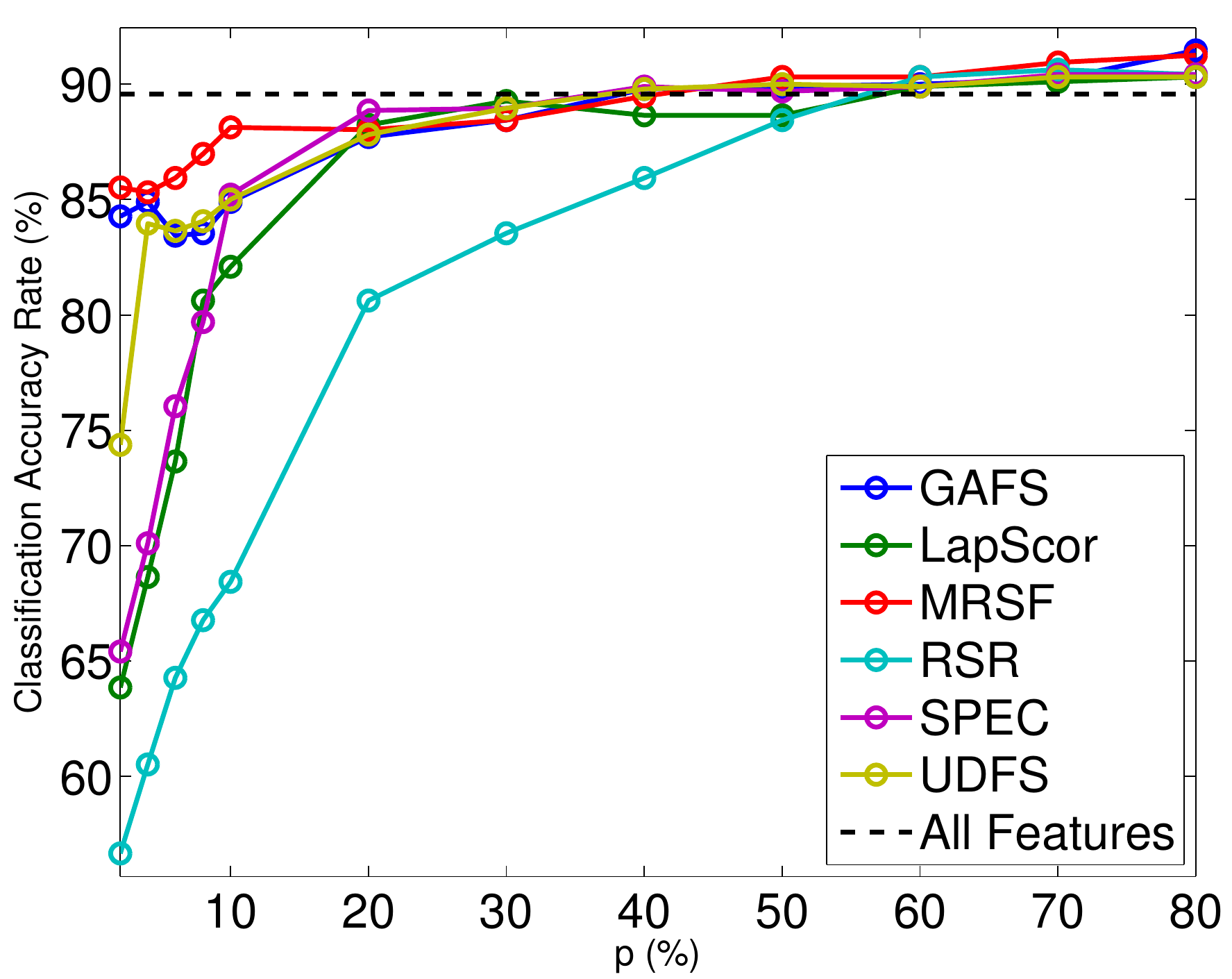}}
  \centerline{(d) PCMAC}
\end{minipage}
\vfill
\begin{minipage}{0.2\linewidth}
  \centerline{\includegraphics[width=4.0cm]{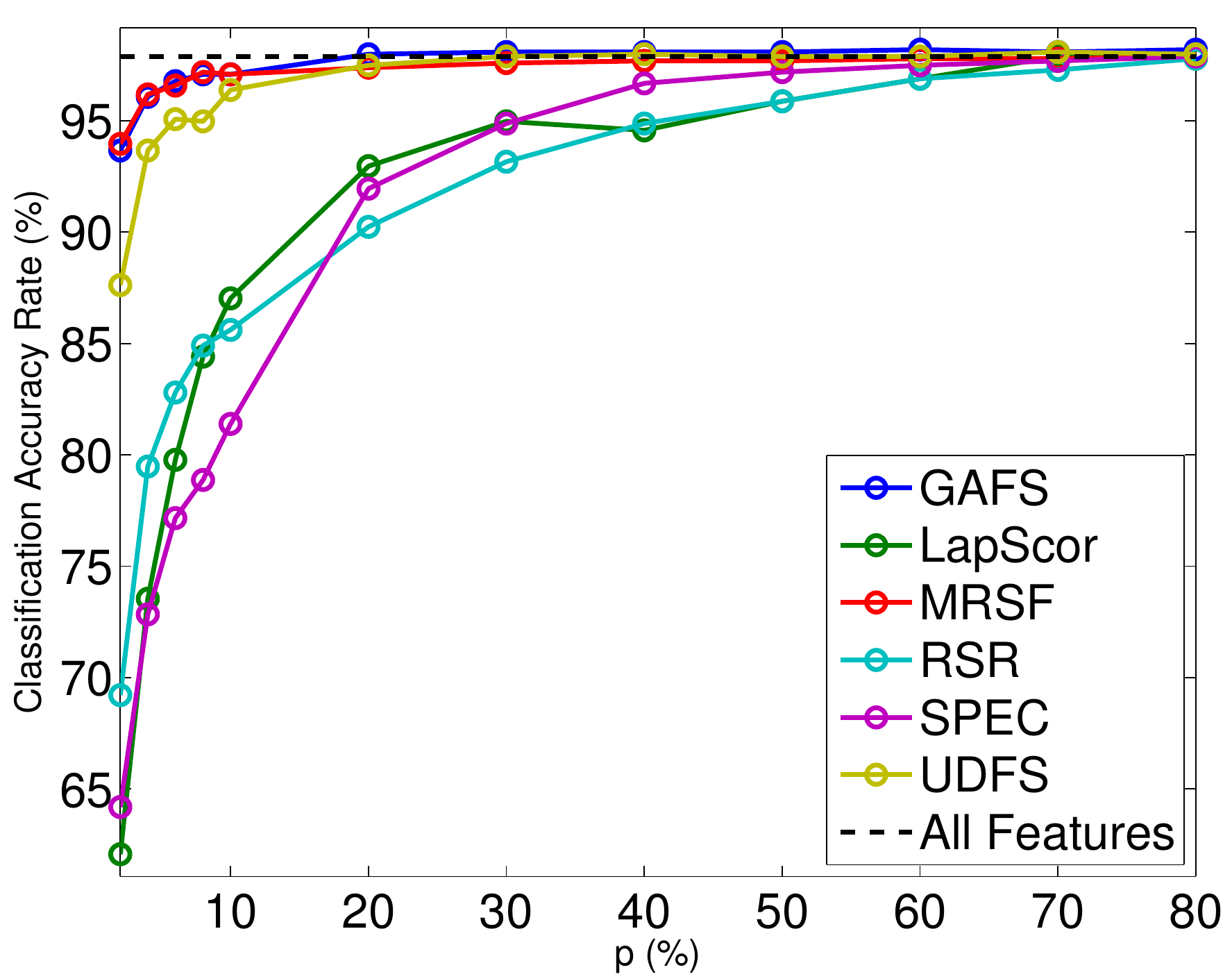}}
  \centerline{(e) BASEHOCK}
\end{minipage}
\hfill
\begin{minipage}{0.2\linewidth}
  \centerline{\includegraphics[width=4.0cm]{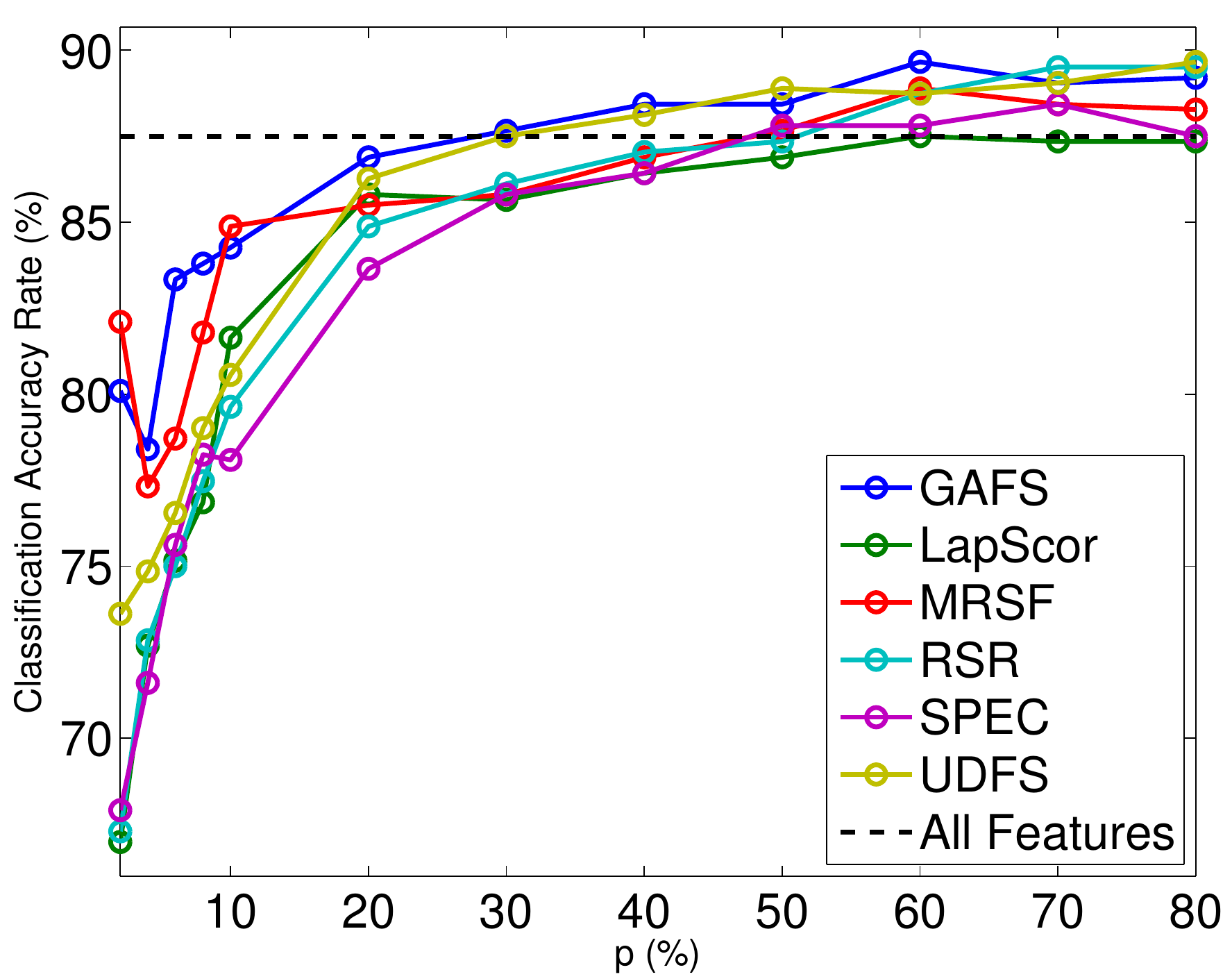}}
  \centerline{(f) RELATHE}
\end{minipage}
\hfill
\begin{minipage}{0.2\linewidth}
  \centerline{\includegraphics[width=4.0cm]{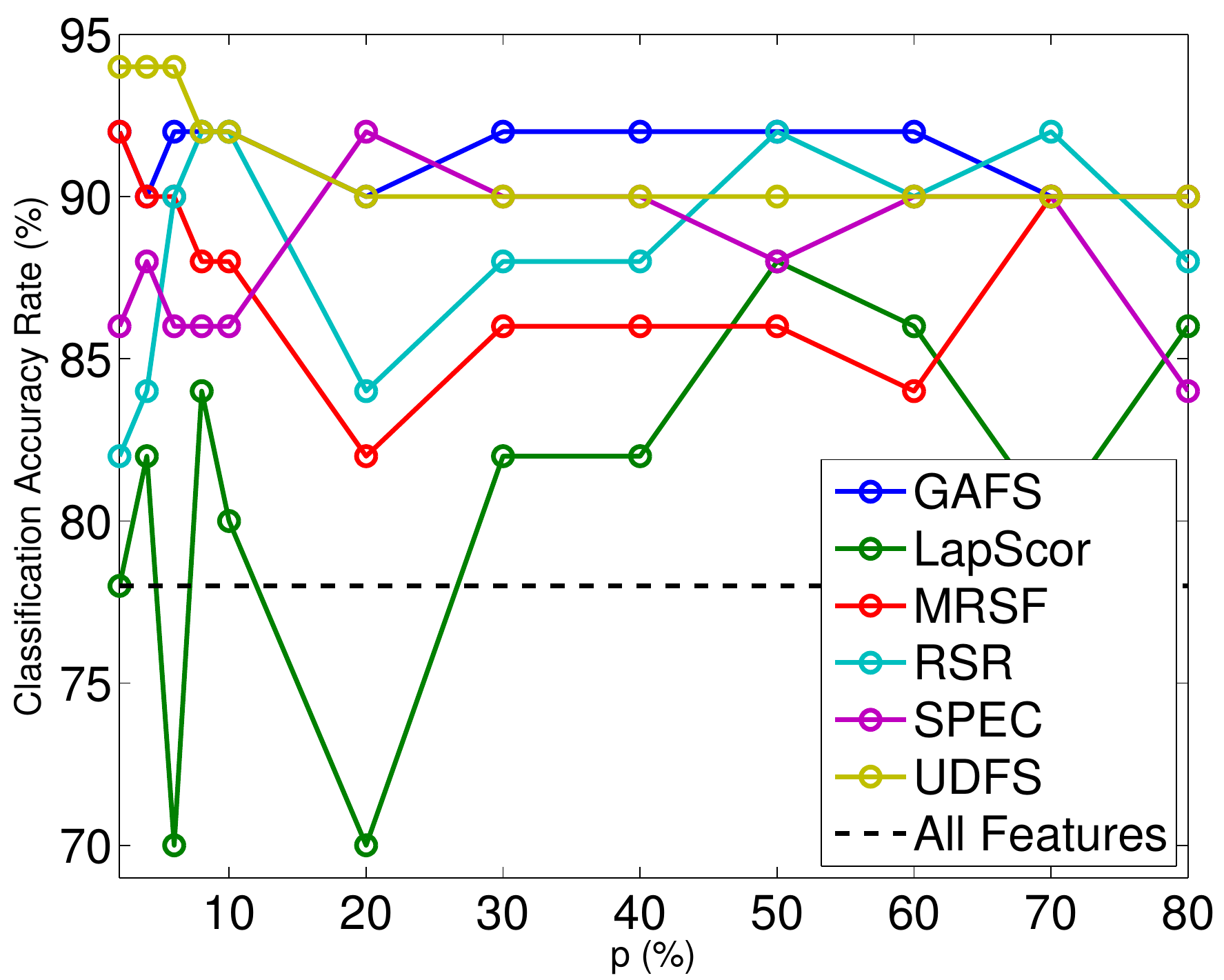}}
  \centerline{(g) Prostate\_GE}
\end{minipage}
\hfill
\begin{minipage}{0.2\linewidth}
  \centerline{\includegraphics[width=4.0cm]{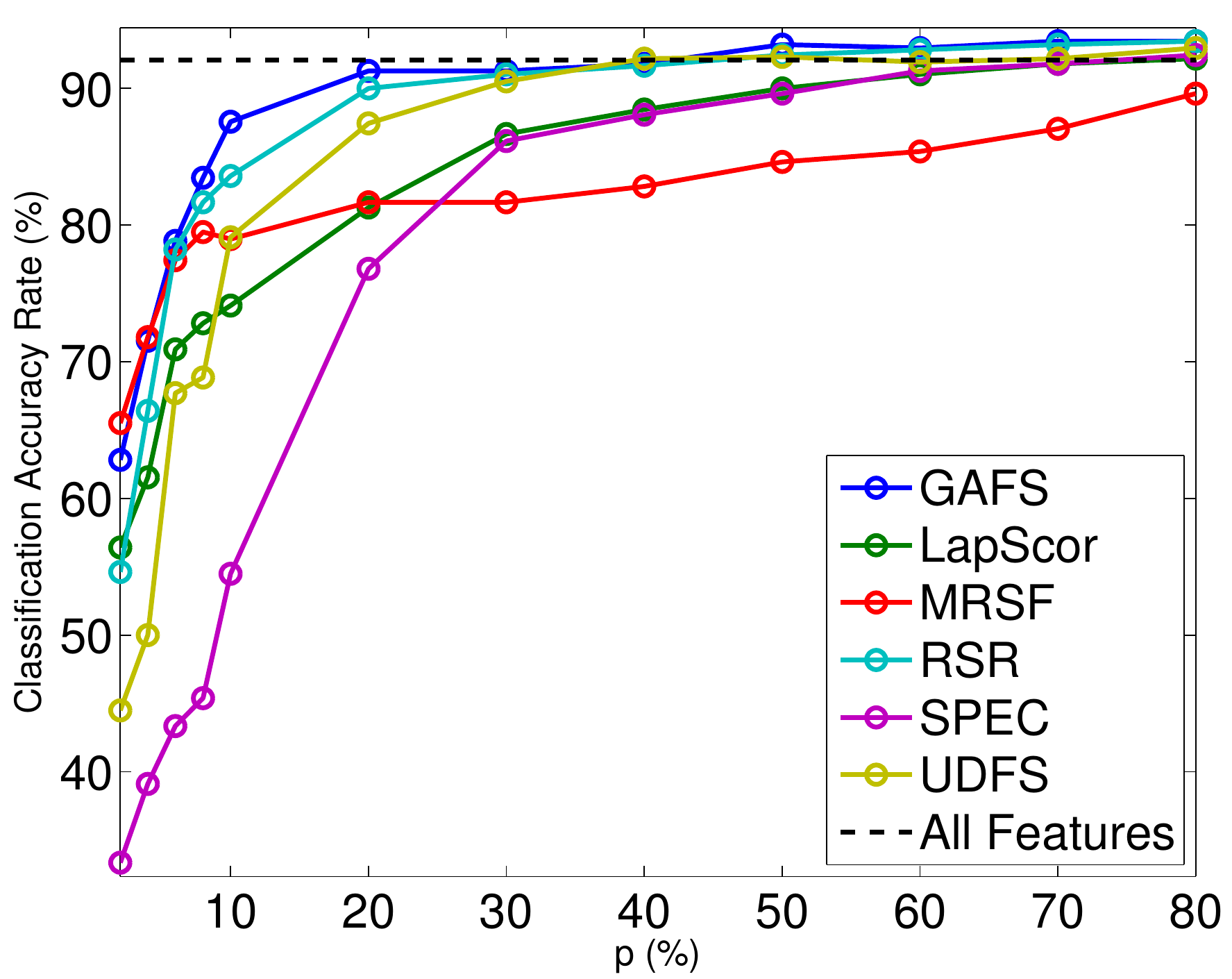}}
  \centerline{(h) Isolet}
\end{minipage}
\vfill
\begin{minipage}{0.2\linewidth}
\end{minipage}
\hfill
\begin{minipage}{0.2\linewidth}
  \centerline{\includegraphics[width=4.0cm]{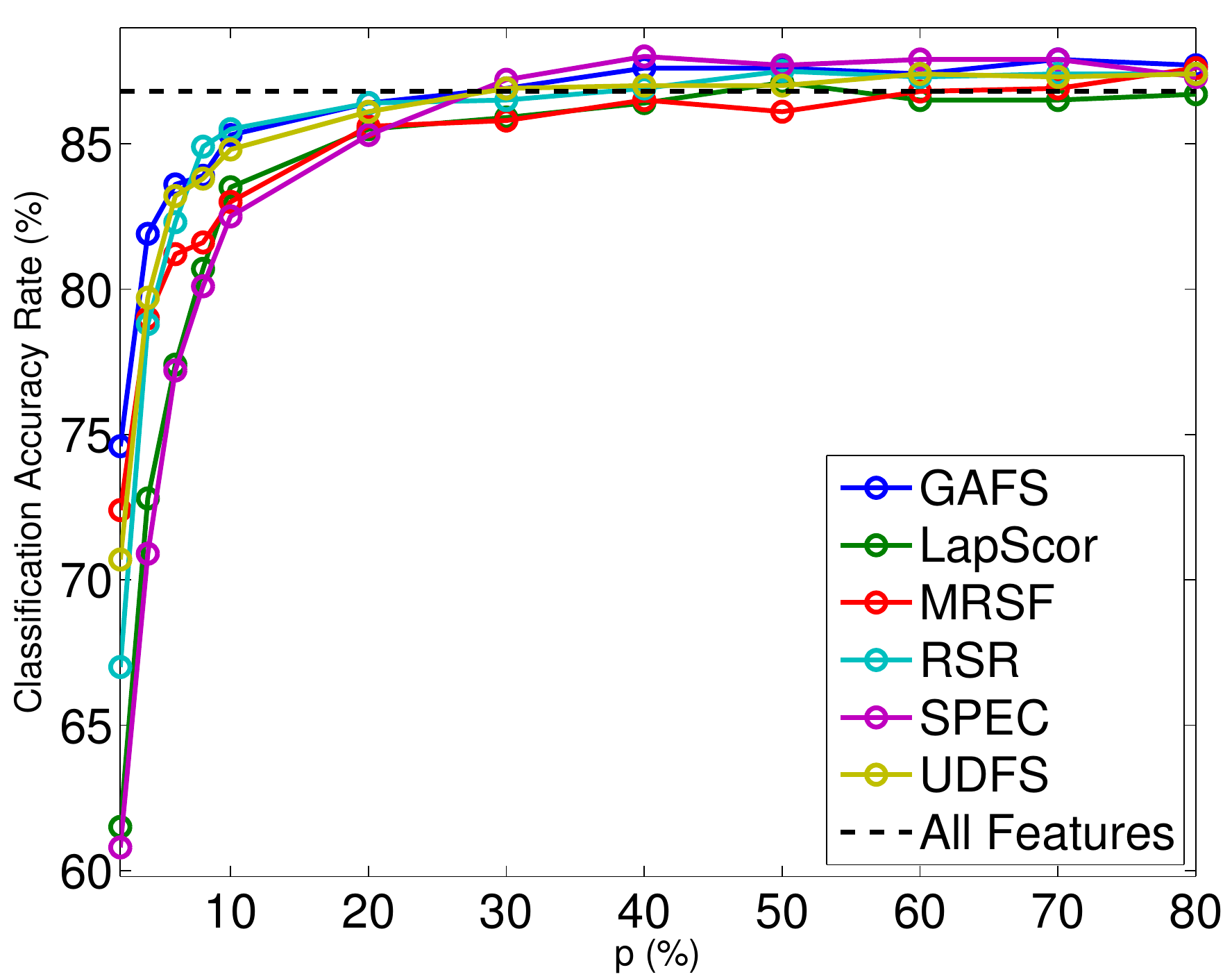}}
  \centerline{(g) Caltech101}
\end{minipage}
\hfill
\begin{minipage}{0.2\linewidth}
  \centerline{\includegraphics[width=4.0cm]{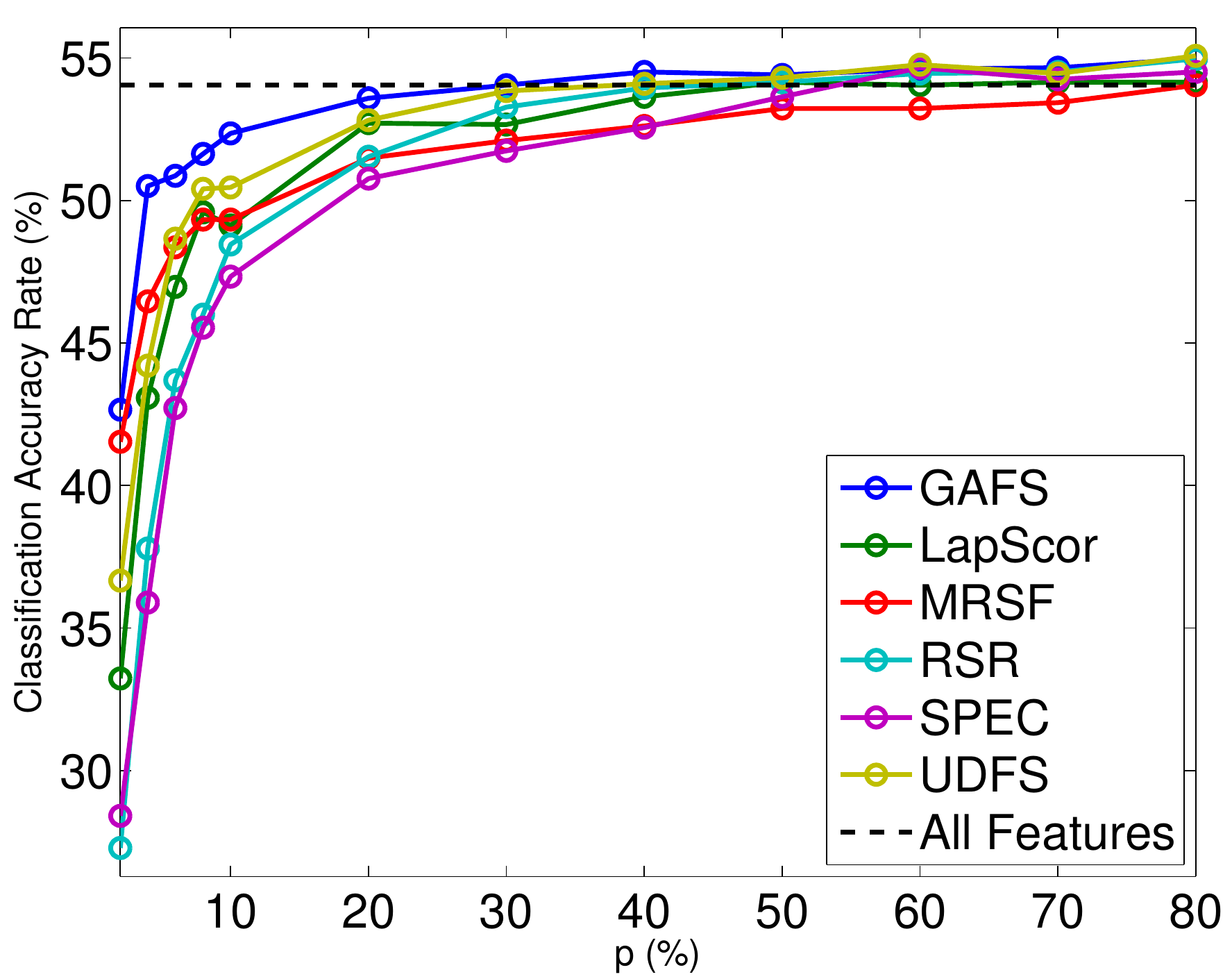}}
  \centerline{(h) CUB200}
\end{minipage}
\hfill
\begin{minipage}{0.2\linewidth}
\end{minipage}
\caption{{Performance of GAFS and competing feature selection algorithms in classification as a function of the percentage of features selected $p$ ($\%$). Classification accuracy is used as the evaluation metric.}}
\label{clsACC}
\end{figure*}

\begin{figure*}
\begin{minipage}{0.2\linewidth}
  \centerline{\includegraphics[width=4.0cm]{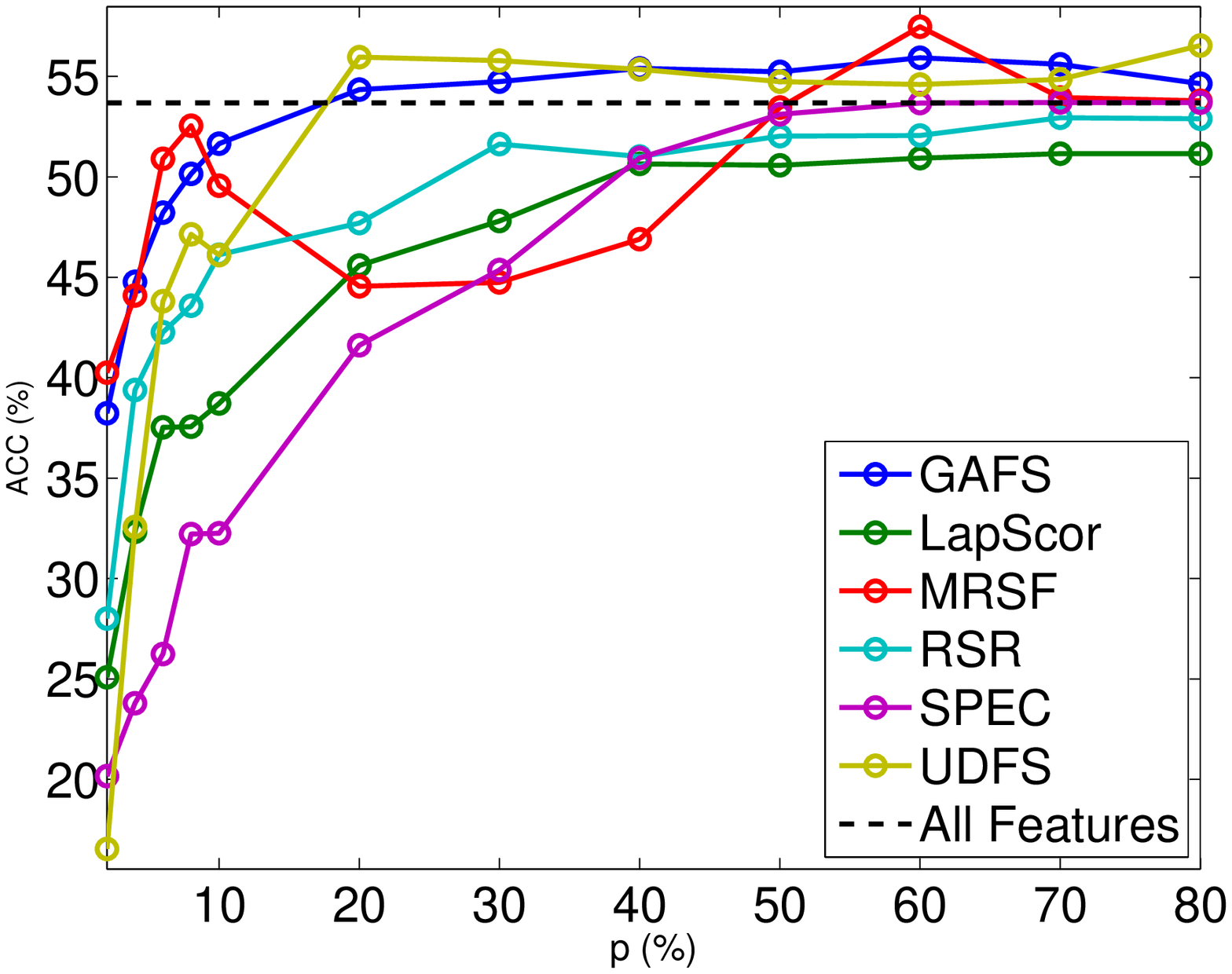}}
  \centerline{(a) MNIST}
\end{minipage}
\hfill
\begin{minipage}{0.2\linewidth}
  \centerline{\includegraphics[width=4.0cm]{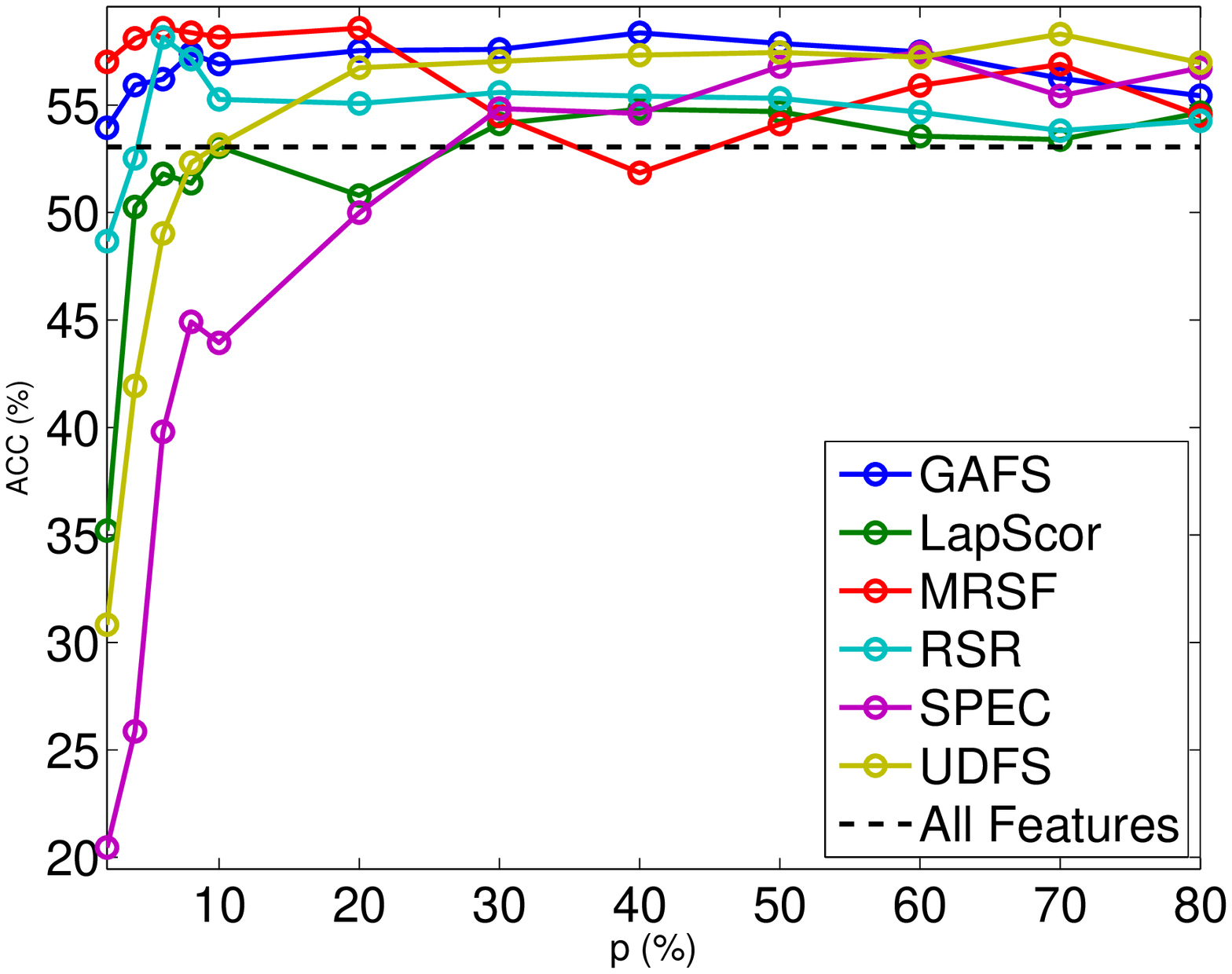}}
  \centerline{(b) COIL20}
\end{minipage}
\hfill
\begin{minipage}{0.2\linewidth}
  \centerline{\includegraphics[width=4.0cm]{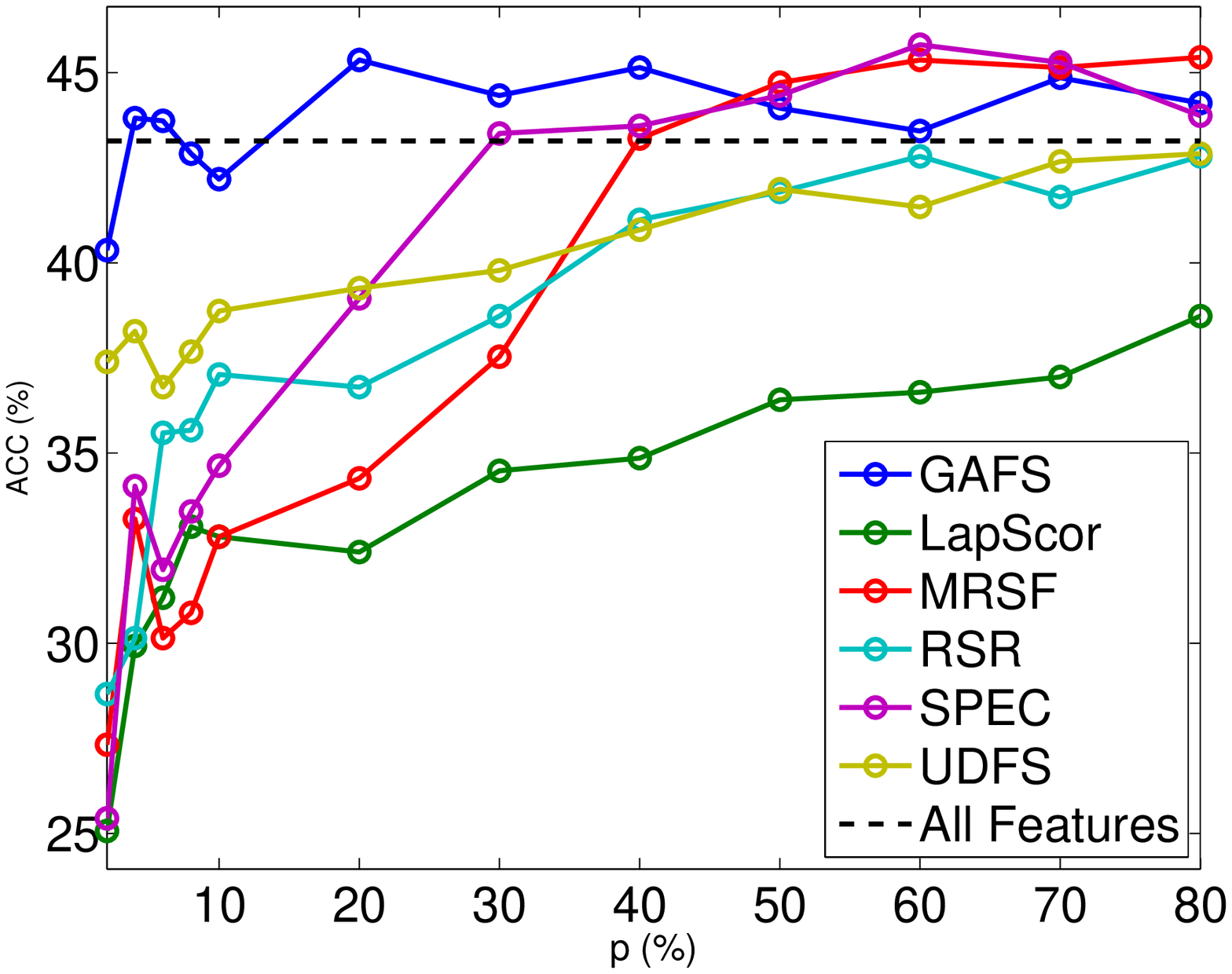}}
  \centerline{(c) Yale}
\end{minipage}
\hfill
\begin{minipage}{0.2\linewidth}
  \centerline{\includegraphics[width=4.0cm]{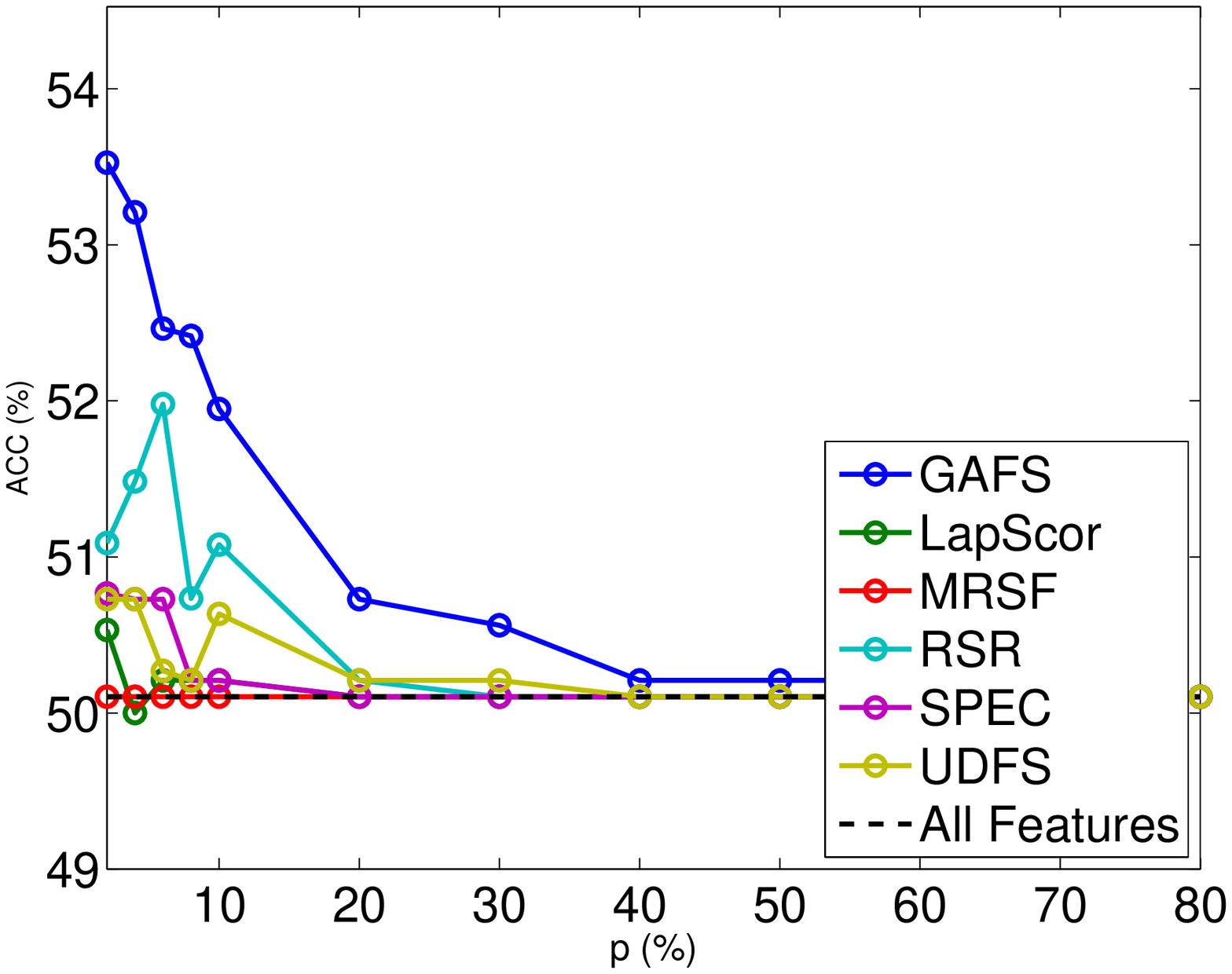}}
  \centerline{(d) PCMAC}
\end{minipage}
\vfill
\begin{minipage}{0.2\linewidth}
  \centerline{\includegraphics[width=4.0cm]{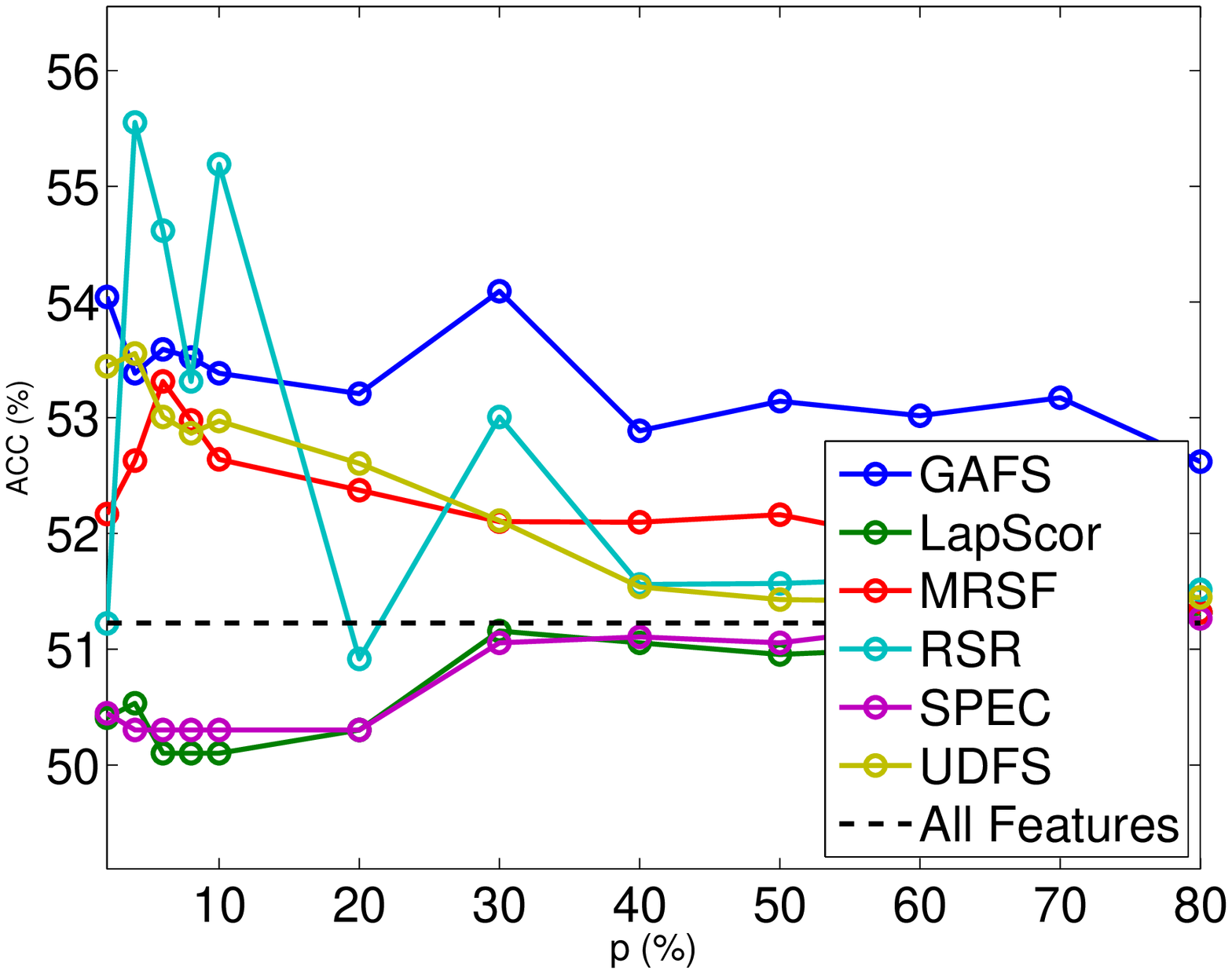}}
  \centerline{(e) BASEHOCK}
\end{minipage}
\hfill
\begin{minipage}{0.2\linewidth}
  \centerline{\includegraphics[width=4.0cm]{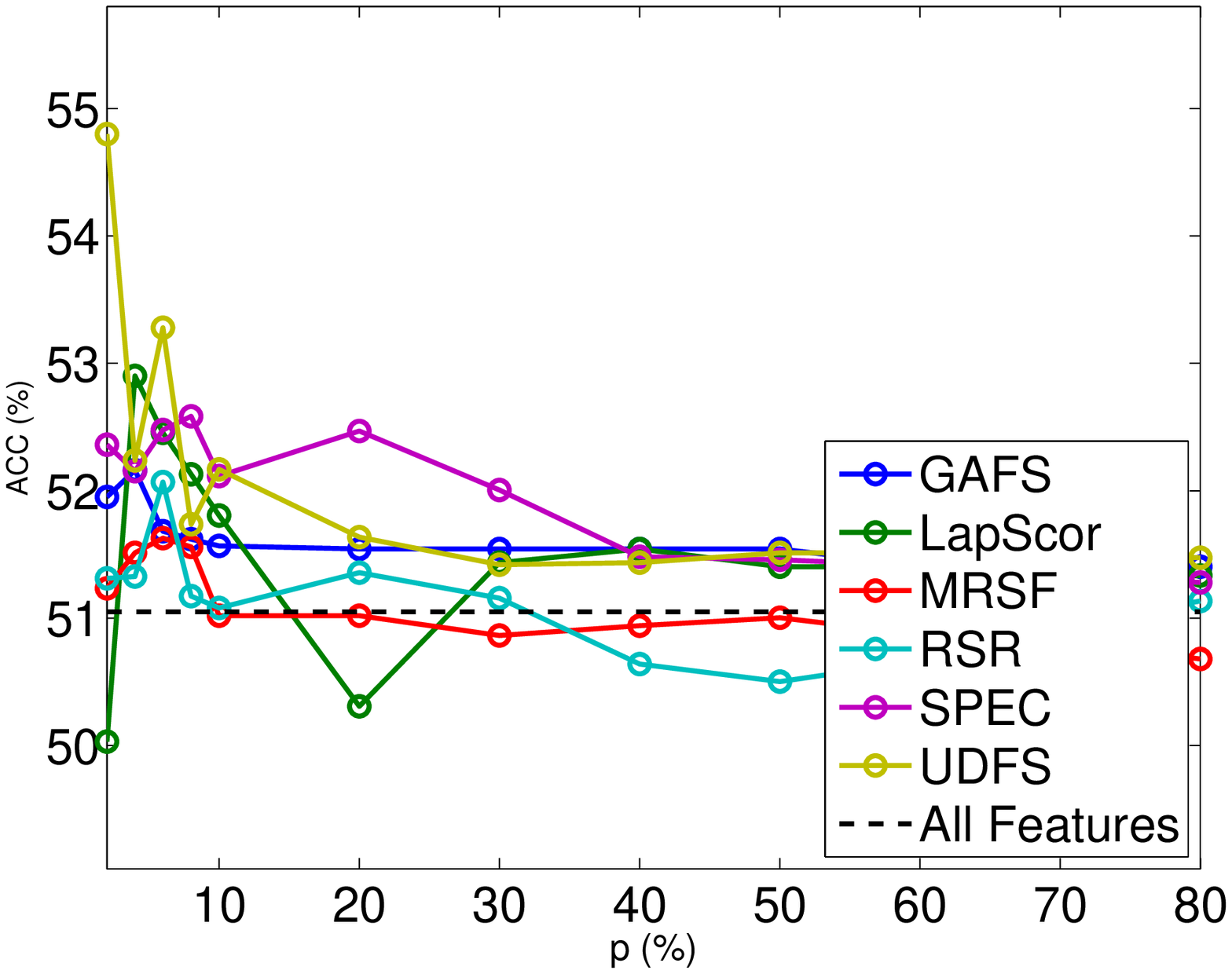}}
  \centerline{(f) RELATHE}
\end{minipage}
\hfill
\begin{minipage}{0.2\linewidth}
  \centerline{\includegraphics[width=4.0cm]{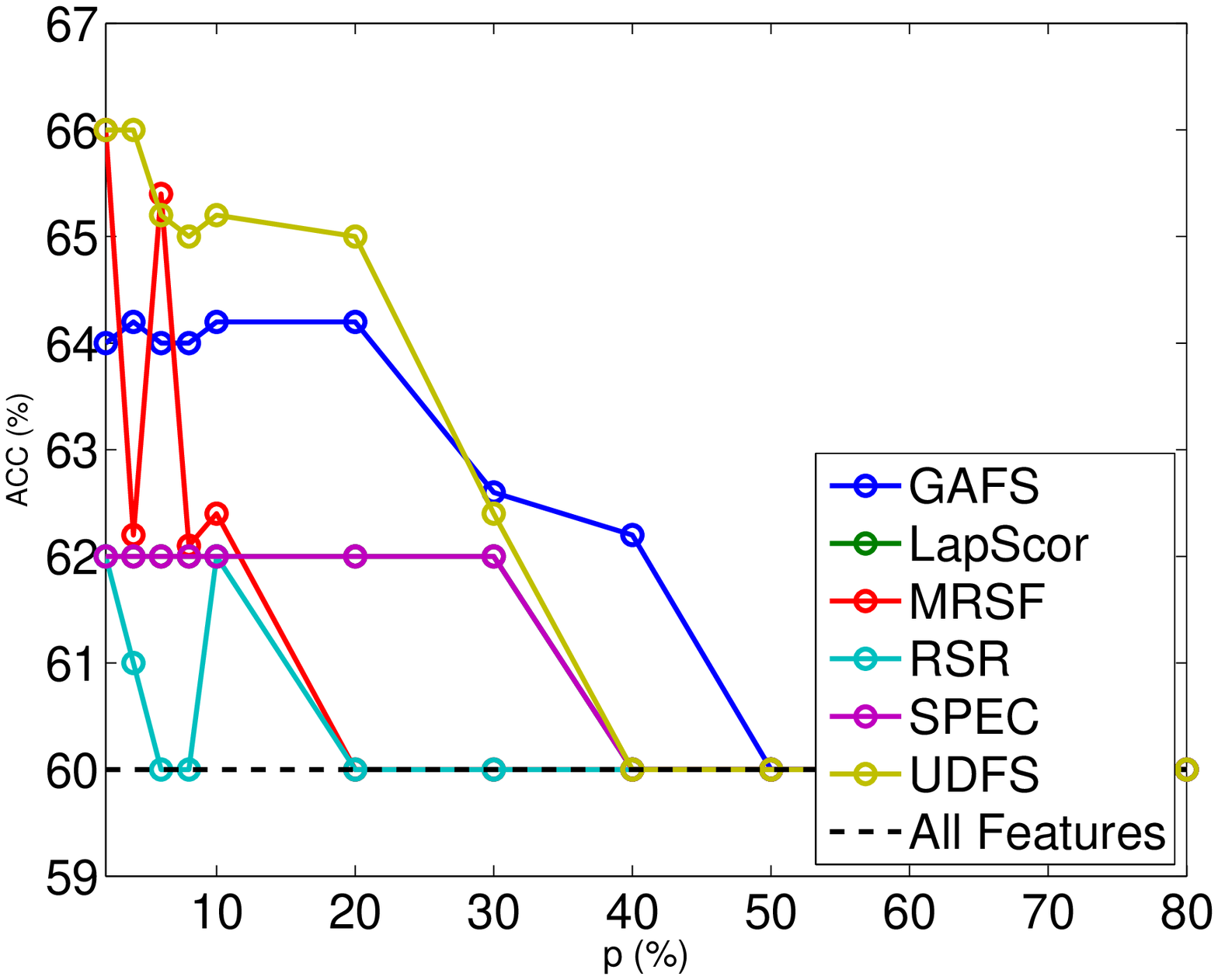}}
  \centerline{(g) Prostate\_GE}
\end{minipage}
\hfill
\begin{minipage}{0.2\linewidth}
  \centerline{\includegraphics[width=4.0cm]{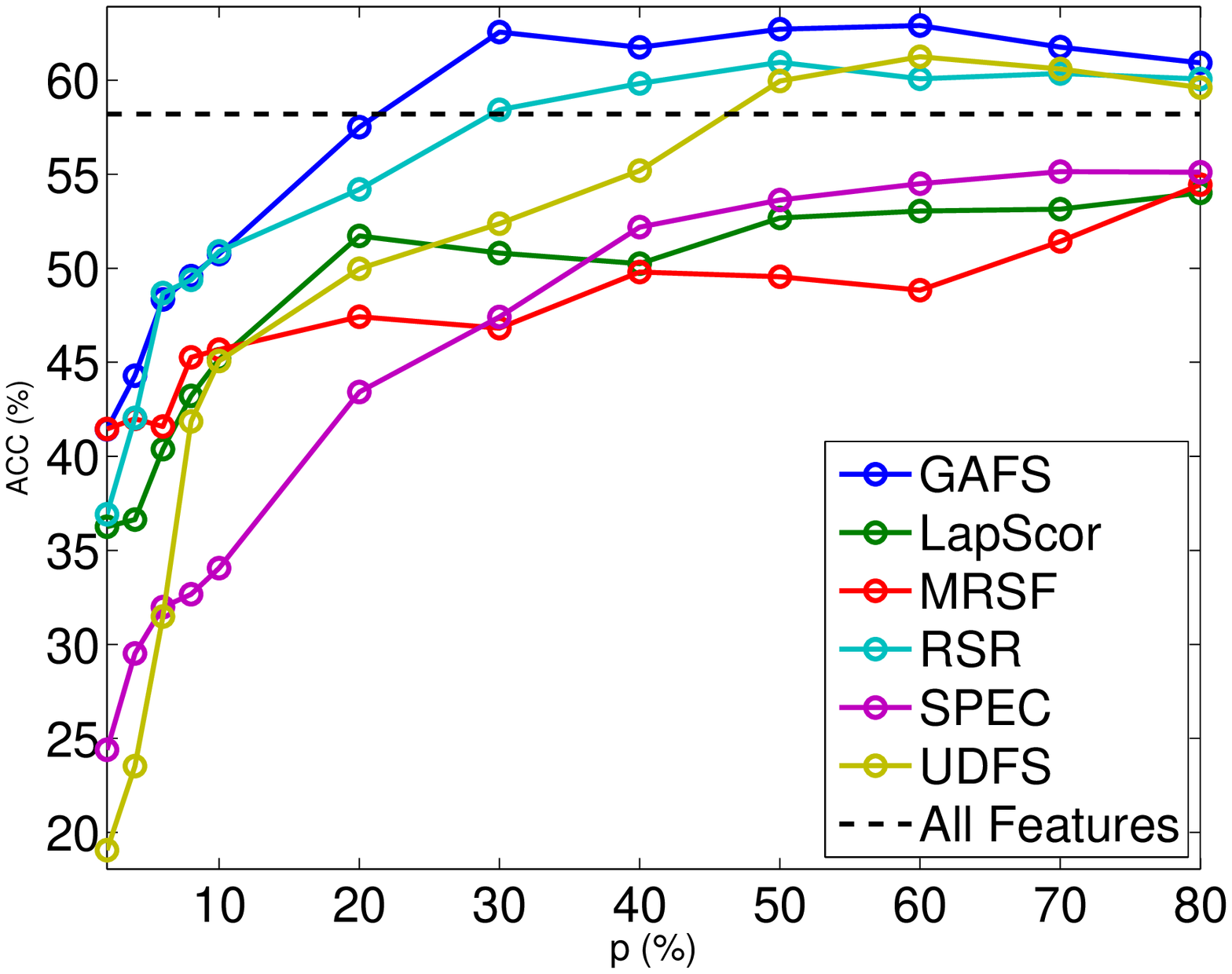}}
  \centerline{(h) Isolet}
\end{minipage}
\vfill
\begin{minipage}{0.2\linewidth}
\end{minipage}
\hfill
\begin{minipage}{0.2\linewidth}
  \centerline{\includegraphics[width=4.0cm]{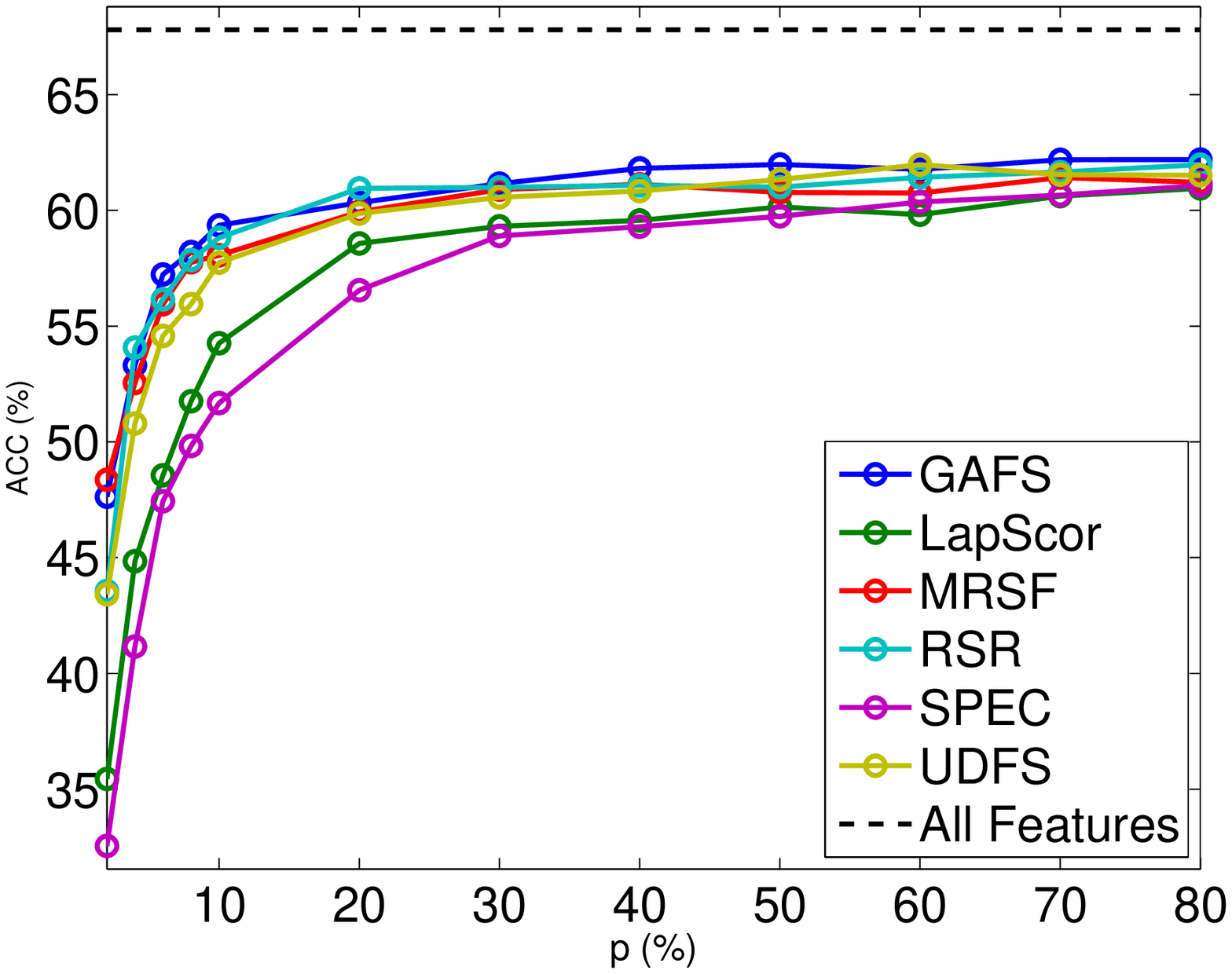}}
  \centerline{(g) Caltech101}
\end{minipage}
\hfill
\begin{minipage}{0.2\linewidth}
  \centerline{\includegraphics[width=4.0cm]{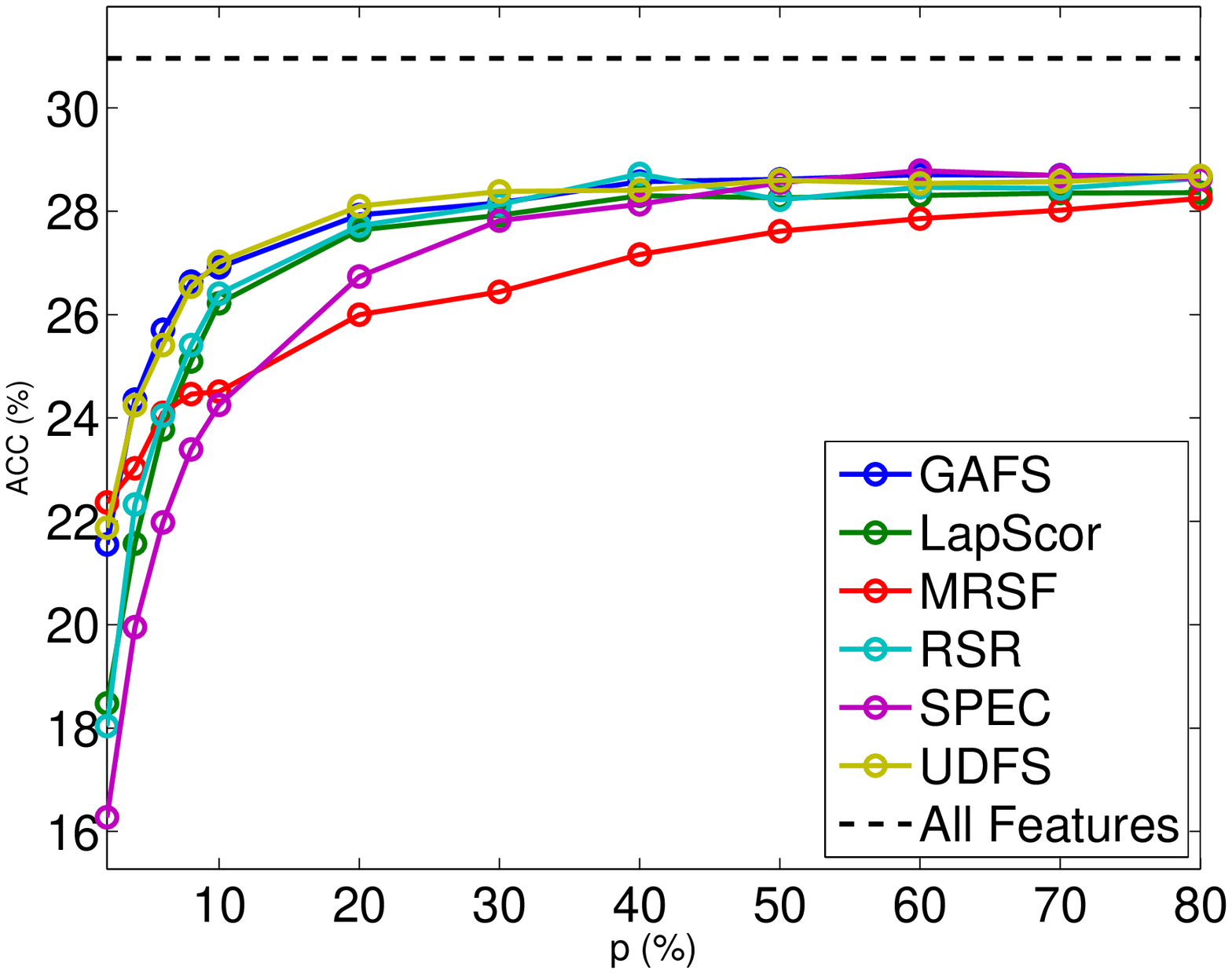}}
  \centerline{(h) CUB200}
\end{minipage}
\hfill
\begin{minipage}{0.2\linewidth}
\end{minipage}
\caption{{Performance of GAFS and competing feature selection algorithms in clustering as a function of the percentage of features selected $p$ ($\%$). Clustering accuracy is used as the evaluation metric.}}
\label{ACC}
\end{figure*}

\begin{figure*}
\begin{minipage}{0.2\linewidth}
  \centerline{\includegraphics[width=4.0cm]{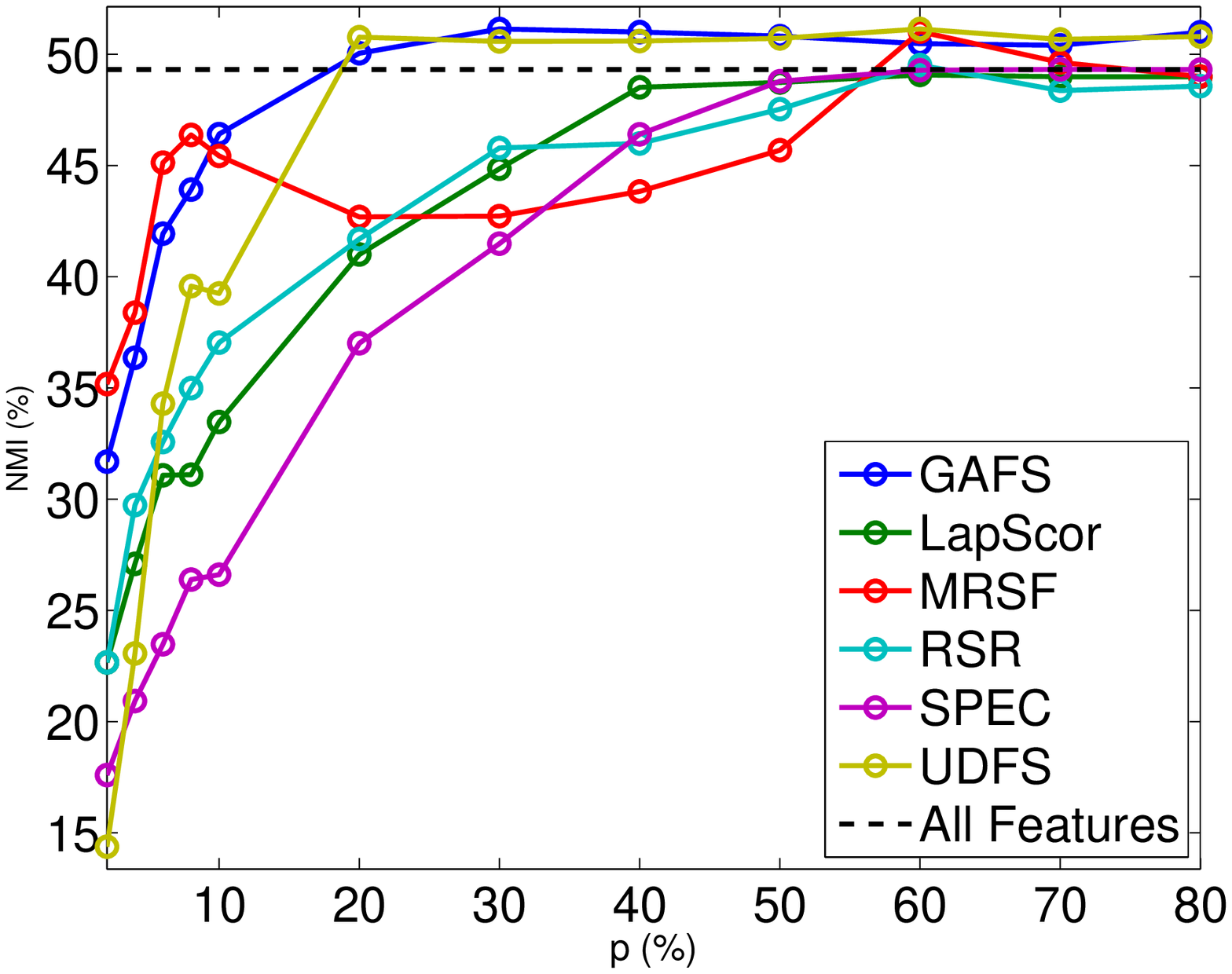}}
  \centerline{(a) MNIST}
\end{minipage}
\hfill
\begin{minipage}{0.2\linewidth}
  \centerline{\includegraphics[width=4.0cm]{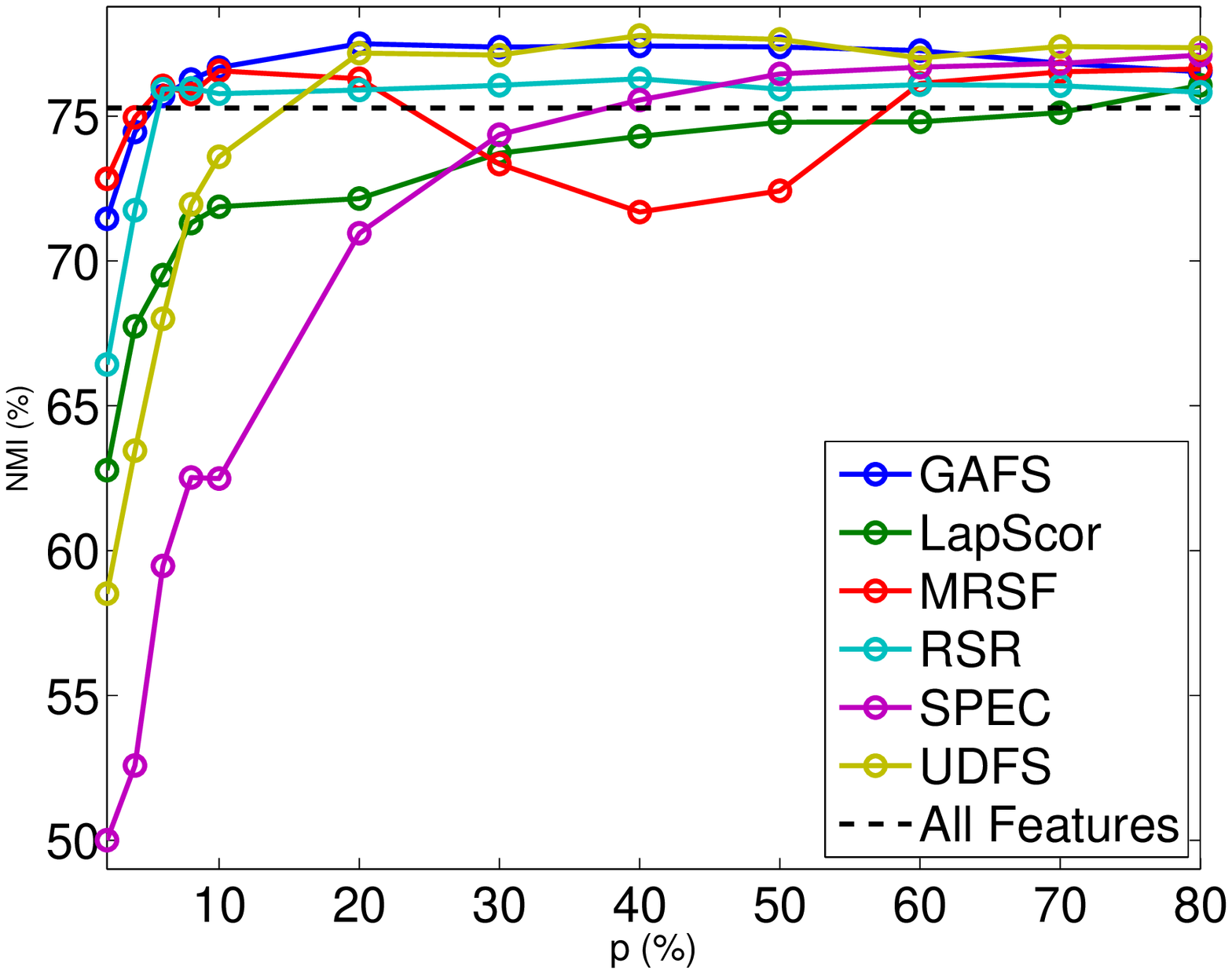}}
  \centerline{(b) COIL20}
\end{minipage}
\hfill
\begin{minipage}{0.2\linewidth}
  \centerline{\includegraphics[width=4.0cm]{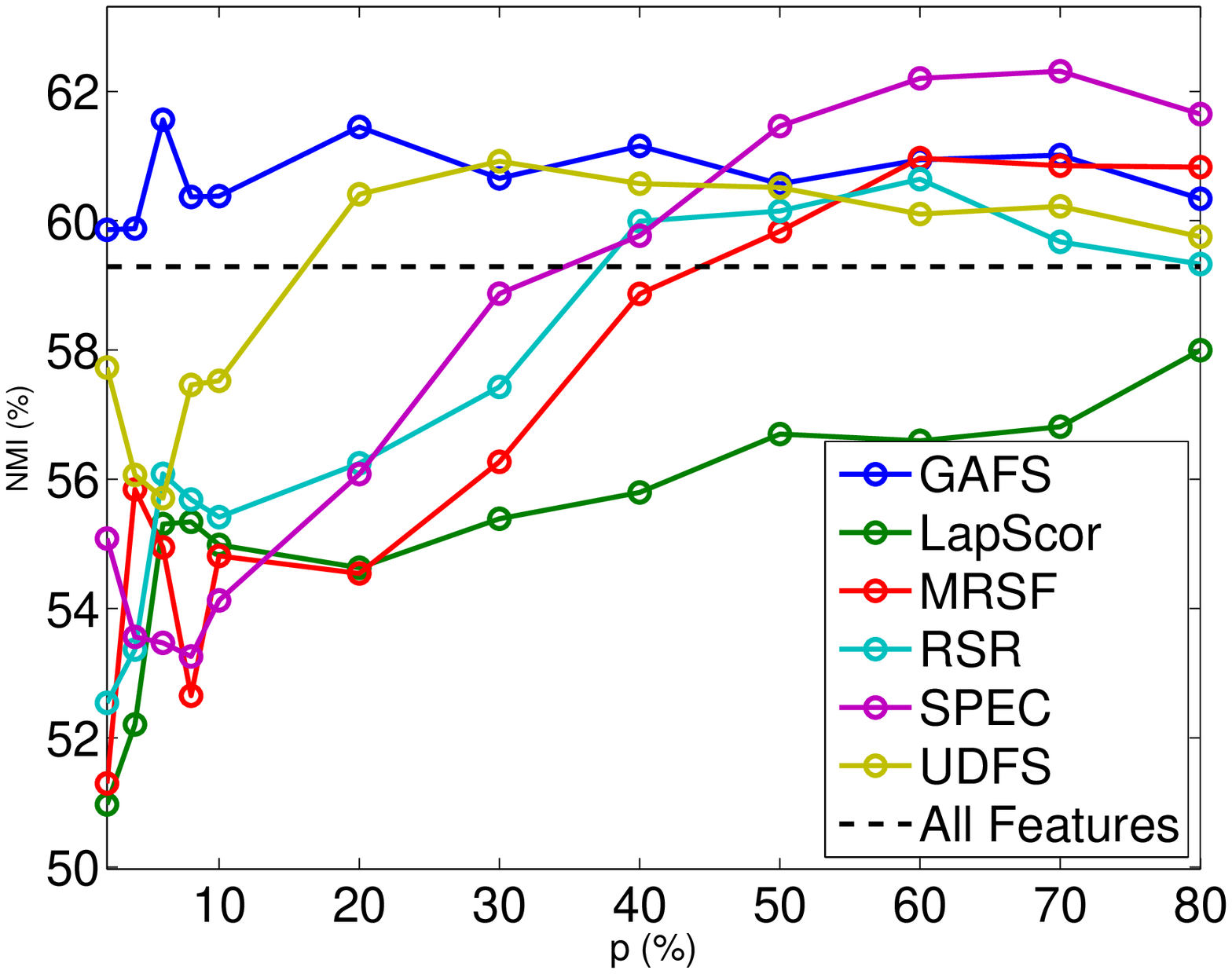}}
  \centerline{(c) Yale}
\end{minipage}
\hfill
\begin{minipage}{0.2\linewidth}
  \centerline{\includegraphics[width=4.0cm]{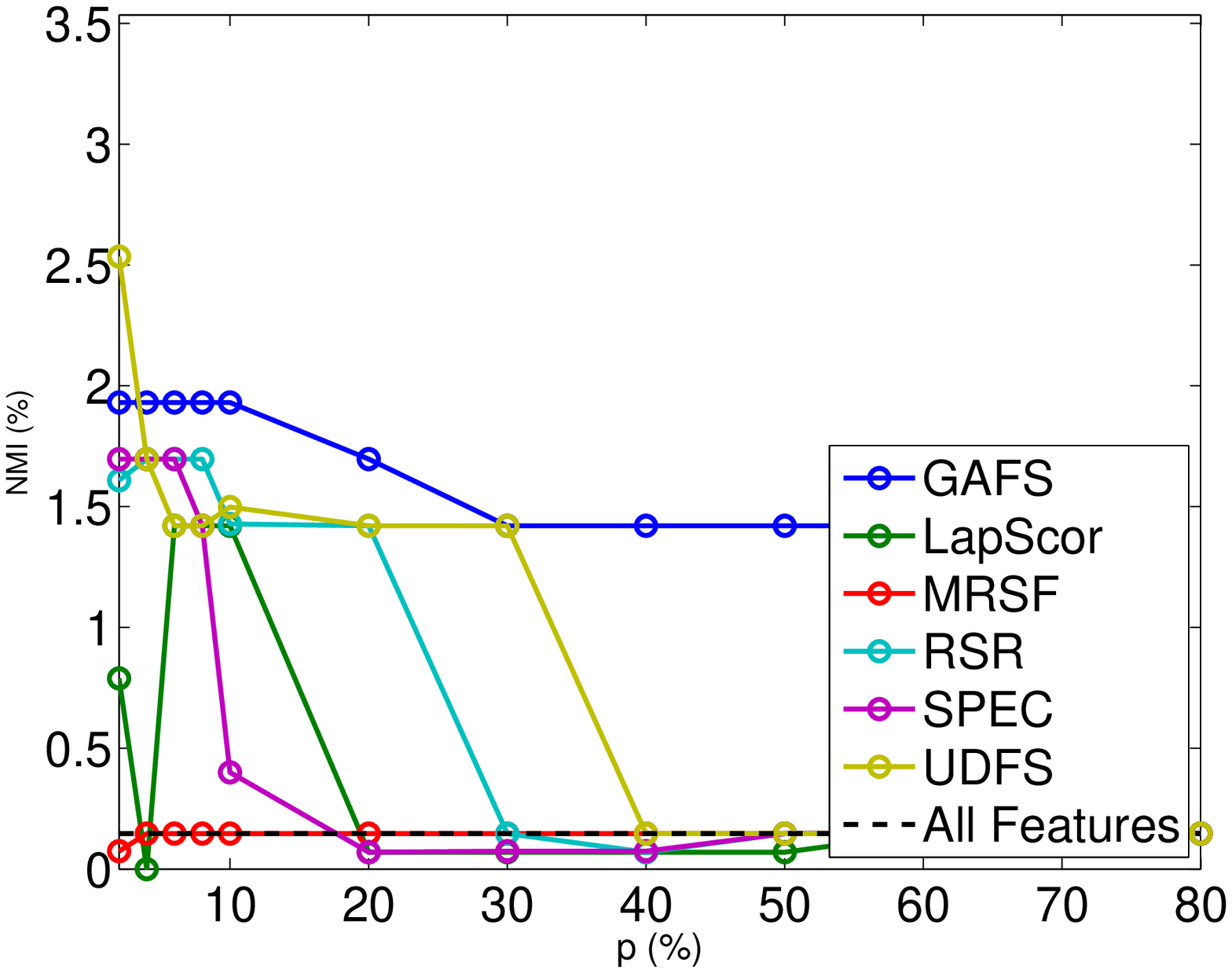}}
  \centerline{(d) PCMAC}
\end{minipage}
\vfill
\begin{minipage}{0.2\linewidth}
  \centerline{\includegraphics[width=4.0cm]{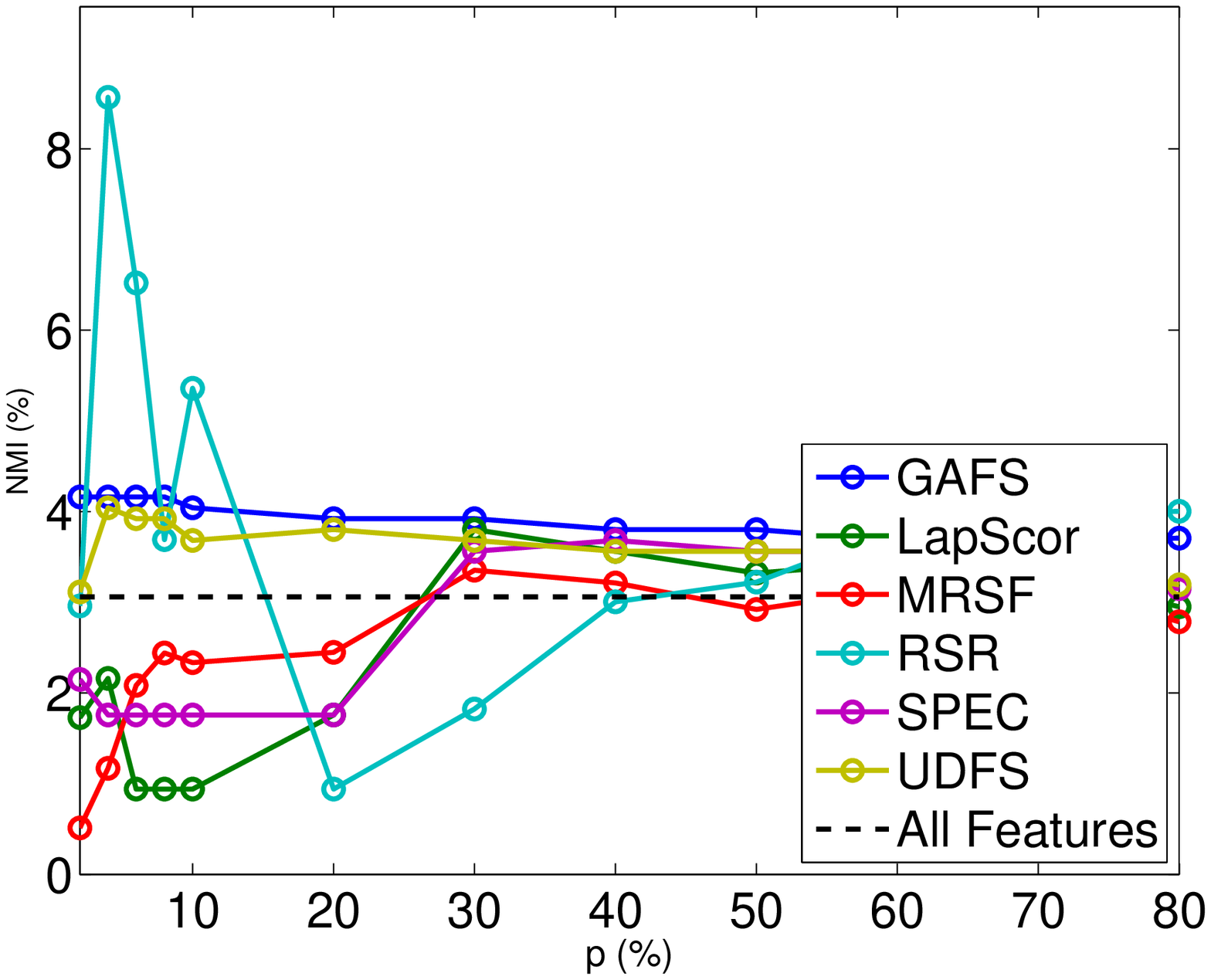}}
  \centerline{(e) BASEHOCK}
\end{minipage}
\hfill
\begin{minipage}{0.2\linewidth}
  \centerline{\includegraphics[width=4.0cm]{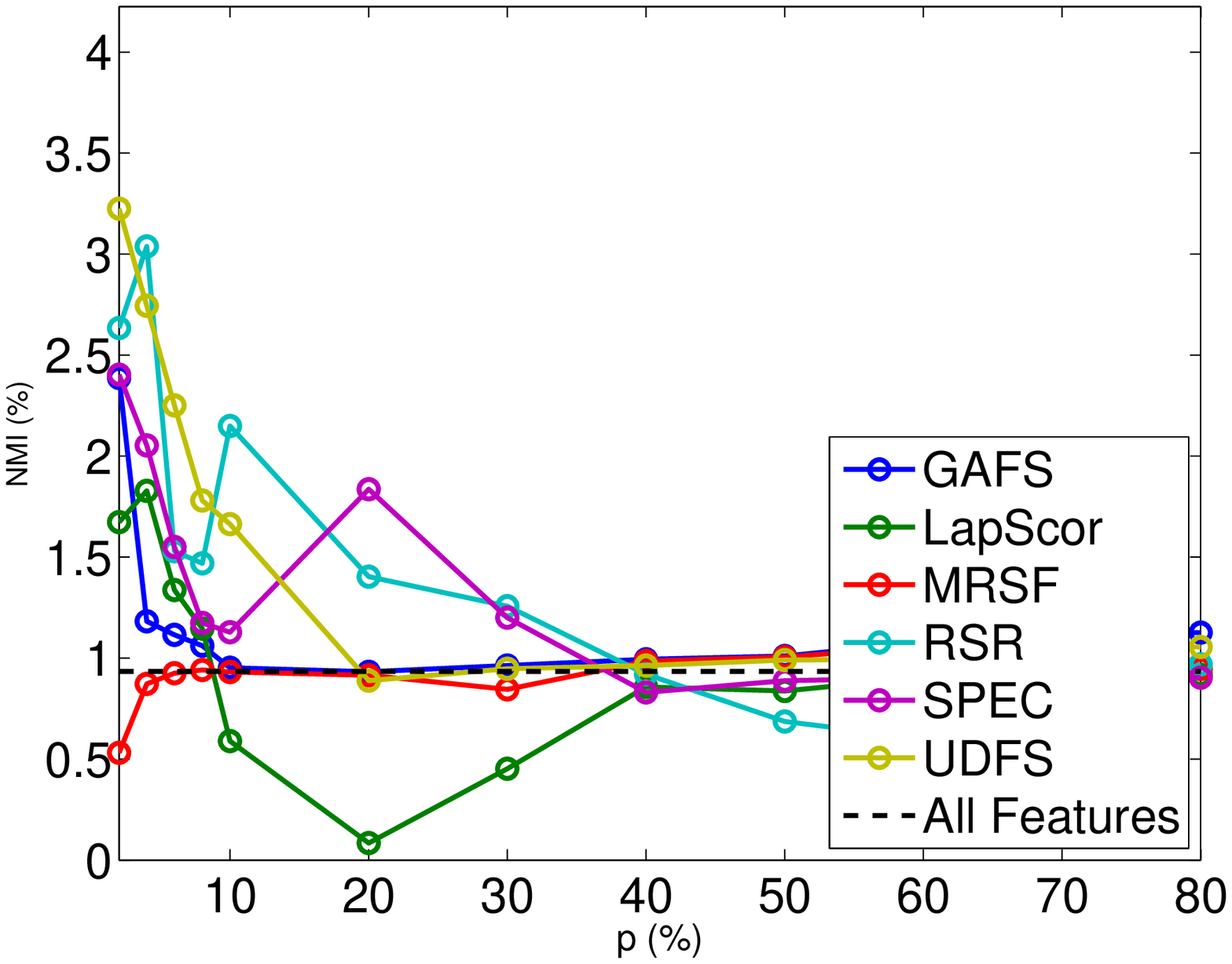}}
  \centerline{(f) RELATHE}
\end{minipage}
\hfill
\begin{minipage}{0.2\linewidth}
  \centerline{\includegraphics[width=4.0cm]{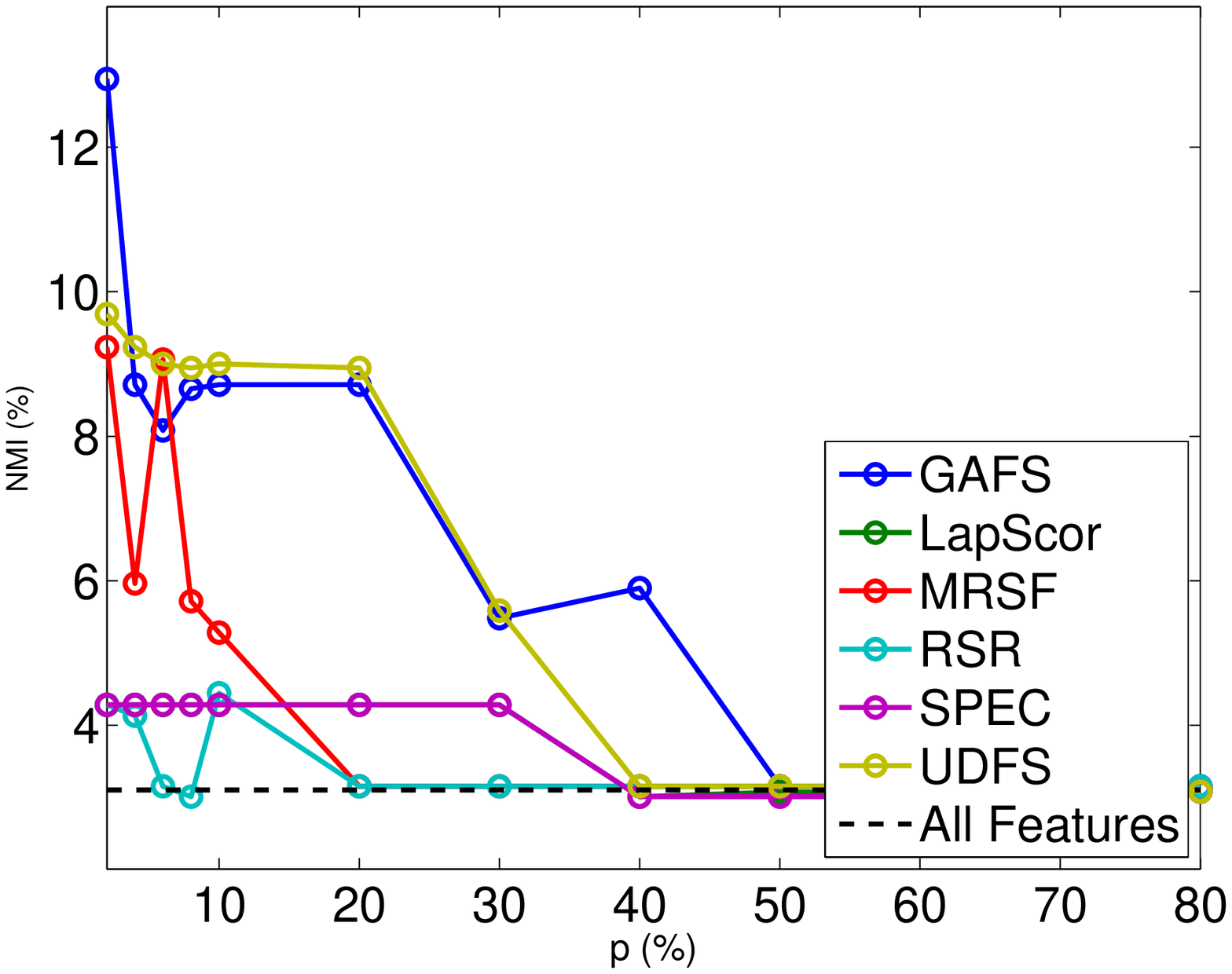}}
  \centerline{(g) Prostate\_GE}
\end{minipage}
\hfill
\begin{minipage}{0.2\linewidth}
  \centerline{\includegraphics[width=4.0cm]{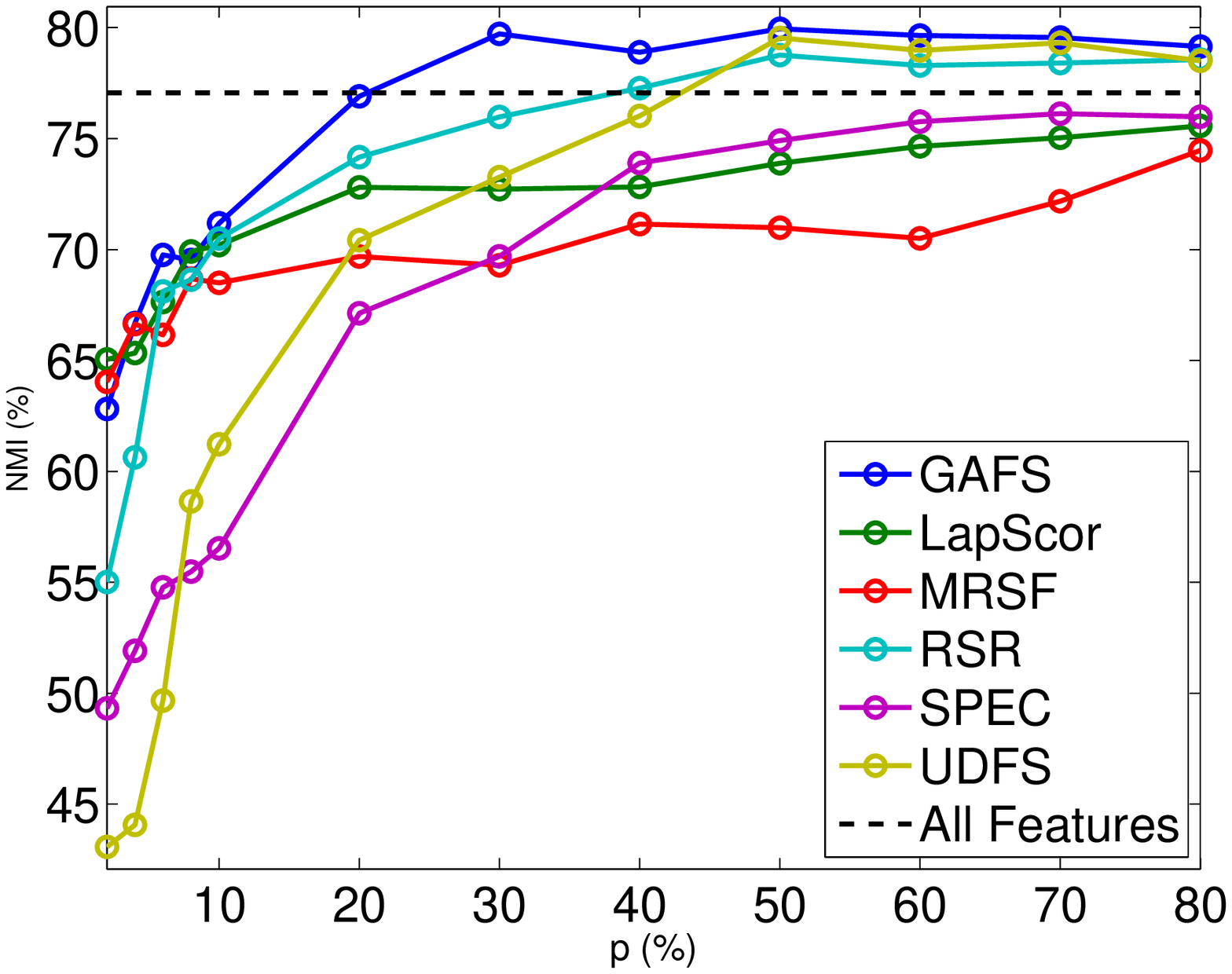}}
  \centerline{(h) Isolet}
\end{minipage}
\vfill
\begin{minipage}{0.2\linewidth}
\end{minipage}
\hfill
\begin{minipage}{0.2\linewidth}
  \centerline{\includegraphics[width=4.0cm]{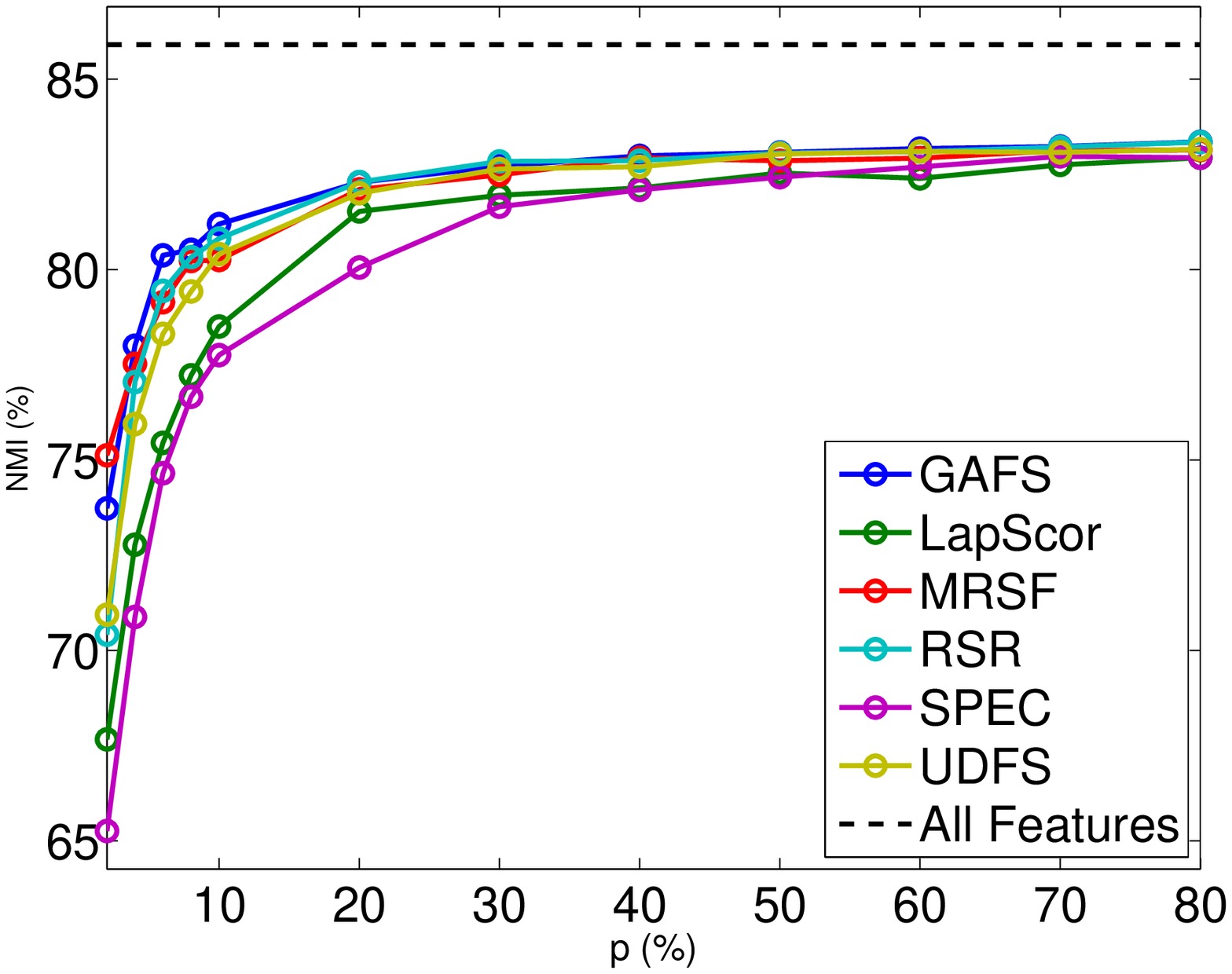}}
  \centerline{(g) Caltech101}
\end{minipage}
\hfill
\begin{minipage}{0.2\linewidth}
  \centerline{\includegraphics[width=4.0cm]{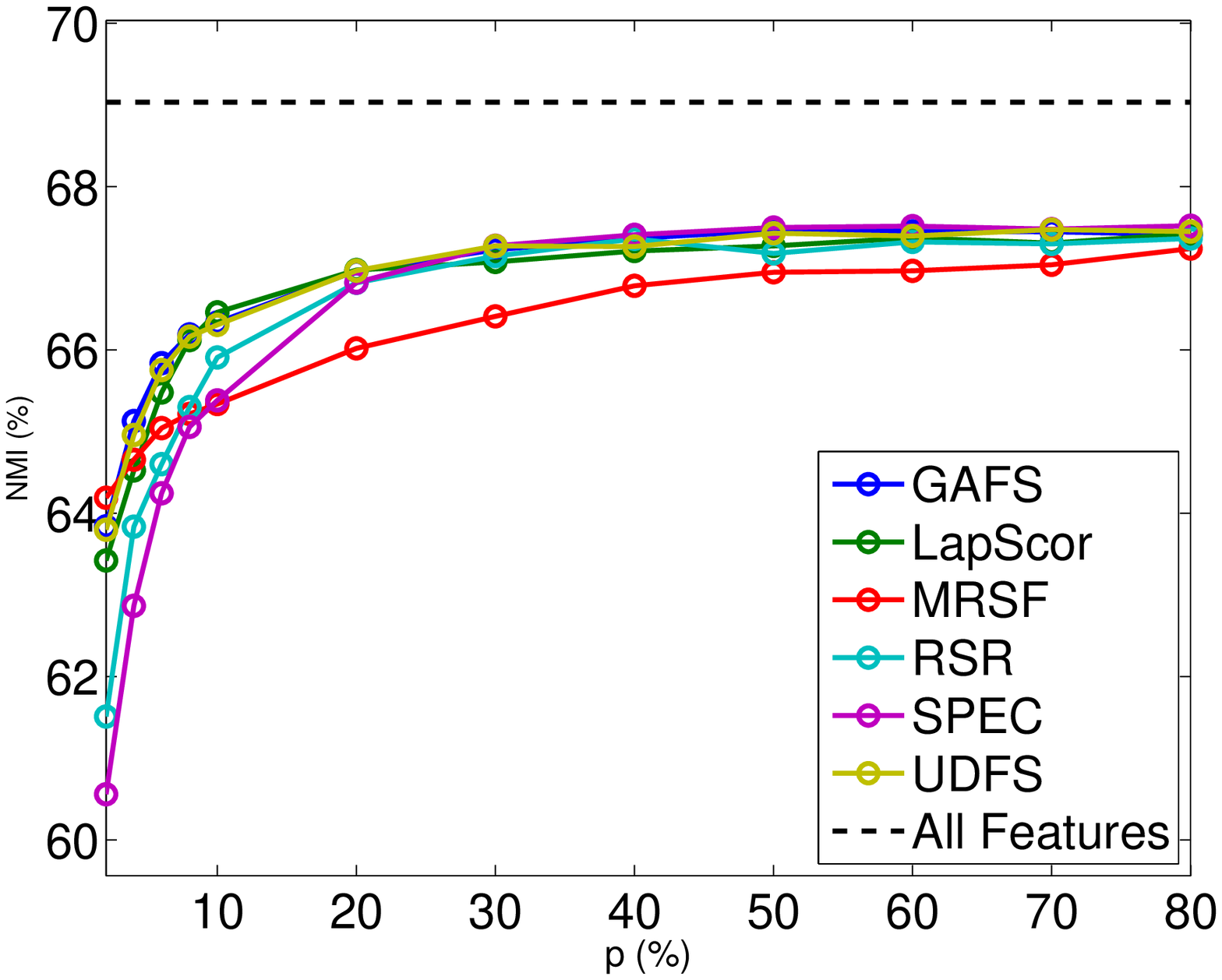}}
  \centerline{(h) CUB200}
\end{minipage}
\hfill
\begin{minipage}{0.2\linewidth}
\end{minipage}
\caption{{Performance of GAFS and competing feature selection algorithms in clustering as a function of the percentage of features selected $p$ ($\%$). Normalized mutual information is used as the evaluation metric.}}
\label{NMI}
\end{figure*}

We present the classification accuracy, ACC, and NMI results of GAFS and the comparison feature selection algorithms on all datasets in Fig.~\ref{clsACC}, Fig.~\ref{ACC}, and Fig.~\ref{NMI}, respectively. From these figures, we can find that GAFS performs better than other compared algorithms in most cases. {There exist some cases that GAFS does not provide the best performance (e.g. PCMAC in classification tasks), but it defeats all its competitors in clustering tasks on PCMAC. We can also observe that GAFS can defeat most competing methods in most cases for both classification and clustering. Therefore, we can say that GAFS provides the best overall performance among the methods that we consider in those figures. Providing a justification for the degradation in performance can be complicated because many factors such as the number of features, evaluation metric, dataset properties, etc. can affect the final performance.} Comparing the performance of GAFS with that of using all features, which is represented by a black dashed line in each figure, we can find that GAFS can always achieve better performance with far less features. {Meanwhile}, with fewer features, the computational load in corresponding classification and clustering tasks can be decreased. These results demonstrate the effectiveness of GAFS in terms of removing irrelevant and redundant features in classification and clustering tasks.

\subsection{Convergence Analysis}
\begin{figure*}
\begin{minipage}{0.2\linewidth}
  \centerline{\includegraphics[width=4.0cm]{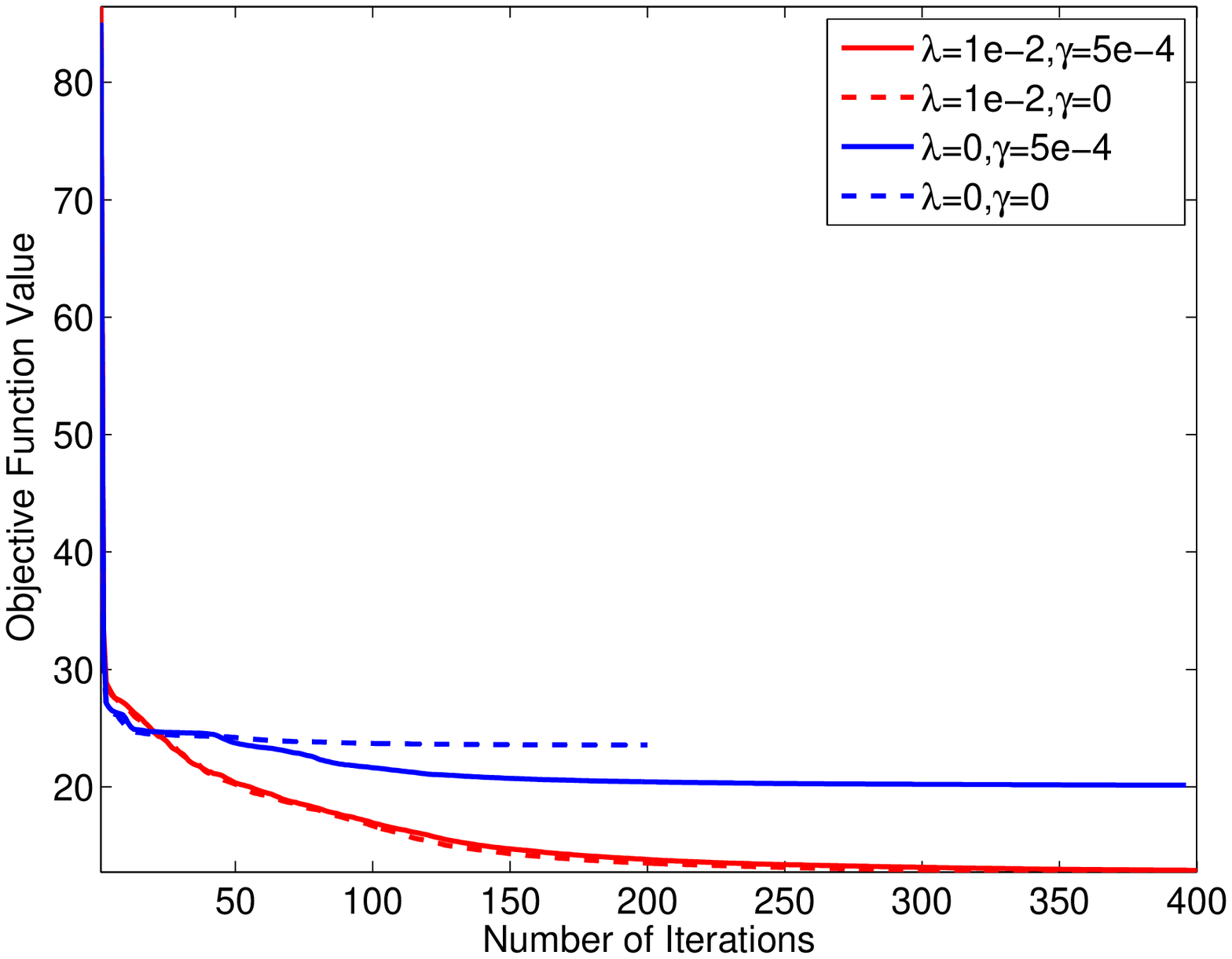}}
  \centerline{(a) MNIST}
\end{minipage}
\hfill
\begin{minipage}{0.2\linewidth}
  \centerline{\includegraphics[width=4.0cm]{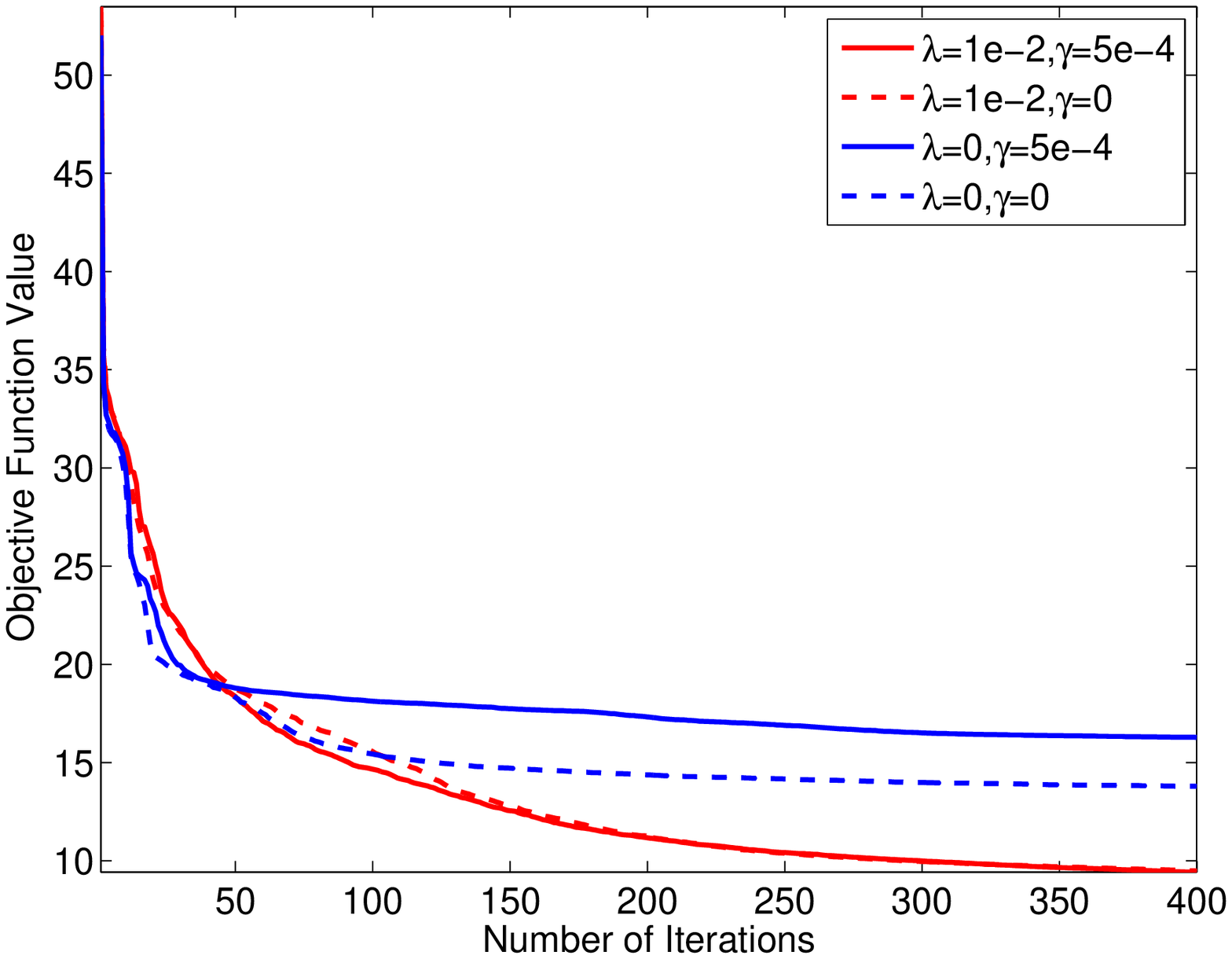}}
  \centerline{(b) COIL20}
\end{minipage}
\hfill
\begin{minipage}{0.2\linewidth}
  \centerline{\includegraphics[width=4.0cm]{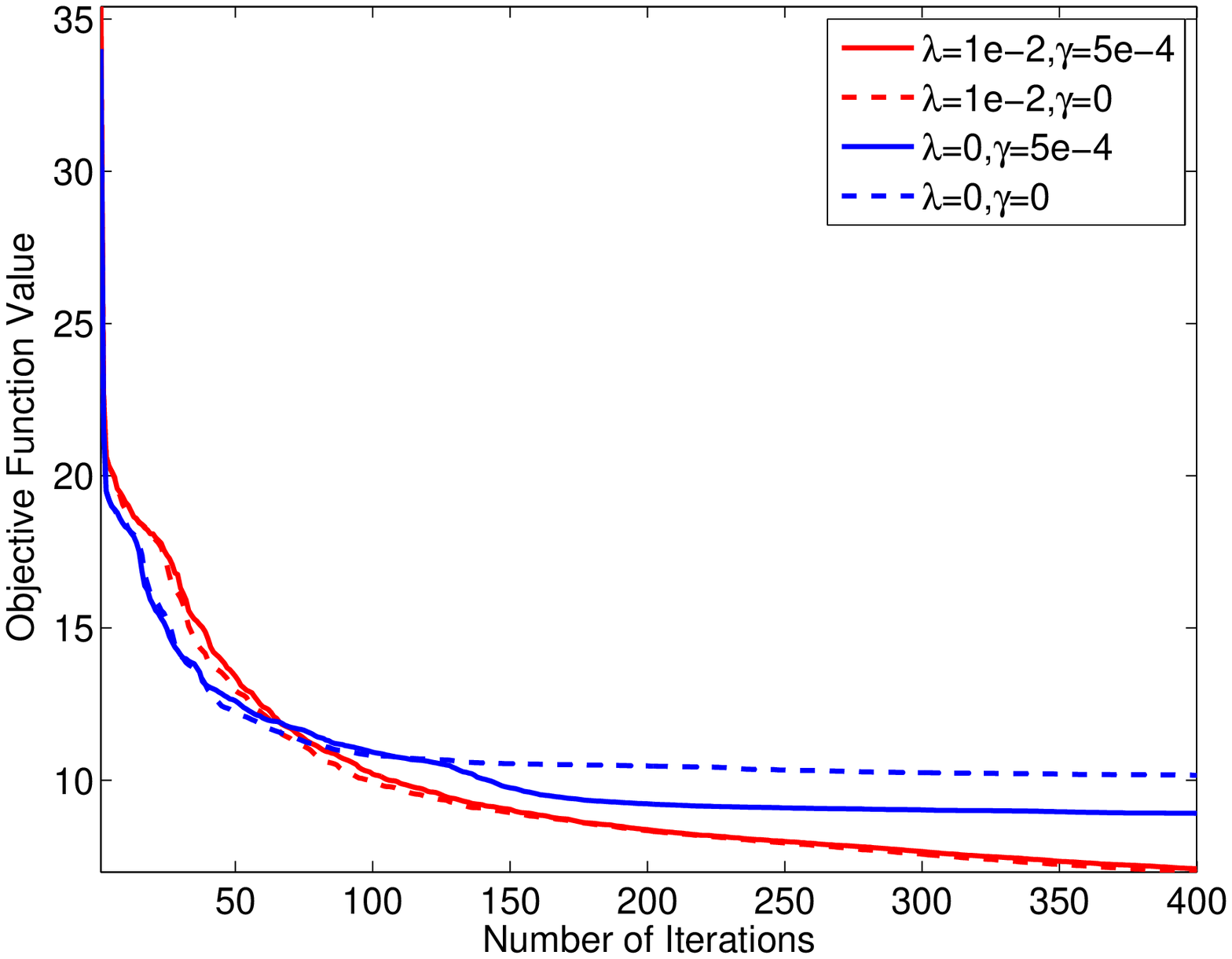}}
  \centerline{(c) Yale}
\end{minipage}
\hfill
\begin{minipage}{0.2\linewidth}
  \centerline{\includegraphics[width=4.0cm]{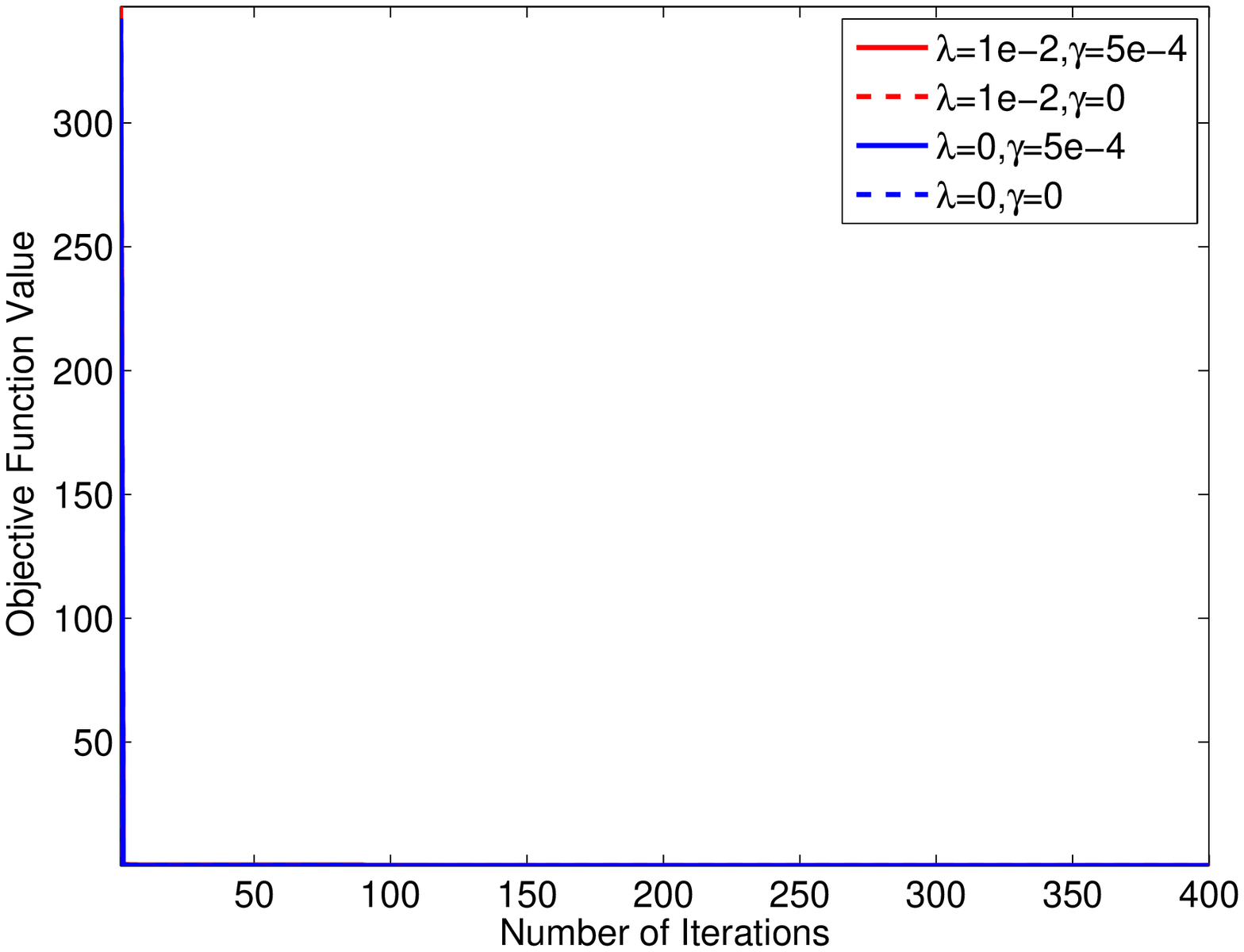}}
  \centerline{(d) PCMAC}
\end{minipage}
\vfill
\begin{minipage}{0.2\linewidth}
  \centerline{\includegraphics[width=4.0cm]{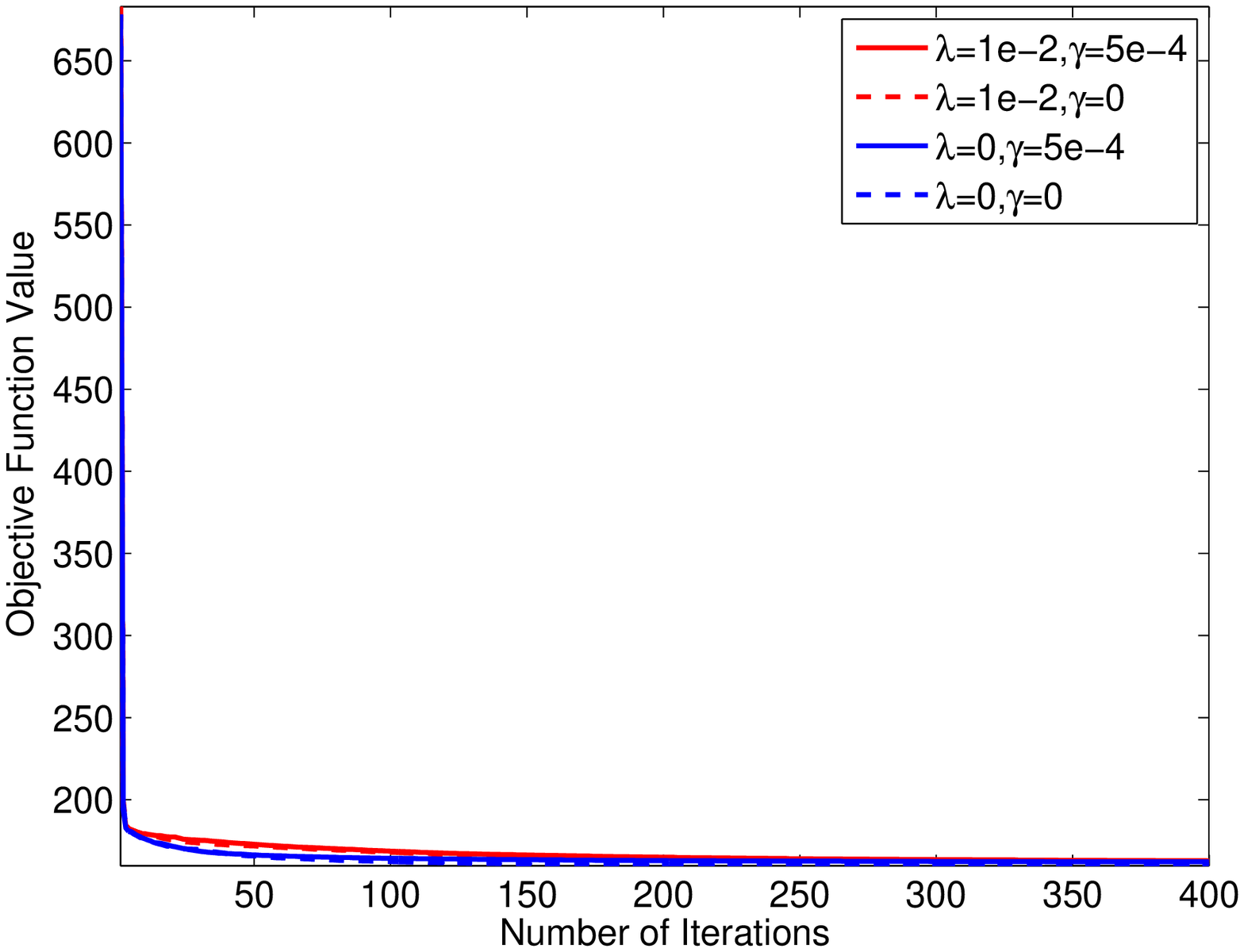}}
  \centerline{(e) BASEHOCK}
\end{minipage}
\hfill
\begin{minipage}{0.2\linewidth}
  \centerline{\includegraphics[width=4.0cm]{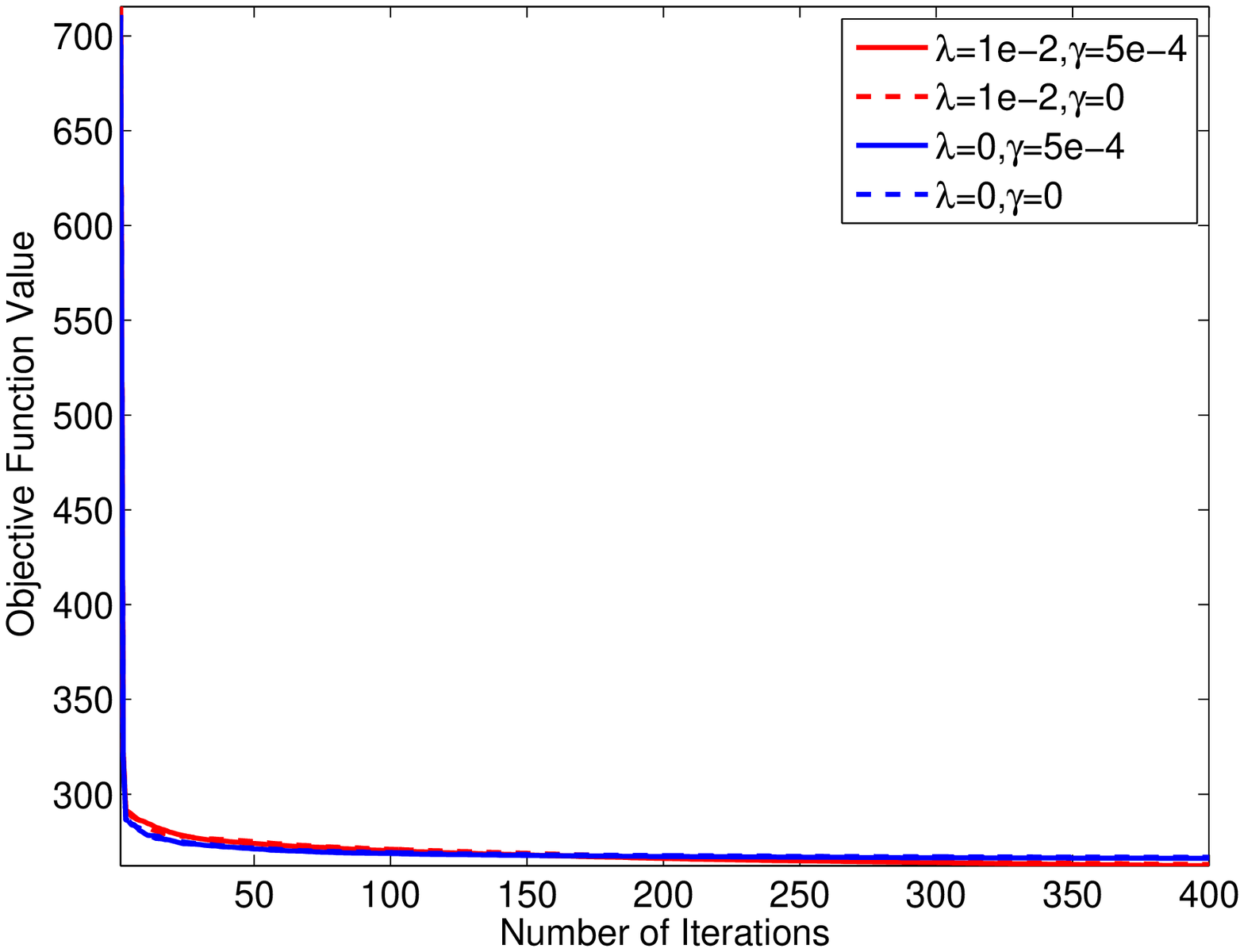}}
  \centerline{(f) RELATHE}
\end{minipage}
\hfill
\begin{minipage}{0.2\linewidth}
  \centerline{\includegraphics[width=4.0cm]{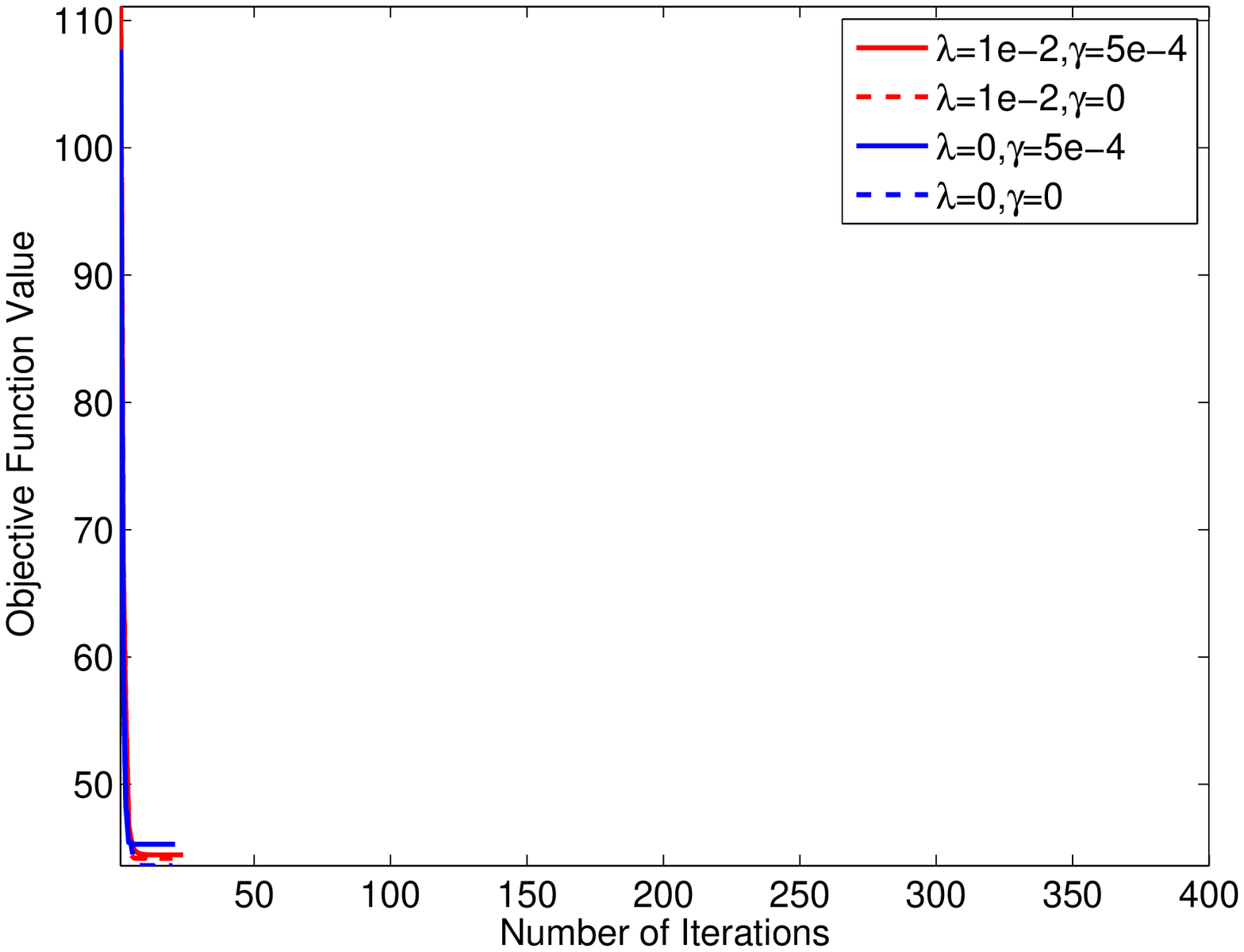}}
  \centerline{(g) Prostate\_GE}
\end{minipage}
\hfill
\begin{minipage}{0.2\linewidth}
  \centerline{\includegraphics[width=4.0cm]{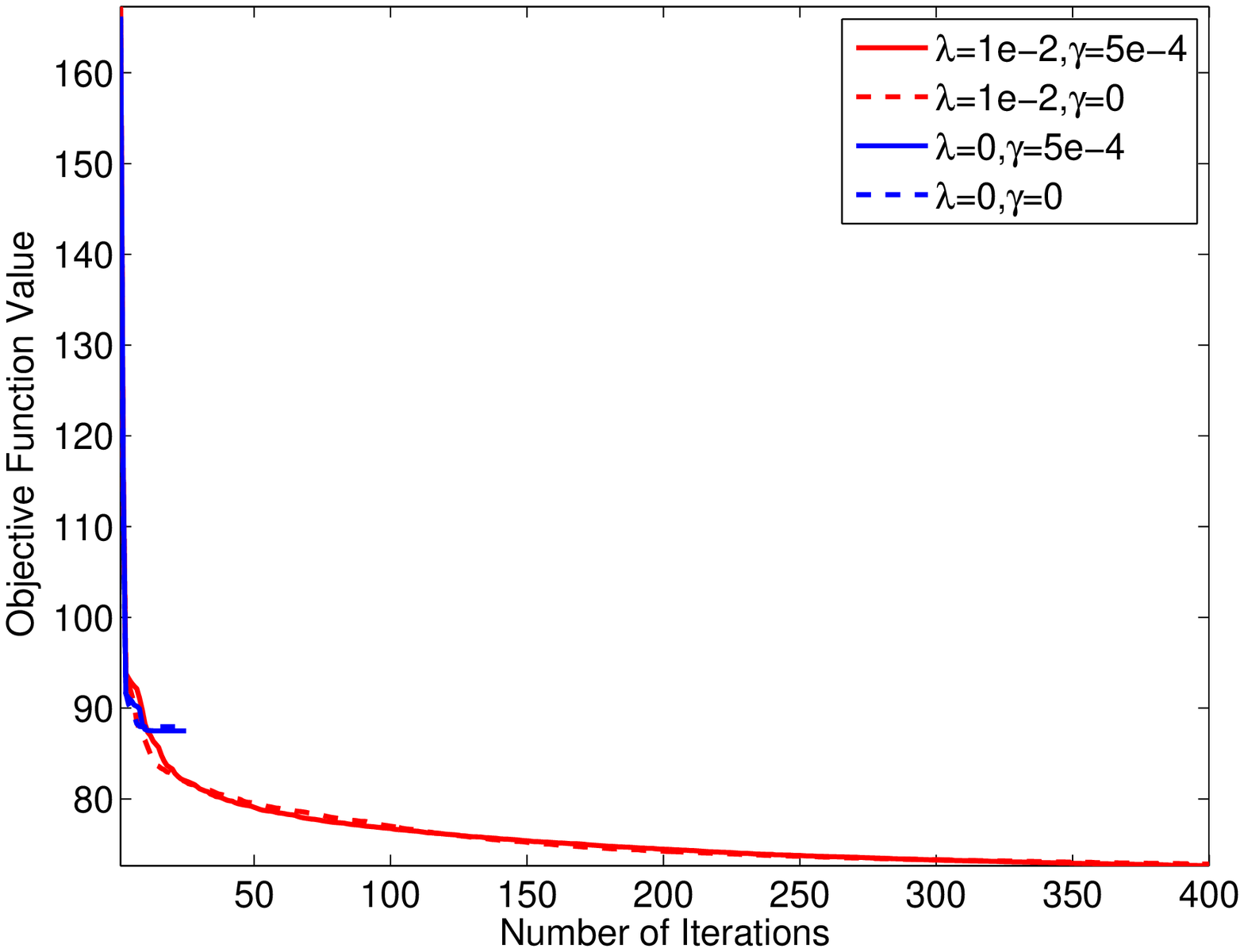}}
  \centerline{(h) Isolet}
\end{minipage}
\vfill
\begin{minipage}{0.2\linewidth}
\end{minipage}
\hfill
\begin{minipage}{0.2\linewidth}
  \centerline{\includegraphics[width=4.0cm]{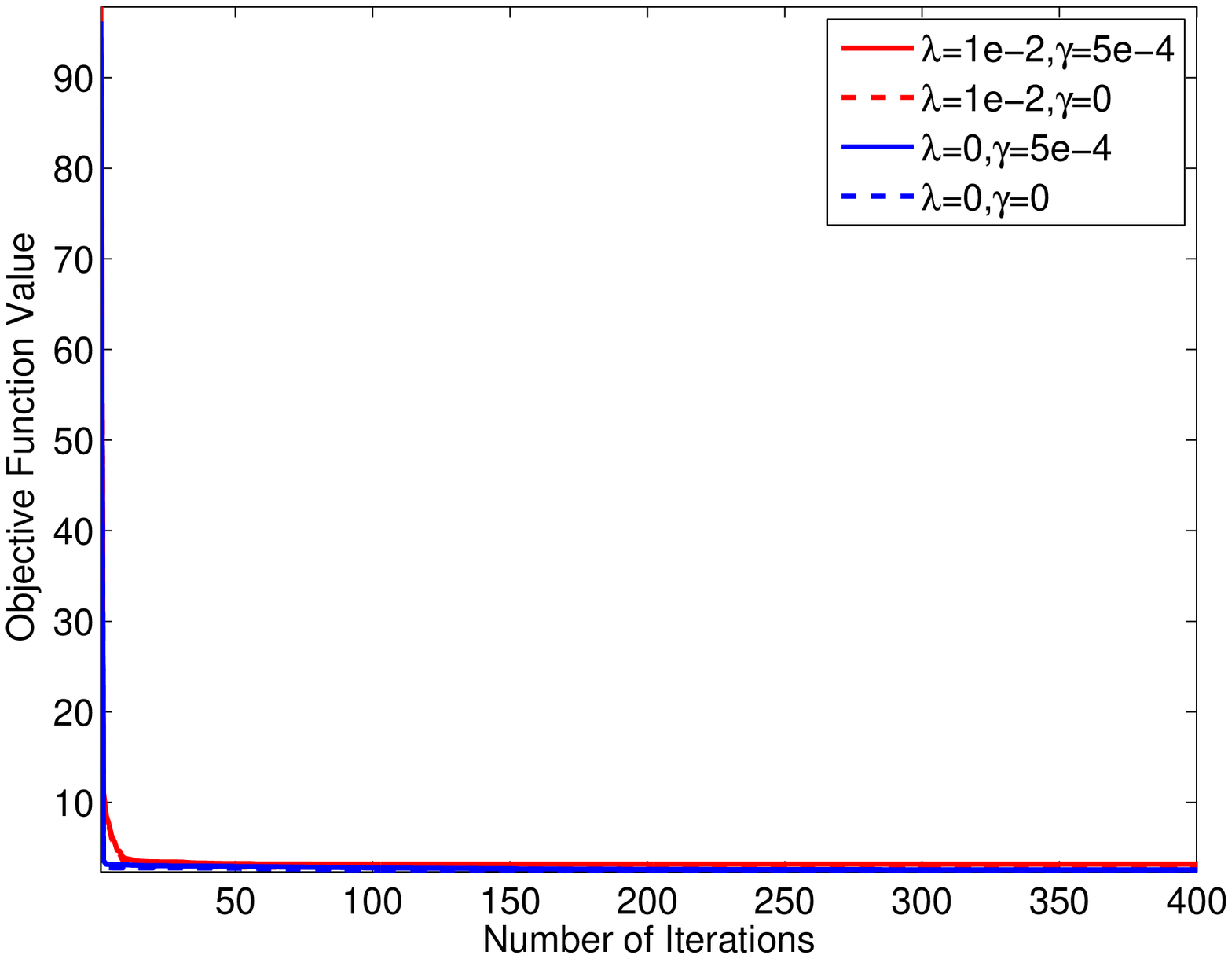}}
  \centerline{(g) Caltech101}
\end{minipage}
\hfill
\begin{minipage}{0.2\linewidth}
  \centerline{\includegraphics[width=4.0cm]{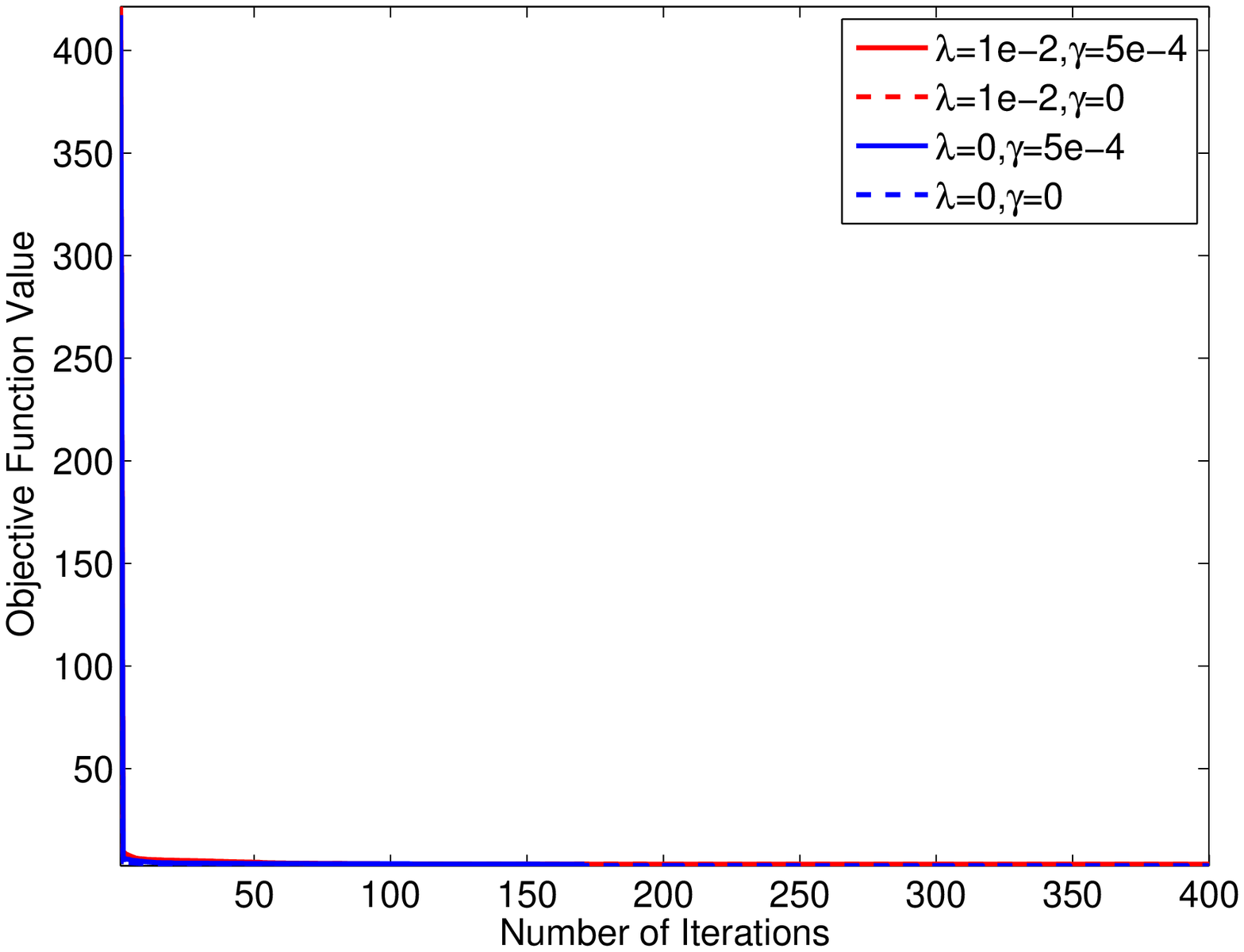}}
  \centerline{(h) CUB200}
\end{minipage}
\hfill
\begin{minipage}{0.2\linewidth}
\end{minipage}
\caption{{Value of the GAFS objective function as a function of the number of iterations for several values of the balance parameters. }}
\label{convergAnalysis}
\end{figure*}

{{In this section, we study the convergence performance of our proposed algorithm. The convergence curves of GAFS on all datasets are shown in Fig.~\ref{convergAnalysis}. Each figure displays the objective function value as a function of the number of iterations under $4$ parameter combinations. The maximum number of iterations in each figure is $400$. In some cases the curves do not reach the maximum number of iterations because they meet the stopping criterion, i.e., the relative difference of objective function values between two iterations is less than $10^{-5}$. We can find that the overall convergence rates on MNIST, COIL20, Yale, and Isolet are slower than other datasets. The reason may be the redundancy between features. For PCMAC, BASEHOCK, RELATHE, and Prostate\_GE, the input features have little connections between each other. Though both Caltech101 and CUB200 are natural image datasets, we employ VGG19 to generate features, which also reduces redundancy among features. But for MNIST, COIL20, Yale, and Isolet, we use the original visual and audio features for feature selection. The high spatial or temporal redundancy among features may reduce the convergence rate. \par
For most datasets, the choice of parameters does not affect convergent results. However, for MNIST, COIL20, Yale, and Isolet, the condition of $\lambda=0$, which implies the abandon of $\ell_{2,1}-$norm term, can lead to a faster convergence rate. This may be due to the use of subgradients during optimization. Since we set the gradient value to be $0$ when the elements in the corresponding column of $\mathbf{W}_1$ are all zeros. The setup of gradients can lead to suboptimality during optimization, which may reduce the speed of convergence. However, when $\lambda \neq 0$ ($\lambda=10^{-2}$ in this illustration), the objective function can be optimized to smaller values. This indicates that the introduction of $\ell_{2,1}$-norm leads to reduction in reconstruction errors. }}

\subsection{Training Time Analysis}
\begin{figure}
\centerline{\includegraphics[width=6.0cm]{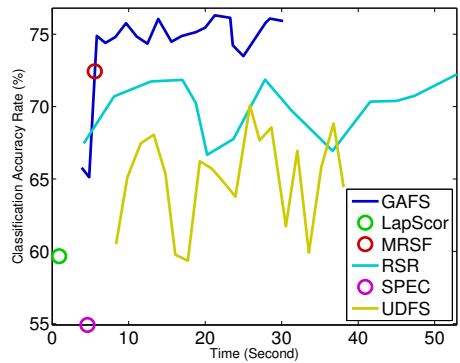}}
\caption{{Classification performance on the MNIST dataset as a function of training time for GAFS and competing feature selection algorithms for $p = 10\%$.}}
\label{trainingtime}
\end{figure}

{{In this section we study the training costs of GAFS as well as competing methods. We record the training time for each method as well as the corresponding feature selection performance. Due to limited space, we only show the results generated from the MNIST dataset mentioned in Section \ref{data}. We use classification accuracy on $10\%$ of all features as the evaluation metric. For Laplacian Score, SPEC, and MRSF, we report the training time and classification accuracy. For GAFS, RSR, and UDFS, we fixed the number of iterations in the optimization procedure from 20 to 400 with a step of 20, while also limiting execution times to a maximum of 50 seconds. We recorded training time for each number of iterations and the corresponding classification rates. The results are displayed in Fig. \ref{trainingtime}. Though Laplacian Score and SPEC have the smallest training times, they cannot provide comparable classification performance. The performance of GAFS starts to stabilize when the training time is around $5$ seconds, which corresponds to $60$ iterations. After that, GAFS consistently provides the best performance among all methods. The performance of RSR is better than GAFS for small training times, but its upper limit in terms of classification accuracy is lower than that of GAFS. MRSF provided satisfactory classification performance ($72.44\%$) with a short training time (4.70s). However, it is also outperformed by GAFS.}}

\section{Conclusion}
\label{conclusion}
In this paper, we proposed a graph and autoencoder-based unsupervised feature selection (GAFS) method. Unlike {similar existing techniques that combine} sparse learning and feature selection, the proposed method projects the data to a lower-dimensional space using a single-layer autoencoder, in contrast to the linear transformation used by most existing methods. With our proposed framework, we bypass the limitation of existing methods {with linear dimensionality reduction schemes}, which may lead to performance degradation for datasets with richer structure {that is}  predominant in modern datasets. Experimental results demonstrate the advantages of GAFS versus methods in the literature for both classification and clustering tasks. \par
The work we present here is our first attempt to leverage autoencoders for unsupervised feature selection purposes. Therefore, we use the most standard setting for the construction of the autoencoder, e.g., there is no desired or particular structure to the activations or the reconstruction error. In the future, we plan to explore the effectiveness of more elaborate versions of an autoencoder for feature selection purposes. {{Furthermore, by employing label information, we can also extend our work to a supervised feature selection framework.}} \par





\bibliographystyle{elsarticle-num}
\biboptions{sort&compress}
\bibliography{Literature_Review.bib}







\end{document}